\documentclass{article}

\usepackage{arxiv}

\usepackage[utf8]{inputenc}
\usepackage[T1]{fontenc}
\usepackage{url}
\usepackage{booktabs}
\usepackage{nicefrac}
\usepackage{microtype}
\usepackage{lipsum}

\usepackage[english]{babel}

\usepackage[backref,hyperindex=true,pagebackref=true]{hyperref}

\usepackage{mathptmx}

\usepackage{longtable}
\usepackage[font=normalsize]{caption}
\usepackage{subcaption}

\usepackage{algorithm}
\usepackage{algorithmic}

\usepackage{amsfonts}
\usepackage{amsmath}
\newcommand{\Mod}[1]{\ (\mathrm{mod}\ #1)}

\usepackage[titletoc]{appendix}

\usepackage{graphicx}

\usepackage[printonlyused,withpage]{acronym}

\usepackage{listings}
\usepackage{xcolor}
\usepackage[printonlyused,withpage]{acronym}

\usepackage{titlesec}

\titleformat{\paragraph}
  {\normalfont\normalsize\bfseries}{\theparagraph}{1em}{}
\titlespacing*{\paragraph}
  {0pt}{2.25ex plus 1ex minus .2ex}{0.75ex plus .2ex}

\title{Path Planning through Multi-Agent Reinforcement Learning in Dynamic Environments}

\author{
Jonas De Maeyer\\
  Department of Computer Science\\
  University of Antwerp (UA)\\
  \texttt{jonas.demaeyer@student.uantwerpen.be} \\
\And
Hossein Yarahmadi\\
  Department of Computer Engineering, Faculty of Engineering,\\
  Ayatollah Boroujerdi University, Boroujerd, Iran\\
  \texttt{hyarahmadi@abru.ac.ir} \\
\And
Moharram Challenger\\
  Department of Computer Science\\
  University of Antwerp (UA) and Flanders Make\\
\texttt{moharram.challenger@uantwerpen.be} \\
}

\begin{document}

\maketitle

\phantomsection
\addcontentsline{toc}{section}{Abstract}

\begin{abstract}

In this technical report, we address the path planning problem in dynamic environments, i.e., settings in which obstacles can change over time, introducing uncertainty. Effective path planning in such environments is essential for mobile robots, as many real-world scenarios naturally exhibit dynamic elements. While prior research has proposed various approaches, our work tackles the problem from a relatively unexplored angle.
We start by relaxing the common assumption that environmental changes are entirely unlocalizable. In practice, such changes are often confined to a bounded region, even if their exact positions are unknown. Leveraging this assumption enables more efficient replanning, as only affected regions need to be updated. This principle underlies the methodology proposed by Yarahmadi et al., which addresses the path planning problem in two ways. If no changes occur, the environment is treated as static and fully known, and a global path planner (e.g., A*) is used to generate a route to a charging station. If changes do occur, the environment is partitioned into sub-environments, with each assigned to a dedicated agent. Agents responsible for the affected sub-environments adapt by performing local path planning based on Q-learning, a \ac{RL} method.
While promising, their methodology has notable limitations. First, relying on a global path planner is impractical in unknown environments where the layout, including free spaces, obstacles, and charging stations, is initially hidden. Second, traditional path planning algorithms such as A* exhibit poor scalability and efficiency in large environments. Furthermore, their method triggers \ac{RL}-based replanning whenever a sub-environment experiences a change. This strategy is inefficient, as some changes have little to no impact on the overall path planning problem, making constant replanning unnecessary.
Another key limitation is that the sub-environments created by partitioning the environment have no defined relationships with one another. As a result, if a charging station is absent or unreachable within a sub-environment, the local planner is stuck and cannot utilize adjacent regions to find an alternative path. This lack of interconnection prevents fallback strategies and offers no guarantee of reaching a charging station. Finally, their evaluation is conducted in overly simplistic environments with minimal dynamic changes, i.e., only one obstacle change per time step.
To address these limitations, we propose the following contributions:

\begin{itemize}
  \item A scalable, region-aware \ac{RL} framework based on a hierarchical decomposition of the environment, which supports efficient, targeted retraining.
  \item A retraining condition based on sub-environment success rates, which determines if retraining is necessary after a change or potentially multiple changes.
  \item Both single-agent and multi-agent (federated Q-learning) \ac{RL}-based training methods, where the multi-agent version aggregates local Q-tables to accelerate the learning process.
  \item A more realistic evaluation setup that spans three levels of environment difficulty and simulates multiple simultaneous obstacle changes per time step.
  \item A visualization tool that demonstrates the evolution of learned policies over time, helping to illustrate the system’s adaptability.
\end{itemize}

The simulation results show that the two multi-agent approaches, \texttt{fedAsynQ\_EqAvg} and \texttt{fed\allowbreak AsynQ\_ImAvg}, consistently perform best, achieving high accuracy across all environment sizes and difficulty levels. Their success rates approach those of the \texttt{A* Oracle} approach (an idealized upper bound) while maintaining low adaptation times, demonstrating excellent scalability. Although their path lengths are slightly longer than those of \texttt{A* Oracle}, they remain acceptable, given the emphasis on efficient and robust adaptation over optimality.
Our findings demonstrate the effectiveness of \ac{DAI} and show that \ac{MARL} outperforms the single-agent counterparts. Nonetheless, our approach has limitations. The hierarchical decomposition currently only supports environments with known dimensions and rectangular or square shapes. Initial training in large, unknown environments is time-intensive due to the exploratory nature of tabular \ac{RL} and should be considered a practical drawback.
This report offers a promising foundation for future work. Potential directions include integrating \ac{Deep RL} for improved generalization and performance, as well as extending the decomposition framework to support arbitrary or unknown environment shapes, thereby making the methodology more universally applicable.

\end{abstract}


\clearpage

\tableofcontents

\clearpage

\listoffigures

\clearpage

\section*{List of Acronyms}
\markboth{List of Acronyms}{List of Acronyms}

\begin{acronym}
    \setlength{\parskip}{-0.5ex}
    \acro{RL}{Reinforcement Learning}
    \acro{DAI}{Distributed Artificial Intelligence}
    \acro{MARL}{Multi-Agent Reinforcement Learning}
    \acro{Deep RL}{Deep Reinforcement Learning}
    \acro{MDP}{Markov Decision Process}
    \acro{UAV}{Unmanned Aerial Vehicle}
    \acro{MAS}{Multi-Agent System}
    \acro{DNN}{Deep Neural Network}
    \acro{GPU}{Graphics Processing Unit}
    \acro{CPU}{Central Processing Unit}
    \acro{CTCE}{Centralized Training and Centralized Execution}
    \acro{DTDE}{Decentralized Training and Decentralized Execution}
    \acro{CTDE}{Centralized Training and Decentralized Execution}
    \acro{TD}{Temporal-Difference}
    \acro{DQN}{Deep Q-Network}
    \acro{IMARL}{Improved Multi-Agent Reinforcement Learning}
    \acro{E-DCM-MULTI-Q}{Evolving Dynamic Correlation Matrix-based Multi-Q learning}
    \acro{MADRL}{Multi-Agent Deep Reinforcement Learning}
    \acro{ASV}{Autonomous Surface Vehicle}
    \acro{M-MDP}{Multi-Agent Markov Decision Process}
    \acro{SE-MDP}{Sequentially Extended Markov Decision Process}
    \acro{CL-MADDPG-PP}{Curriculum Learning Multi-Agent Deep Deterministic Policy Gradient with Proportional Prioritization}
    \acro{MADDPG}{Multi-Agent Deep Deterministic Policy Gradient}
    \acro{MSD-SPP}{multi-source-destination shortest path problems}
    \acro{MAPF}{Multi-Agent Path Finding}
    \acro{LNS2}{Large Neighborhood Search}
    \acro{DWA}{Dynamic Window Approach}
    \acro{CNN}{Convolutional Neural Network}
    \acro{DQL}{Deep Q-Learning}
    \acro{MAPPER}{Multi-Agent Path Planning with Evolutionary Reinforcement Learning}
    \acro{A2C}{Advantage Actor-Critic}
    \acro{AB-Mapper}{Attention and BicNet-based Multi-agent path planning with effective reinforcement}
    \acro{MCAL}{Mobile robot Collision Avoidance Learning}
    \acro{MCAL-P}{Mobile robot Collision Avoidance Learning with Path}
    \acro{SAC}{Soft Actor-Critic}
    \acro{USV}{Unmanned Surface Vehicle}
    \acro{G2RL}{Globally Guided Reinforcement Learning}
    \acro{FOV}{Field Of View}
    \acro{LSTM}{Long Short-Term Memory}
    \acro{DDPG}{Deep Deterministic Policy Gradient}
    \acro{TD-MATD3}{Task Decomposed Multi-Agent Twin Delayed Deep Deterministic Policy Gradient}
    \acro{DPPDRL}{Decentralized Path Planning model using Deep Reinforcement Learning}
    \acro{AGV}{Automated Guided Vehicles}
    \acro{RMFS}{Robotic Mobile Fulfillment System}
    \acro{DOA}{Dynamic Obstacle Avoidance}
    \acro{BFS}{Breadth-First Search}
    \acro{FedAsynQ}{Federated Asynchronous Q-Learning}
    \acro{SFML}{Simple and Fast Multimedia Library}
\end{acronym}

\clearpage

\section{Introduction} \label{sec:introduction}

\subsection{Problem Context}

Path planning is a well-established problem encountered across a wide range of disciplines, from civil engineering~\cite{morad1992path, sivakumar2003automated, kim2003construction, song2019construction, cai2023prediction, lu2025fire} to industrial engineering~\cite{ting2002path, gochev2017path, zhang2018path, fu2018improved}, and is especially prominent in computer science. The goal of path planning is to determine the shortest, collision-free route between two points within a given environment. Desirable characteristics of path planning algorithms include accuracy, i.e., they consistently find valid paths connecting two positions via a sequence of actions. But, they must also be efficient, ensuring that these paths are identified within reasonable time constraints. Given the widespread relevance of path planning, the development of accurate and time-efficient path planning solutions holds significant practical importance across many application domains.

A major distinction among path planning algorithms lies in the type of environment they are designed to handle. In the most common scenario, the environment is fully known in advance and remains static, meaning it does not change over time. Numerous accurate and efficient solutions have been developed for these static settings, including classical algorithms such as Dijkstra’s algorithm, Bellman-Ford, A*, and Floyd-Warshall. The environments can typically be modeled as a graph or a \ac{MDP}. 

The assumption that environments remain static over time is often unrealistic in real-world scenarios. In practice, environments are typically dynamic, evolving due to the movement of obstacles or other environmental factors, introducing significant uncertainty into the path planning process. Robots frequently encounter dynamic settings in contexts such as automated factories or warehouses, where they must navigate efficiently while avoiding collisions with other robots, temporarily placed items, or fixed obstacles such as shelves~\cite{chen2023intelligent}. Similarly, \acp{UAV} operate in even more complex settings, navigating through dynamic three-dimensional spaces with moving obstacles~\cite{tordesillas2021mader}.

The dynamic nature of real-world environments can lead to situations where the environment is partially or entirely unknown, making it difficult, or even impossible, to solve the path planning problem using traditional techniques alone. Classical algorithms such as Dijkstra’s or A* become inadequate under such uncertainty, as they require a complete representation or model of the environment. This limitation creates a pressing need for new algorithms, or adaptations of existing ones, that can handle dynamicity while still producing accurate and efficient paths. Meeting these requirements significantly increases the complexity of the path planning problem and calls for more flexible, adaptive, and intelligent approaches.

\ac{RL} provides a promising alternative to classical path planning, particularly in dynamic and uncertain environments. Because most \ac{RL} techniques are model-free, agents can learn optimal behaviors, referred to as policies, directly through interaction with the environment, without requiring explicit knowledge of its internal dynamics. This is particularly useful in dynamic settings, where it is difficult or impossible to predict how actions will influence future states. The ability to learn purely from experience allows \ac{RL}-based methods to adapt to a wide variety of conditions, making them highly suitable for real-world path planning tasks.

Many \ac{RL}-based approaches have already been proposed to address dynamic environments, and several notable contributions are discussed in Section~\ref{sec:literature-review}. However, a specific research direction remains underexplored. Most \ac{MARL}-based solutions assume that changes in the environment, such as moving obstacles, are unlocalizable. That is, agents are expected to adapt to environmental changes without knowing where those changes occurred. This often leads to inefficient and unnecessary re-planning: agents may revise their strategies even when no relevant changes have occurred within their operational region.

To address this inefficiency, Yarahmadi et al.~\cite{yarahmadi2024comp} proposed a novel \ac{MARL}-based approach that introduces a localized change detection assumption. Their method assumes that changes in the environment can be detected and localized to specific regions, allowing replanning to occur only where it is needed. If no changes are detected in the environment, a global path planner (e.g., A*) is used to navigate to a charging station. If changes are detected, the environment is divided into sub-environments, each managed by an agent responsible for finding paths to a charging station using Q-learning. This targeted replanning strategy greatly reduces unnecessary computation. 

\subsection{Problem Statement and Motivation}

Despite its merits, the approach by Yarahmadi et al. has notable limitations. First, the evaluation environments in their experiments are relatively simple, featuring many open spaces, numerous charging stations, and few obstacles, which makes the path-planning problem far less challenging than in realistic scenarios. Consequently, it is unclear how well their method would perform in more complex or constrained settings. Second, their simulation of dynamic environments is highly restricted: only one obstacle changes position per time step. While this introduces some uncertainty, it fails to capture simultaneous changes that can occur in real-world dynamic systems.

Regarding their methodology, another key limitation is the reliance on the global path planner. In environments not affected by changes, a global path planner (e.g., A*) is used to plan a path to a charging station. In unknown environments where the layout of free spaces, obstacles, and charging stations is initially hidden, the global planner becomes unusable, leaving only \ac{RL}-based planning as an option. Moreover, their approach triggers \ac{RL}-based replanning every time a sub-environment is affected by a change. While this ensures maximum accuracy, it is inefficient; in practice, some changes have little to no impact on the overall path planning problem, making constant replanning unnecessary.

The most critical limitation, in my view, concerns the handling of infeasible sub-environments. Because each agent is confined to its assigned region, it cannot replan a route to a charging station if none exists within its area or if obstacles block all possible paths to one. With no relationships or connections between sub-environments, agents cannot leverage neighboring regions to find alternative routes. This restriction becomes increasingly problematic in complex environments with sparse goals or narrow passages. Addressing this would require enabling agents to reason beyond the boundaries of their assigned sub-environments, a capability that the original approach does not provide.

By addressing these limitations, the methodology becomes both more practical and more scalable in dynamic scenarios. Eliminating the reliance on a global path planner aligns the approach more closely with the principles of \ac{DAI}, where decision-making is distributed and does not depend on a complete view of the environment. Avoiding unnecessary replanning whenever a sub-environment is affected by changes improves efficiency. Relying solely on \ac{RL} also makes the approach applicable to initially unknown environments, where the layout of free spaces, obstacles, and charging stations cannot be assumed in advance. Together, these improvements strengthen the methodology and allow it to handle more realistic and challenging dynamic path planning scenarios.

\subsection{Contributions and Findings}

This technical report addresses these limitations by introducing a hierarchical decomposition strategy. While the core idea remains the same, partitioning the environment into sub-environments, we extend it by recursively subdividing regions, forming a hierarchy of sub-environments structured as a tree. The root represents the entire environment, and the leaves correspond to the smallest, atomic sub-regions. Within this hierarchy, relationships naturally emerge: sub-environments at the same level that share a parent are considered siblings and correspond to adjacent regions in the maze. Agents typically operate at the leaf level, but if no valid path is found or the local environment proves infeasible, replanning is escalated one level up in the hierarchy. This corresponds to replanning across all sibling sub-environments together, expanding the search area and increasing the likelihood of finding valid paths in challenging scenarios.

Additionally, the hierarchical decomposition incorporates a retraining condition to determine whether retraining is necessary after one or more changes occur within a sub-environment. The key idea is that certain obstacle changes have a greater impact than others, and this impact is not necessarily proportional to the number of changes. The retraining condition evaluates the effectiveness of the current policy in the affected sub-environment based on its success rate. If the policy’s effectiveness drops beyond a defined threshold, the agent retrains to adapt to the changes and to restore performance. This approach enhances efficiency by ensuring that retraining is only performed when necessary.

We implement four approaches within the hierarchical decomposition framework. The first is a naive variant that replans only at the leaf level, ignoring higher levels of the hierarchy, closely mirroring the method of Yarahmadi et al. The second approach actively exploits the hierarchical structure, enabling replanning at multiple levels when necessary. Both methods use single-agent tabular Q-learning for simplicity reasons. The third and fourth approaches introduce federated Q-learning, a \ac{MARL} variant that allows agents to learn in parallel and share knowledge through a federated averaging process. These two methods differ in the mechanism used to combine Q-tables. By enabling parallel learning and information sharing, the federated approaches improve training efficiency and accelerate the path planning process.

The experimental setup is also expanded to include a more diverse set of environments. We categorize mazes by difficulty: easy, medium, and hard, based on the density of obstacles, charging stations, and free space. Harder mazes contain fewer navigable paths and goal locations (charging stations), providing a more challenging testbed. This allows for a more rigorous and meaningful evaluation of each approach. We compare our four \ac{RL}-based methods against two A*-based baselines, which serve as theoretically optimal references. Although effective, these baselines rely on assumptions that make them impractical in real-world settings, such as complete knowledge of the dynamic environment at all times during the simulation.

In this report, we show that efficient and accurate path planning is achievable under the region-aware dynamicity assumption. The proposed federated Q-learning methods achieve success rates comparable to A*, while significantly reducing adaptation time in dynamic environments. Although the resulting paths are slightly longer than those generated by A*, this trade-off is balanced by improved adaptability and scalability.

\subsection{Technical Report Structure}

The remainder of this report is organized as follows. Section~\ref{sec:background} introduces the foundational concepts necessary for understanding the reviewed literature and proposed methodology. Section~\ref{sec:literature-review} presents an overview of related work in dynamic path planning and \ac{MARL}, including the approach by Yarahmadi et al. Section~\ref{sec:methodology} introduces the hierarchical decomposition framework, the retraining condition, and the four developed \ac{RL}-based approaches. Section~\ref{sec:evaluation} describes the experimental setup and presents the results. Section~\ref{sec:discussion} offers a high-level analysis of the findings and relates these to the addressed limitations. Finally, Section~\ref{sec:conclusion-future-work} concludes the report and outlines promising directions for future research.

Two appendices are included. Appendix~\ref{sec:policy-visualization} illustrates the visualization tool with an example, and Appendix~\ref{sec:implementation-running-details} provides implementation details and instructions for replicating the experiments.

\clearpage

\section{Background} \label{sec:background}

This section provides the background necessary to understand the reviewed literature in Section~\ref{sec:literature-review} and the methodology presented in Section~\ref{sec:methodology}. It begins with an introduction to \ac{RL} in Section~\ref{sec:reinforcement-learning}, highlighting its two most important types. Section~\ref{sec:multi-agent-systems} then introduces \acp{MAS}, followed by a discussion of \ac{MARL} in Section~\ref{sec:marl}. Section~\ref{sec:mdp} presents a detailed explanation of \acp{MDP}, while Section~\ref{sec:bellman} outlines a foundational solution based on the Bellman equations. Finally, Section~\ref{sec:q-learning} provides an in-depth discussion of the Q-learning algorithm, which is the foundational learning algorithm used in this report.

\subsection{Reinforcement Learning} \label{sec:reinforcement-learning}

\ac{RL} is a machine learning technique that trains agents to take actions to maximize cumulative rewards. It operates via a trial-and-error learning process that resembles human learning: actions that bring the agent closer to its goal are reinforced via positive rewards, while counterproductive actions are discouraged through negative rewards or penalties. Agents interact with an environment: they select actions based on the current state, receive a reward signal, and transition to a new state. Through repeated interactions, agents learn policies that map states to actions to maximize long-term rewards.

\ac{RL} methods can be broadly categorized into two families. Tabular methods explicitly store value estimates for each state-action pair and are tractable in small, discrete environments. These methods enjoy theoretical convergence guarantees to optimal or near-optimal policies. Function-approximation methods, particularly those based on \acp{DNN}, replace lookup tables with learned representations, allowing agents to handle high-dimensional or continuous state spaces. Although more computationally intensive, \ac{Deep RL} has demonstrated impressive results in complex decision-making tasks.

This report focuses on tabular methods for their conceptual simplicity and low hardware requirements. Unlike \ac{Deep RL} methods, which typically require \acp{GPU} and longer training times, tabular approaches can be run efficiently on standard \acp{CPU} with sufficient memory.

\subsection{Multi-Agent Systems} \label{sec:multi-agent-systems}

A \ac{MAS}~\cite{van2008multi} is composed of multiple intelligent agents that interact with the environment to achieve individual or shared goals. Agents may act independently, collaborate, or even compete, depending on the design of the system and the problem being solved.

\acp{MAS} are particularly suitable for addressing problems that are difficult or inefficient for a single agent to solve, especially in large, distributed, or dynamic environments. By dividing the environment or workload among multiple agents, \acp{MAS} offer improved scalability, fault tolerance, and computational efficiency.

\acp{MAS} are a key enabler of \ac{DAI}, where agents can operate in parallel and optionally share information. This is especially advantageous in path planning tasks, where the environment may be too large for a single agent to explore efficiently. The use of multiple agents also enhances robustness: failures or suboptimal decisions by one agent can be compensated for by others.

In summary, combining \acp{MAS} with \ac{RL} enables distributed, scalable, and adaptive path planning strategies that are well-suited for complex and dynamic environments.

\subsection{Multi-Agent Reinforcement Learning} \label{sec:marl}

\ac{MARL} extends \ac{RL} to \acp{MAS}, enabling multiple agents to learn and act within a shared environment. It has attracted growing attention in recent years, with successful applications in domains such as autonomous driving, swarm robotics, and distributed decision-making.

Depending on how coordination is organized and whether agents share observations, different learning paradigms can be distinguished. The definitions below follow the taxonomy presented in a recent survey on multi-agent decision-making systems~\cite{jin2025comprehensive}.

\begin{itemize}
    \item \textbf{\ac{CTCE}:} A central controller governs all agents by aggregating their observations, actions, and rewards to make joint decisions. This means that agents interact with the environment in a coordinated manner. While this paradigm enables high levels of coordination, its scalability is limited in large-scale systems.
    \item \textbf{\ac{DTDE}:} Each agent learns and acts independently, updating its policy based solely on local observations and rewards, without having access to observations from other agents. This paradigm excels in scalability and robustness, especially in scenarios with limited communication.
    \item \textbf{\ac{CTDE}:} A central controller aggregates information from all agents to optimize their policies during training, but each agent operates independently during execution without considering the actions of other agents. This paradigm is widely used in the \ac{MARL} literature and addresses key challenges such as non-stationarity and scalability.
\end{itemize}

These paradigms provide the foundation for comparing existing \ac{MARL} approaches for path planning in Section~\ref{sec:literature-review} and for situating our proposed methodology within the broader MARL landscape.

\subsection{Markov Decision Processes} \label{sec:mdp}

\ac{RL} methods interact with environments, typically modeled as \acp{MDP}. An \ac{MDP} is formally defined as a tuple $(S, A, R_a, P_a, \gamma)$:

\begin{itemize}
    \item $S$: the set of all possible states (state space). In this report, the state space is discrete.
    \item $A$: the set of all possible actions (action space). This is also discrete in our case.
    \item $R_a(s, s')$: the immediate reward received after transitioning from state $s$ to $s'$ by taking action $a$.
    \item $P_a(s, s')$: the probability of transitioning from state $s$ to $s'$ via action $a$. In this report, transitions are deterministic, i.e., $P_a(s, s') \in \{0,1\}$.
    \item $\gamma \in [0,1)$: the discount factor that favors immediate rewards over long-term ones.
\end{itemize}

A policy $\pi$ maps states to actions. The goal of \ac{RL} methods is to find an optimal policy $\pi^*$ that maximizes the expected cumulative reward over time.

An important property of \acp{MDP} is the Markov property, which asserts that the future is conditionally independent of the past given the present. That is, the next state depends only on the current state and action, not on the sequence of past states and actions. This property simplifies decision-making and is fundamental to most \ac{RL} algorithms.

\begin{figure}[h]
    \centering
    \includegraphics[width=0.8\linewidth]{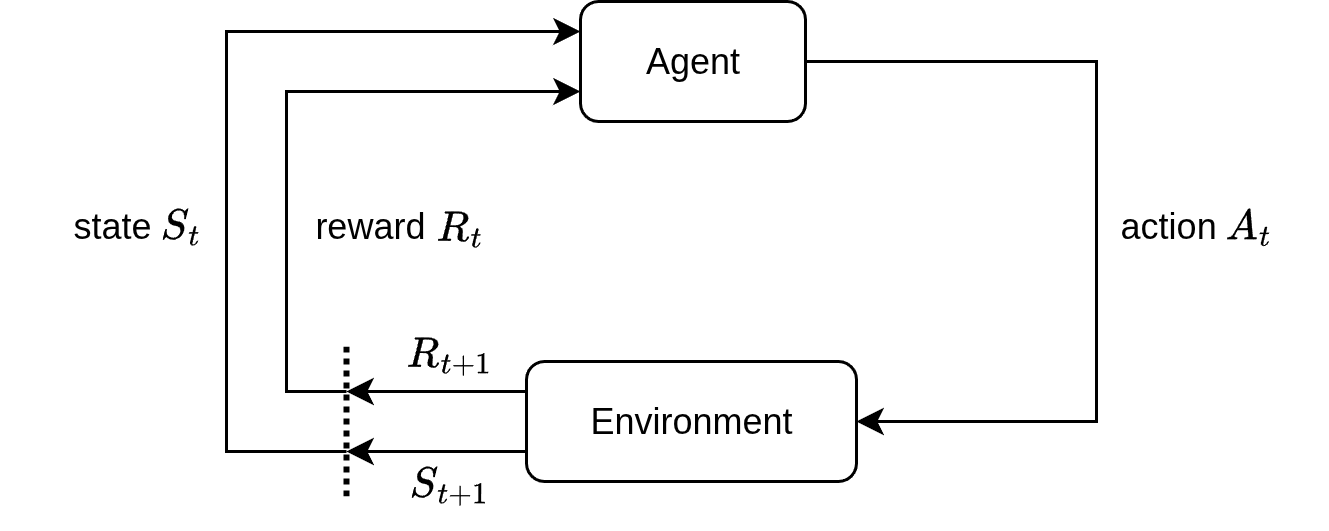}
    \caption{Agent-environment interaction loop: the agent observes state $S_t$, performs action $A_t$, receives reward $R_{t+1}$ and transitions to state $S_{t+1}$.}
    \label{fig:RL-loop}
\end{figure}

The agent-environment interaction loop is depicted in Figure~\ref{fig:RL-loop}. The environment is treated as a black box, meaning the agent does not require prior knowledge of the environment’s internal transition dynamics or its underlying reward function. This characteristic makes \ac{RL} particularly suitable for scenarios where the environment is unknown or subject to changes over time, rendering it uncertain. Through interaction with the environment, the agent gathers experience in the form of tuples $(S_t, A_t, R_{t+1}, S_{t+1})$, which are used to refine its estimate of the optimal policy.

\vspace{10pt}

\subsection{The Bellman Equations} \label{sec:bellman}

The Bellman equations form the mathematical foundation for solving \acp{MDP}. They define the value of a state or a state-action pair under a given policy.

The state-value function $V_\pi(s)$ gives the expected return from starting in state $s$ at time $t$ and following policy $\pi$:
\begin{equation} \label{eq:value}
    V_\pi(s) = \mathbb{E}_\pi \left[ \sum_{k=0}^\infty \gamma^k R_{t+k+1} \biggm| S_t = s \right]
\end{equation}

The action-value function $Q_\pi(s,a)$ gives the expected return from taking action $a$ in state $s$ at time $t$ and thereafter following policy $\pi$:
\begin{equation} \label{eq:action}
    Q_\pi(s,a) = \mathbb{E}_\pi \left[ \sum_{k=0}^\infty \gamma^k R_{t+k+1} \biggm| S_t = s, A_t = a \right]
\end{equation}

When the environment’s transition dynamics and reward function are known, these value functions can be computed exactly, and dynamic programming techniques, such as value iteration and policy iteration~\cite{pashenkova1996value}, can be used to find optimal policies. These methods rely on the recursive forms of the Bellman equations:
\begin{align}
    V_\pi(s) &= \sum_a \pi(a \mid s) \sum_{s',r} P_a(s, s') \left[ r + \gamma V_\pi(s') \right] \label{eq:value-dynamics} \\
    Q_\pi(s,a) &= \sum_{s',r} P_a(s, s') \left[ r + \gamma \max_{a'} Q_\pi(s', a') \right] \label{eq:action-dynamics}
\end{align}

In dynamic environments, both the transition dynamics and the underlying reward function change over time, causing dynamic programming methods to become ineffective. Furthermore, solving \acp{MDP} with large state spaces using the Bellman equations is computationally intractable, as it requires evaluating all states in the state space. These limitations necessitate the use of model-free \ac{RL} algorithms, such as Q-learning.

\subsection{Q-Learning} \label{sec:q-learning}

Q-learning~\cite{watkins1992q} is a model-free \ac{RL} algorithm that learns the optimal action-value function $Q_*(s, a)$ through direct interaction with the environment. It does not require knowledge of its transition probabilities or underlying reward function, making it particularly suitable for dynamic environments where the reward function changes over time due to moving obstacles.

Q-learning is based on \ac{TD} learning, enabling the agent to update its knowledge after each interaction step, rather than waiting for an entire episode to complete, as is required in Monte Carlo-based methods. The update rule is defined as follows:

\begin{equation} \label{eq:q-learning}
Q(S_t,A_t) \leftarrow Q(S_t,A_t) + \alpha\left[ R_{t+1} + \gamma \max_a Q(S_{t+1},a) - Q(S_t,A_t) \right]
\end{equation}

Where:
\begin{itemize}
    \item $\alpha \in (0,1]$ is the learning rate, which controls how much new information overrides old estimates.
    \item $\gamma \in [0,1)$ is the discount factor.
    \item $\displaystyle\max_a Q(S_{t+1}, a)$ estimates the value of the best possible action in the next state.
\end{itemize}

Higher learning rates can accelerate learning but may cause instability. Conversely, lower values improve stability but slow down convergence.

\subsection{Convergence} \label{sec:convergence}

Q-learning typically requires numerous interactions with the environment to accurately estimate state-action values. In this report, convergence is assessed by comparing Q-tables across iterations. For each table entry, the absolute difference between successive iterations is calculated. If the maximum difference falls below a predefined threshold for a specified number of iterations, the algorithm is deemed to have converged. This convergence criterion is employed by the \texttt{onlyTrainLeafNodes} and \texttt{singleAgent} approaches, as described in Section~\ref{sec:methodology}. However, the two federated Q-learning approaches do not utilize this criterion.

\clearpage

\section{Literature Review} \label{sec:literature-review}

This literature review examines existing path-planning approaches for dynamic environments, with a focus on \ac{MARL}. The reviewed works are organized into two categories based on their applicability to dynamic settings. The first category includes approaches that may be suitable for dynamic environments but lack explicit testing in such contexts. These works are discussed in Section~\ref{sec:non-dynamic-tested}. The second category comprises approaches explicitly designed and tested for dynamic environments. These are examined in Section~\ref{sec:dynamic-tested}, which is further divided into subsections based on shared methodologies or concepts.

The work by Yarahmadi et al.~\cite{yarahmadi2024comp} is analyzed separately in Section~\ref{sec:yarahmadi}, where we evaluate its methodology and identify its limitations. This work forms the foundation for this technical report. The section concludes with a summary table (Table~\ref{tab:summary}), which compares the reviewed works based on a set of key criteria.

\subsection{Approaches Not Tested in Dynamic Environments} \label{sec:non-dynamic-tested}

The earliest work reviewed in this category was published in 2018 by Panov et al.~\cite{panov2018grid}, who developed a \ac{Deep RL}-based approach for grid path planning. Specifically, they employed a \ac{DQN} that receives the agent’s locally visible area as input and outputs a probability distribution over all possible actions. While effective in single-agent, static environments, this approach does not align with our focus on \acp{MAS} and dynamic environments.

Lin et al.~\cite{lin2019end} proposed a novel \ac{Deep RL} framework enabling teams of robots to navigate unknown and complex environments. Their method follows the \ac{CTDE} paradigm, learning decentralized policies executed independently by each agent. The approach maps raw LiDAR measurements directly to velocity control commands without constructing obstacle maps. While innovative, it was not explicitly designed or evaluated for dynamic environments.

Gan et al.~\cite{gan2019new} introduced a \ac{MARL} method called \ac{E-DCM-MULTI-Q} that updates an agent’s action-value estimates using both its own state and the state values of its neighboring agents. Neighbors are defined as agents within the local neighborhood of the current agent. The goal is to reduce the number of learning trials needed. The method employs an evolutionary algorithm to optimize multiple hyperparameters efficiently. However, it was not tested in dynamic environments.

Luis et al.~\cite{luis2021multiagent} presented a \ac{MADRL}-based approach for efficient path planning of \acp{ASV} to monitor water quality in the Ypacara{\'\i} lake. The objective was to develop a scalable \ac{MAS} that minimizes the average time since each zone in the lake was last visited. The authors employed Double Deep Q-Learning and Dueling Deep Q-Learning to estimate state-action values. The method does not consider dynamic obstacles as part of the environment.

Li et al.~\cite{li2023ace} proposed ACE, a method that converts a \ac{M-MDP} into a \ac{SE-MDP}. This transformation offers two main benefits: the resulting \ac{SE-MDP} can be solved using standard single-agent \ac{RL} methods, which are theoretically guaranteed to converge, and it mitigates the non-stationarity problem in \acp{MAS}, where learning becomes unstable due to other agents’ evolving policies. Despite these strengths, the method is mathematically complex and was not tested in dynamic environments with moving obstacles.

Shen et al.~\cite{shen2024autonomous} developed \ac{CL-MADDPG-PP}, which combines \ac{MADDPG} with proportional prioritization in an experience replay buffer. Following the \ac{CTDE} paradigm, the method trains agents with shared experiences but allows independent execution. While evaluated in a warehouse setting, the study does not provide details on the presence of moving obstacles, limiting its relevance to fully dynamic environments.

Yin et al.\cite{yin2024cooperative} proposed asyn-MARL, an asynchronous \ac{MARL} framework for solving the \ac{MSD-SPP} in dynamic road networks. The method employs two-stage route planning: inter-region planning using an actor–critic network, and intra-region planning using a \ac{Deep RL} model to find the shortest path within a region to a designated boundary edge. The environment is divided into sub-graphs, an idea similar to the regional decomposition used by Yarahmadi et al.\cite{yarahmadi2024comp} and in our own methodology. While traffic congestion is modeled as a dynamic factor, the method does not account for changing network topologies, such as newly added, blocked, or removed roads.

Wang et al.~\cite{wang2025lns2+} introduced LNS2+RL, a \ac{MAPF} algorithm that integrates \ac{MARL} with \ac{LNS2}. In early iterations, \ac{MARL} is used for low-level replanning, while later iterations adaptively switch to prioritized planning to efficiently resolve remaining collisions. The authors define the \ac{MAPF} problem on a 2D four-neighbor grid with a finite set of static obstacles, excluding dynamic environments from consideration.

Overall, these works contribute valuable insights to \ac{MARL} and path planning, but their lack of focus on truly dynamic environments makes them less relevant regarding the interests of this report. 

\subsection{Approaches Tested in Dynamic Environments} \label{sec:dynamic-tested}

\subsubsection{Perception‑Driven Approaches}

Chang et al.~\cite{chang2021reinforcement} proposed a mobile robot path planning approach that enhanced the \ac{DWA} for local path optimization. No global path planning is used since the assumption was made that obtaining a global map is difficult and not always feasible. The robot is equipped with LiDAR, providing the distance to the nearest obstacle in each direction of a circle at each time step. The method focuses on identifying optimal motion parameters, such as velocity and steering angle, to guide the agent’s movement toward its destination. Their approach employs neither a \ac{MAS} nor \ac{MARL}.

This particular action selection highlights an important aspect of path planning: the type of action space considered. Path planning algorithms may operate on high-level actions (e.g., moving forward, backward, or in cardinal directions) or on low-level actions, such as adjusting acceleration or rotation to avoid collisions. For this technical report, we restrict ourselves to high-level, discrete actions based on cardinal directions. This simplification reduces the complexity of the path planning problem, facilitating clearer evaluation and implementation.

Bae et al.~\cite{bae2019multi} proposed a method that combines \acp{CNN} with \ac{DQL}. In their approach, the \ac{CNN} analyzes the exact situation using image information of its environment, and the robot navigates the environment based on the situation analyzed through \ac{DQL}. The \ac{CNN} learns features from the input images passed to the \ac{DQL} algorithm, which consists of a Q-network and a target Q-network. While this method supports both static and dynamic environments due to its direct environmental representation, its reliance on image-based inputs makes it less practical for real-world applications where such precise representations may be unavailable. In our implementation, we assume that image-based inputs are inaccessible.

Liu et al.~\cite{liu2020mapper} introduced the \ac{MAPPER} method, which also uses an image-based representation to model dynamic obstacle behavior. The approach is decentralized, meaning agents learn optimal actions on their own using local observations without using any form of communication. It employs evolutionary \ac{RL} in a decentralized setting to train agents. \ac{A2C} is used to update the model parameters of the actor-critic network. The authors acknowledge that collisions with dynamic obstacles in complex environments may still occur.

Guan et al.~\cite{guan2022ab} proposed \ac{AB-Mapper}, which builds upon \ac{MAPPER} and introduces agent communication via a BicNet (bidirectional LSTM). It operates within an actor-critic framework and is designed for dynamic environments. It also operates an attention mechanism in the critic network that allows agents to focus on the actions and states of the most relevant neighboring agents. The approach aligns with the \ac{CTDE} paradigm, allowing decentralized execution. Its reliance on \acp{DNN} makes it computationally intensive, however.

Gao et al.~\cite{gao2025scalable} introduced a multi-\ac{USV} path planning algorithm built on the \ac{MADDPG} framework. Their main contribution is a novel observation space construction strategy based on adjacent observation, which extracts key information from nearby \acp{USV} and obstacles. For each \ac{USV}, the local observation includes only the essential details of the nearest \ac{USV} and obstacle, along with its own state and target information. They verified the performance of the algorithm in an environment with three dynamic obstacles having the same structure as other \acp{USV}, which move randomly in the environment. However, restricting the testbed to only three dynamic obstacles does not resemble a real dynamic environment.

These studies demonstrate that agents or robots with access to a local perception of their environment can effectively navigate dynamic settings.

\subsubsection{Hierarchical Path Planning Frameworks}

Several works focus on hierarchical approaches for tackling the path-planning problem in dynamic environments.

Choi et al.~\cite{choi2021reinforcement} presented an improvement of the \ac{MCAL} algorithm, called \ac{MCAL-P}. The original \ac{MCAL} algorithm is a purely \ac{RL}-based obstacle avoidance method that uses \ac{SAC}. The algorithm is decentralized, and no communication between agents takes place. The method suffers from several issues, which is why \ac{MCAL-P} was proposed. This new algorithm integrates \ac{MCAL} with a global path planner, such as A*. The global path is used as guidance to where the robot should move, while still being able to avoid dynamic obstacles based on multiple steps of LiDAR information collected locally to predict the obstacle trajectories.

Wang et al.~\cite{wang2020mobile} presented the \ac{G2RL} algorithm, which integrates a global planner with a local \ac{RL}-based module to enable end-to-end learning in dynamic environments. The A* algorithm serves as the global planner, assuming the environment is well-known. The local \ac{RL} planner takes as input a local \ac{FOV} of the robot, a local segment of global guidance, and a history of local observations that together form a tensor. A series of three-dimensional \ac{CNN} layers extracts features and reshapes the features into one-dimensional vectors. A \ac{LSTM} layer further extracts temporal information by aggregating the vectors. Fully connected layers estimate the quality of each state-action pair and choose the action that maximizes the corresponding value. In large-scale environments, the global planner exhibits poor scalability, leading to a substantial decline in performance.

Chang et al.~\cite{chang2023hierarchical} introduced a hierarchical framework for multi-robot navigation and formation maintenance in unknown environments containing both static and dynamic obstacles. The framework consists of three modules. The navigation module utilizes a \ac{DDPG} network to train each robot individually to navigate toward the global target. The formation module determines the optimal formation configuration for the robots in real time. The velocity adjustment module adjusts the navigation velocity of each robot toward its formation target. While effective, it operates in a continuous action space, which increases the mathematical and computational complexity compared to discrete approaches like cardinal direction movement, used in this report.

Zhou et al.~\cite{zhou2024novel} introduced \ac{TD-MATD3} for multi-\ac{UAV} autonomous path planning in complex dynamic environments. Based on MATD3, a combination of TD3 and \ac{MADDPG}, it handles both static and dynamic obstacles in mixed environments. The approach aligns with the \ac{CTDE} paradigm and is rather complex. It involves advanced mathematical modeling and \ac{Deep RL} expertise.

These studies show that decomposing the path planning task into distinct layers, either through global and local planners or through specialized modules with dedicated functions, can significantly improve efficiency in dynamic environments. This principle of decomposition also forms the foundation of our own methodology.

\subsubsection{Optimization‑based Path Planning}

Tordesillas et al.~\cite{tordesillas2021mader} proposed MADER, a trajectory planning framework for \acp{UAV} operating in both static and dynamic multi-agent environments. The method leverages the MINVO framework to model other agents' trajectories and formulates path planning as an optimization problem, without relying on \ac{RL}. It employs decentralized planning, where each agent determines its own trajectory without a central coordinator, and ensures collision avoidance by incorporating the predicted trajectories of other agents as constraints in the optimization process. Optimization-based path planning typically relies on robot kinematics models defined by systems of differential equations, which fall outside the scope of this technical report. 

\subsubsection{Fully Decentralized MARL}

Guo et al.~\cite{guo2024decentralized} proposed the \ac{DPPDRL}, a decentralized approach for coordinating \acp{AGV} in dynamic \ac{RMFS} environments. Following the \ac{DTDE} paradigm, the method enables agents to make real-time decisions based solely on local observations. A carefully designed reward function and state space allow the system to remain robust under dynamic conditions, with each agent learning and acting independently. 

Other fully decentralized \ac{MARL} approaches discussed in this literature review include those by Liu et al.~\cite{liu2020mapper} and Tordesillas et al.~\cite{tordesillas2021mader}.

\subsubsection{Survey and Review Papers}

Cai et al.~\cite{cai2020mobile} presented a comprehensive survey of path planning methods in dynamic environments, with a particular focus on hierarchical path planning, which separates planning into global and local layers. Examples of \ac{RL}-based methods include IRL, CADRL, SA-CADRL, and CRL. The hierarchical structure, where local planners operate within sub-regions of a larger environment, is a key inspiration for the methodology proposed by Yarahmadi et al.~\cite{yarahmadi2024comp}.

Madridano et al.~\cite{madridano2021trajectory} provided a literature review comparing different categories of multi-robot path planning methods, categorized as decomposition graph-based, sampling-based, model-based, and bio-inspired methods. Each category is characterized by certain advantages and disadvantages, which are presented in a table. A few of these approaches are tailored to dynamic environments. Flexible Unit A* (FU-A*)~\cite{boroujeni2017flexible} is a graph-based method that incorporates the prediction of dynamic obstacle positions. Several \ac{RL}-based methods capable of handling dynamic conditions are also noted, including those by~\cite{bae2019multi, qie2019joint, cruz2017path}, all of which utilize \acp{DNN}.

Almazrouei et al.~\cite{almazrouei2023dynamic} reviewed \ac{RL}-based techniques for \ac{DOA} and highlighted several promising directions. Their survey reinforces the strength of \ac{Deep RL} for dynamic navigation tasks, while also identifying unresolved challenges, such as creating appropriate reward functions in dynamic environments and evaluating the performance of RL algorithms in real-world environments.

\subsection{Region-Aware MARL Methodology of Yarahmadi et al.} \label{sec:yarahmadi}

In this section, we examine the work by Yarahmadi et al.~\cite{yarahmadi2024comp} on the path planning problem in dynamic environments. This section outlines the problem statement and motivation behind their work, the methodology that distinguishes it from other approaches, and their experimental findings. Finally, we identify several limitations in their work that serve as a foundation for the contributions of this report.

\subsubsection{Problem Statement and Motivation}

Yarahmadi et al.~\cite{yarahmadi2024comp} identify a critical limitation in classical path planning algorithms such as A*, which, while effective in static environments, become inefficient in dynamic settings where obstacles change over time. These algorithms lack incremental update mechanisms and instead require full replanning whenever the environment changes, leading to substantial computational overhead.

To address this inefficiency, the authors introduce a key assumption: although the environment is dynamic, the locations of changes can be detected and localized with reasonable precision. Based on this assumption, they propose a region-aware replanning strategy that confines path updates to only the affected parts of the environment, avoiding unnecessary global recomputation. This perspective is rarely considered in the \ac{MARL} literature. 

This assumption aligns well with many real-world scenarios where environmental changes are localized and their impact is confined to specific regions. For example, in warehouse settings, a blocked aisle or a stalled robot typically affects only a limited area. Similarly, in urban traffic systems, disruptions such as accidents primarily influence specific intersections. Even in extreme cases, such as targeted strikes, damage is often restricted to discrete zones. In these scenarios, a region-aware replanning mechanism enhances scalability and responsiveness by focusing updates solely on the affected regions.

\subsubsection{Methodology}

\begin{figure}[h]
    \centering
    \includegraphics[width=0.8\linewidth]{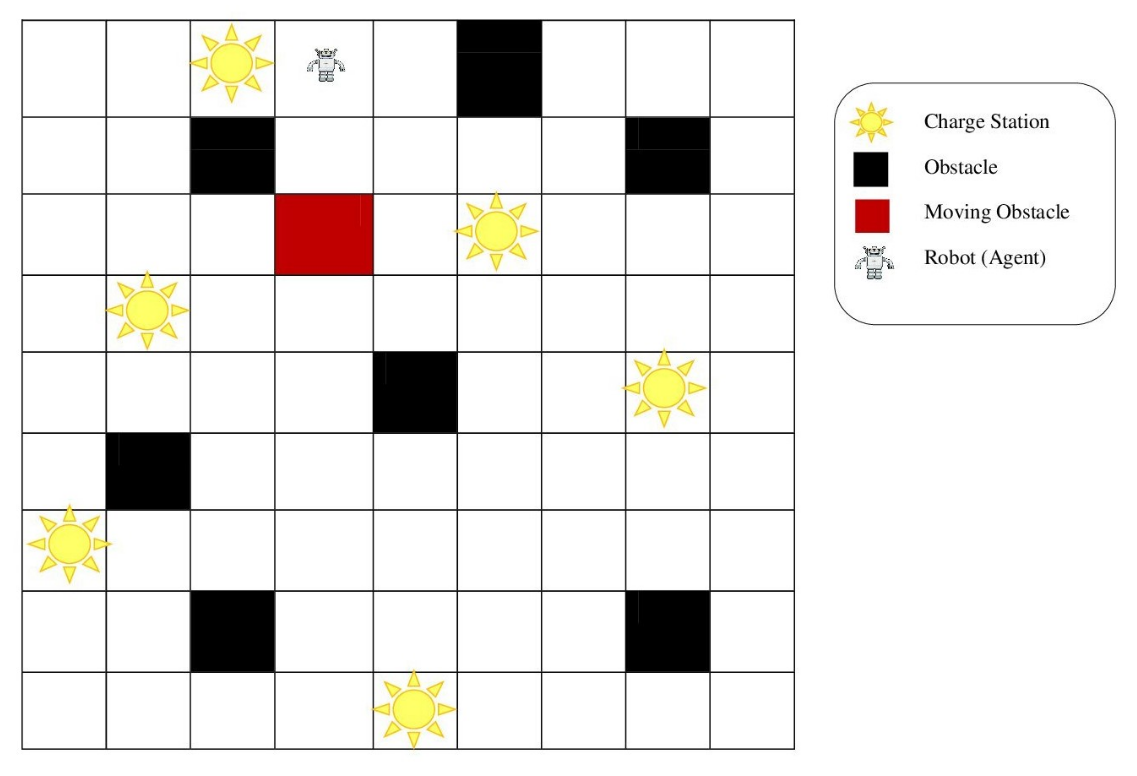}
    \caption{a $9 \times 9$ maze consisting of free space positions (white), static obstacles (black), dynamic obstacles (red), charging stations (yellow suns), and a robot.}
    \label{fig:maze}
\end{figure}

The environments considered in their work are mazes represented as two-dimensional grids, where each cell is either free space, an obstacle, a charging station, or a robot. An example maze of size $9 \times 9$ is depicted in Figure~\ref{fig:maze}. The robot is tasked with navigating to a charging station while avoiding obstacles, which may be either static or dynamic. Dynamic obstacles are allowed to move to adjacent cells at each time step.

\begin{figure}[h]
    \centering
    \includegraphics[width=0.8\linewidth]{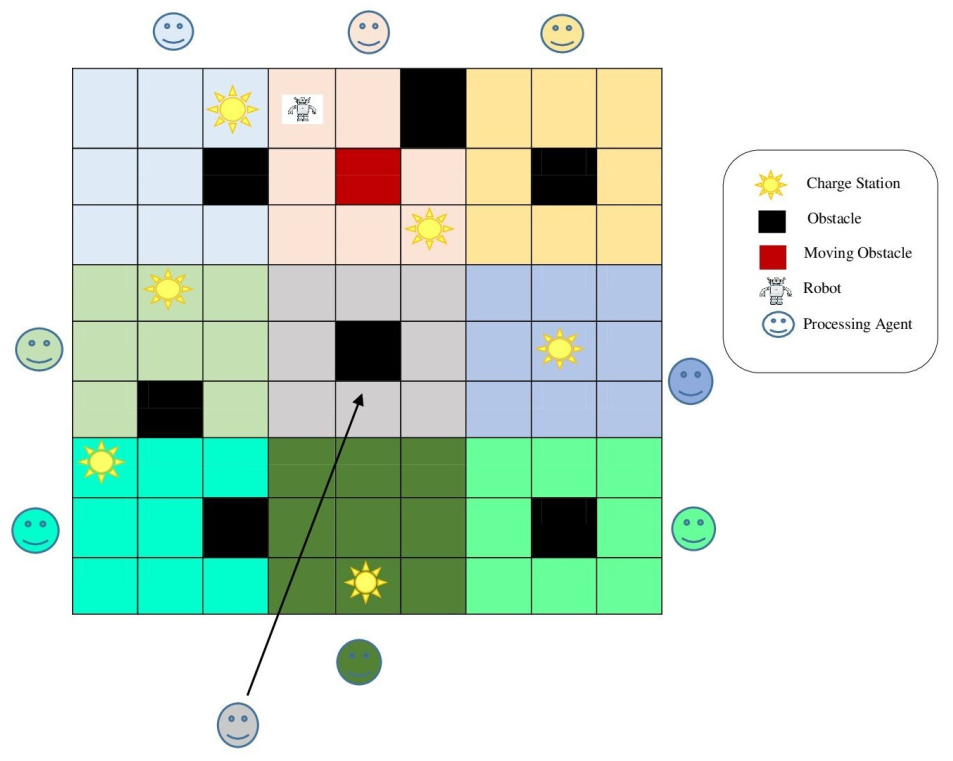}
    \caption{a $9 \times 9$ maze divided into $3 \times 3$ sub-mazes. Each sub-maze is assigned to a dedicated processing agent.}
    \label{fig:sub-mazes}
\end{figure}

To implement the region-aware dynamicity assumption, the maze is partitioned into equally sized sub-environments (e.g., $3 \times 3$ regions), each managed by a dedicated processing agent. Figure~\ref{fig:sub-mazes} illustrates this decomposition into sub-environments. Each agent is responsible for finding paths in its assigned sub-environment using Q-learning when it is affected by changes.

Environmental changes are detected by comparing the maze state at consecutive time steps, such as $t$ and $t+1$. If no changes occur in the environment, the global path planner (e.g., A*) is used to compute paths to a charging station. When changes are detected, the agents in the affected sub-environments replan their paths by retraining using the Q-learning algorithm.

The MasMat algorithm formalizes this procedure by taking the current and previous environment states as input and producing a path to a charging station. It determines which sub-environments have changed and instructs the right agents to use Q-learning to replan a path to a charging station. Notably, agents operate independently without coordination or information sharing, ensuring a highly decentralized and scalable approach.

\subsubsection{Evaluation and Key Findings}

The proposed method is evaluated in both static and dynamic environments. In static settings, A* outperforms the \ac{MARL}-based method in terms of path quality and computation time. This is expected, as Q-learning requires significant exploration and training to converge.

In dynamic environments, the limitations of A* become evident. Without access to a complete environmental representation at each simulation time step, A* cannot perform replanning, rendering it unsuitable for dynamic settings. To evaluate A*’s potential performance in such contexts, we assume it has access to a complete environmental representation at each time step, requiring replanning across the entire environment whenever a change is detected. This approach becomes computationally infeasible as the environment scales. In contrast, the \ac{MARL}-based solution replans only within affected regions, significantly enhancing efficiency.

The results show that while A*'s replanning time grows exponentially with environment size, the \ac{MARL}-based method maintains manageable performance across different scales. Therefore, A* is preferable for static environments with a complete environmental representation, while the proposed method offers a scalable and adaptive solution for dynamic scenarios.

\subsubsection{Limitations}

Despite its promising results, the approach by Yarahmadi et al. presents several limitations, previously noted in the introduction section, that are directly addressed in this technical report:

\begin{itemize}
    \item \textbf{Simplistic environment design:} Their evaluation uses relatively simple environments, with few obstacles and many charging stations. This simplifies the path planning problem and does not test the robustness of the approach in more challenging settings.
    \item \textbf{Single obstacle changes:} The dynamic component of their simulation involves moving only one obstacle per time step. This does not reflect more chaotic or realistic scenarios where multiple simultaneous changes may occur.
    \item \textbf{Dependence on a global planner:} The global path planner (e.g., A*) is used for finding paths in the environment if no changes occurred. However, in completely unknown environments, the global planner becomes unusable as no complete representation of the environment is available, leaving \ac{RL}-based replanning as the only viable option. Moreover, a global planner does not scale well with the environment's dimensions.
    \item \textbf{Naive replanning strategy:} Each agent initiates replanning whenever a change occurs in its sub-environment, potentially leading to unnecessary replanning when obstacle changes have minimal impact on the path-planning problem.
    \item \textbf{Restricted agent scope:} Each agent is limited to replanning within its assigned sub-environment. If no charging station exists within that region, the agent may be unable to find a valid path, potentially becoming stuck. This limitation stems from the absence of defined relationships between sub-environments, which could enable replanning across larger regions to increase the likelihood of finding a path to a charging station. Consequently, the system does not guarantee the discovery of valid paths under all conditions.

\end{itemize}

\subsection{Summary Table}

Table~\ref{tab:summary} summarizes the studies reviewed in this section, excluding the work by Yarahmadi et al.~\cite{yarahmadi2024comp}, which is analyzed separately below. The table evaluates each approach based on a set of key criteria: use of \ac{RL}, employment of a \ac{MAS}, support for dynamic environments (Dyn. Env.), and the employed learning paradigm (\ac{CTCE}, \ac{DTDE}, or \ac{CTDE}, where applicable). These criteria systematically weigh strengths, such as scalability and robustness enhanced by paradigms like \ac{DTDE}, against weaknesses, such as limited dynamic support or the lack of a \ac{MAS}. These criteria also assess the extent to which other studies align with our goal of developing an efficient, scalable \ac{MARL}-based path planning framework for dynamic environments, building on the work of Yarahmadi et al. Other studies on similar problems can also be evaluated via this set of criteria.

If the work of Yarahmadi et al.~\cite{yarahmadi2024comp} were to be included in this table, it would satisfy the criteria for \ac{RL}, \ac{MAS}, and dynamic environment support. To classify their method under a specific learning paradigm, we examine it more closely: the environment is partitioned into sub-environments, each managed by a single agent that learns a local policy independently and executes it without coordinating with others. This places their approach within the \ac{DTDE} paradigm.

\newpage

\begin{small}
\begin{longtable}[c]{|c|c|c|c|c|c|c|c|c|}
\hline
\textbf{Ref.} & \textbf{Title} & \textbf{Year} & \textbf{RL} & \textbf{MAS} & \textbf{\begin{tabular}[c]{@{}c@{}}Dyn.\\ Env.\end{tabular}} & \textbf{CTCE} & \textbf{DTDE} & \textbf{CTDE} \\ \hline
\endhead
\cite{panov2018grid} & \begin{tabular}[c]{@{}c@{}}Grid path planning with \\ deep reinforcement learning: \\ Preliminary results\end{tabular} & 2018 & x &  &  &  &  &  \\ \hline
\cite{gan2019new} & \begin{tabular}[c]{@{}c@{}}A new multi-agent \\ reinforcement learning \\ method based on evolving \\ dynamic correlation matrix\end{tabular} & 2019 & x & x &  &  & x &  \\ \hline
\cite{bae2019multi} & \begin{tabular}[c]{@{}c@{}}Multi-robot path \\ planning method using \\ reinforcement learning\end{tabular} & 2019 & x & x & x &  &  & x \\ \hline
\cite{wang2019improved} & \begin{tabular}[c]{@{}c@{}}Improved multi-agent \\ reinforcement learning \\ for path planning-based \\ crowd simulation\end{tabular} & 2019 & x & x &  &  & x &  \\ \hline
\cite{lin2019end} & \begin{tabular}[c]{@{}c@{}}End-to-end decentralized \\ multi-robot navigation \\ in unknown complex \\ environments via deep \\ reinforcement learning\end{tabular} & 2019 & x & x &  &  &  & x \\ \hline
\cite{cai2020mobile} & \begin{tabular}[c]{@{}c@{}}Mobile robot path \\ planning in dynamic \\ environments: A survey\end{tabular} & 2020 & x & x & x &  &  &  \\ \hline
\cite{liu2020mapper} & \begin{tabular}[c]{@{}c@{}}Mapper: Multi-agent \\ path planning with \\ evolutionary reinforcement \\ learning in mixed \\ dynamic environments\end{tabular} & 2020 & x & x & x &  & x &  \\ \hline
\cite{wang2020mobile} & \begin{tabular}[c]{@{}c@{}}Mobile robot path planning \\ in dynamic environments \\ through globally guided \\ reinforcement learning\end{tabular} & 2020 & x & x & x &  & x &  \\ \hline
\cite{tordesillas2021mader} & \begin{tabular}[c]{@{}c@{}}MADER: Trajectory \\ planner in multiagent \\ and dynamic environments\end{tabular} & 2021 &  & x & x &  & x &  \\ \hline
\cite{chang2021reinforcement} & \begin{tabular}[c]{@{}c@{}}Reinforcement based \\ mobile robot path \\ planning with improved \\ dynamic window approach \\ in unknown environment\end{tabular} & 2021 & x &  & x &  &  &  \\ \hline
\cite{choi2021reinforcement} & \begin{tabular}[c]{@{}c@{}}Reinforcement learning-\\ based dynamic obstacle \\ avoidance and integration \\ of path planning\end{tabular} & 2021 & x & x & x &  &  & x \\ \hline
\cite{madridano2021trajectory} & \begin{tabular}[c]{@{}c@{}}Trajectory planning \\ for multi-robot systems: \\ Methods and applications\end{tabular} & 2021 & x & x & x &  &  &  \\ \hline
\cite{luis2021multiagent} & \begin{tabular}[c]{@{}c@{}}A multiagent deep \\ reinforcement learning \\ approach for path \\ planning in autonomous \\ surface vehicles: \\ The Ypacarai lake \\ patrolling case\end{tabular} & 2021 & x & x &  & x &  &  \\ \hline
\cite{guan2022ab} & \begin{tabular}[c]{@{}c@{}}Ab-mapper: Attention \\ and bicnet based multi-\\ agent path planning \\ for dynamic environment\end{tabular} & 2022 & x & x & x &  &  & x \\ \hline
\cite{almazrouei2023dynamic} & \begin{tabular}[c]{@{}c@{}}Dynamic obstacle avoidance \\ and path planning through \\ reinforcement learning\end{tabular} & 2023 & x &  & x &  &  &  \\ \hline
\cite{li2023ace} & \begin{tabular}[c]{@{}c@{}}Ace: Cooperative multi- \\ agent q-learning with bi- \\ directional action-dependency\end{tabular} & 2023 & x & x &  &  &  & x \\ \hline
\cite{chang2023hierarchical} & \begin{tabular}[c]{@{}c@{}}Hierarchical multi-robot \\ navigation and formation \\ in unknown environments \\ via deep reinforcement \\ learning and distributed \\ optimization\end{tabular} & 2023 & x & x & x &  &  & x \\ \hline
\cite{guo2024decentralized} & \begin{tabular}[c]{@{}c@{}}A decentralized path \\ planning model based \\ on deep reinforcement \\ learning\end{tabular} & 2024 & x & x & x &  & x &  \\ \hline
\cite{shen2024autonomous} & \begin{tabular}[c]{@{}c@{}}Autonomous navigation \\ with minimal sensors \\ in dynamic warehouse \\ environments: a multi-\\ agent reinforcement \\ learning approach with \\ curriculum learning \\ enhancement\end{tabular} & 2024 & x & x &  &  &  & x \\ \hline
\cite{yin2024cooperative} & \begin{tabular}[c]{@{}c@{}}Cooperative Path Planning \\ with Asynchronous Multi-\\ agent Reinforcement Learning\end{tabular} & 2024 & x & x &  &  &  & x \\ \hline
\cite{zhou2024novel} & \begin{tabular}[c]{@{}c@{}}Novel task decomposed \\ multi-agent twin delayed \\ deep deterministic policy \\ gradient algorithm for\\  multi-UAV autonomous\\  path planning\end{tabular} & 2024 & x & x & x &  &  & x \\ \hline
\cite{wang2025lns2+} & \begin{tabular}[c]{@{}c@{}}LNS2+RL: Combining \\ Multi-Agent Reinforcement \\ Learning with Large \\ Neighborhood Search \\ in Multi-Agent Path Finding\end{tabular} & 2025 & x & x &  &  &  & x \\ \hline
\cite{gao2025scalable} & \begin{tabular}[c]{@{}c@{}}Scalable path planning \\ algorithm for multi-Unmanned \\ Surface Vehicles based on \\ Multi-Agent Deep \\ Deterministic Policy Gradient\end{tabular} & 2025 & x & x & x &  &  & x \\ \hline
\caption{Comparison of the reviewed works (RL = Reinforcement Learning, MAS = Multi-Agent System; Dyn. Env. = Dynamic Environment; CTCE = Centralized Training, Centralized Execution; DTDE = Decentralized Training, Decentralized Execution; CTDE = Centralized Training, Decentralized Execution).}
\label{tab:summary}\\
\end{longtable}
\end{small}

\clearpage

\section{Methodology} \label{sec:methodology}

This section presents the methodology developed to overcome the key limitations identified in the work of Yarahmadi et al.~\cite{yarahmadi2024comp}.

In contrast to their approach, we assume that a global path planner is unavailable due to environmental uncertainty. As a result, all path planning must be performed locally by agents using Q-learning. The most critical limitation in their method is the absence of a fallback mechanism: if an agent cannot find a charging station within its assigned sub-environment, it has no means of exploring alternatives. To address this, we introduce a hierarchical decomposition framework (Section~\ref{sec:hierarchical-decomposition}), which extends the original partitioning approach by recursively subdividing the environment into a tree structure of nested sub-environments.

Within this hierarchy, the root corresponds to the entire environment, while the leaves represent the smallest, most manageable regions. If an agent is unable to find a path in its local region, retraining can be escalated to higher levels of the hierarchy, covering a larger area that may include a reachable charging station. This hierarchical fallback increases the likelihood of finding valid paths.

Another limitation of the original approach is its inefficient replanning strategy, which triggers retraining after every change in a sub-environment, regardless of its impact. To overcome this, we propose a retraining condition (Section~\ref{sec:retraining-condition}) that evaluates the effectiveness of the current policy based on its success rate. If performance degrades beyond a predefined threshold, retraining is triggered; otherwise, the current policy is retained. This mechanism avoids unnecessary computation while maintaining adaptability.

Finally, local retraining in a sub-environment was originally performed using a single-agent Q-learning algorithm, which is effective given the small size of the sub-environments. We adopt this in our methodology (Section~\ref{sec:single-agent-rl}) but further extend it by exploring parallelization through multiple agents. By sharing and aggregating their learned knowledge, training efficiency can be improved. This federated multi-agent extension is described in Section~\ref{sec:marl-methodology}.

\subsection{Hierarchical Decomposition Framework} \label{sec:hierarchical-decomposition}

The hierarchical decomposition framework recursively splits the environment into smaller regions. Specifically, the environment is divided into four parts by bisecting it along both dimensions. This simple and effective strategy supports square and rectangular environmental shapes and scales across different sizes. The splitting process continues recursively until sub-environments reach a maximum size of $20 \times 20$. This size strikes a balance between Q-learning efficiency and the complexity of the paths. 

\begin{figure}[h]
    \centering
    \begin{subfigure}{0.3\textwidth}
        \includegraphics[width=\textwidth]{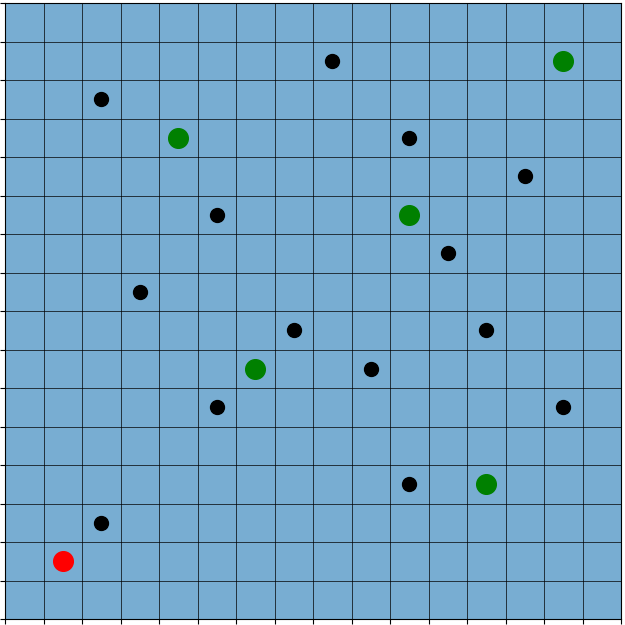}
        \caption{Root level.}
    \end{subfigure}
    \hfill
    \begin{subfigure}{0.3\textwidth}
        \includegraphics[width=\textwidth]{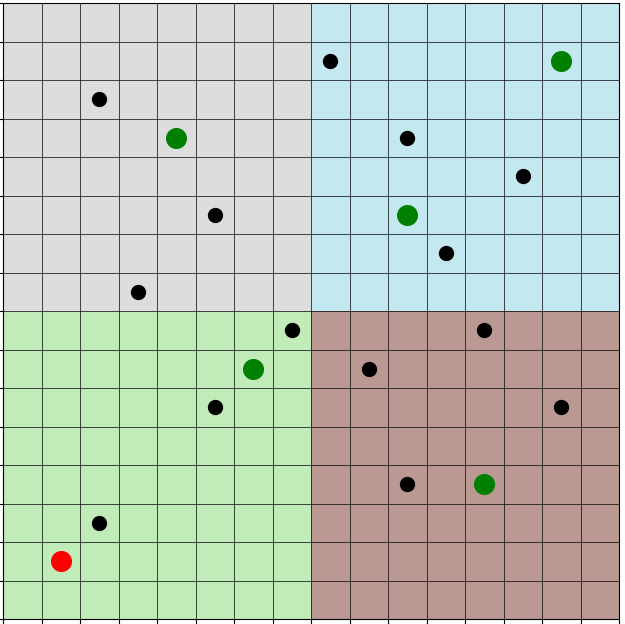}
        \caption{Intermediate level.}
    \end{subfigure}
    \hfill
    \begin{subfigure}{0.3\textwidth}
        \includegraphics[width=\textwidth]{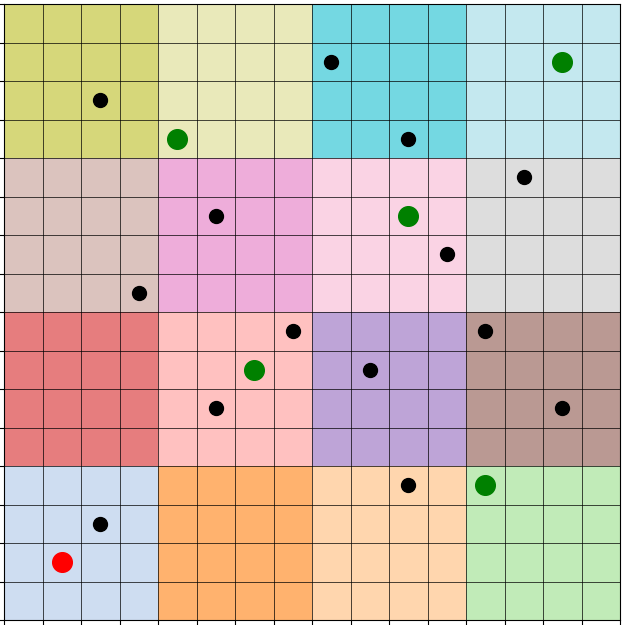}
        \caption{Leaf level.}
    \end{subfigure}
    
    \caption{Hierarchical decomposition until sub-environments of size $4 \times 4$ are reached.}
    \label{fig:hierarchical-decomposition}
\end{figure}

An example decomposition process, which continues splitting until a maximum sub-environment size of $4 \times 4$ is reached, is presented in Figure~\ref{fig:hierarchical-decomposition}.

\begin{figure}[h]
    \centering
    \includegraphics[width=0.85\linewidth]{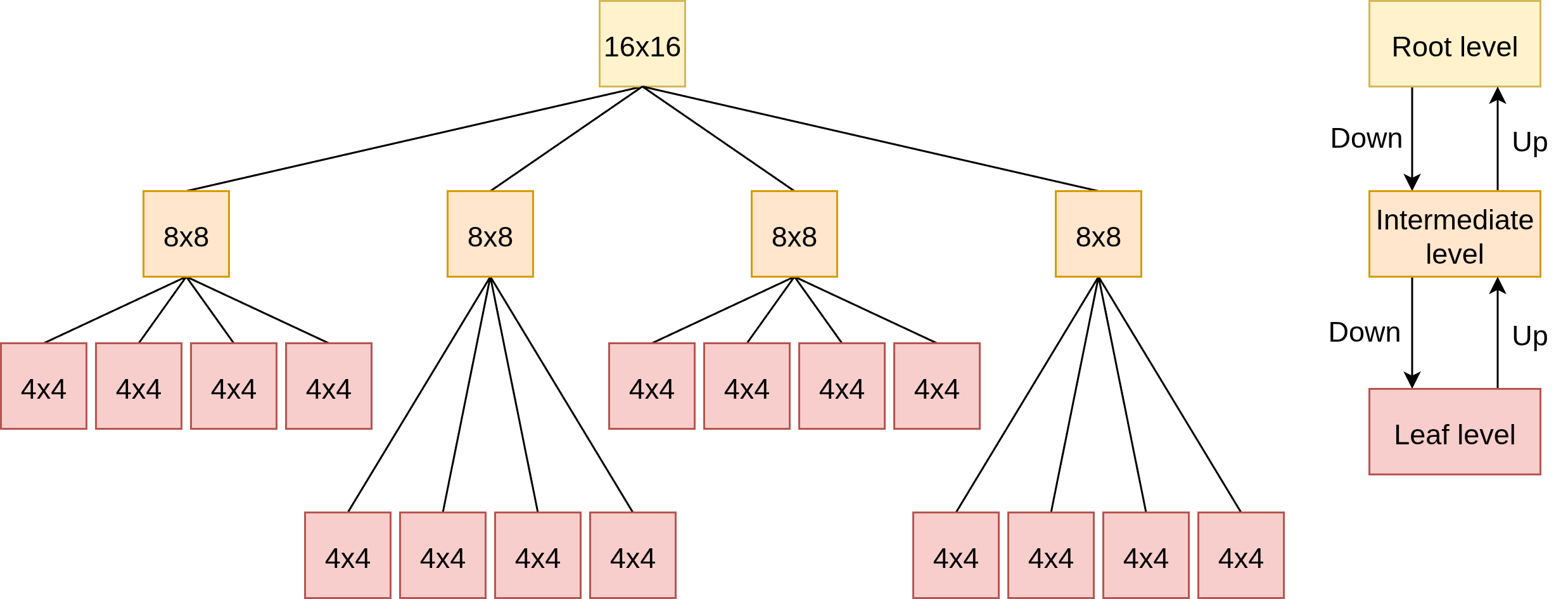}
    \caption{Tree representation of the hierarchical decomposition framework.}
    \label{fig:tree}
\end{figure}

This decomposition naturally forms a tree (shown in Figure~\ref{fig:tree}), consisting of multiple levels:

\begin{itemize}
    \item Root level: the root represents the entire environment.
    \item Intermediate level: nodes represent progressively smaller sub-environments.
    \item Leaf level: leaves represent the smallest sub-environments used for training.
\end{itemize}

Each node in the tree contains a Q-table with the same dimensions as the sub-environment it represents. The hierarchical decomposition mechanism is implemented in the \texttt{createSubEnvironments} function, located in the \texttt{treenode.cpp} file\footnote{\url{https://github.com/micss-lab/MARL4DynaPath/blob/main/src/treenode.cpp}}.

Algorithm~\ref{alg:decomposition} provides the pseudocode for this process, where decomposition starts at the root ($node$) and recursively continues until all sub-environments are no larger than $20 \times 20$, ensuring a balance between Q-learning efficiency and path complexity. The $decompose$ function invokes Algorithm~\ref{alg:decomposition} on the selected child node, reflecting the recursive nature of the algorithm. Each leaf node is assigned a single agent, trained independently via Q-learning in the case of single-agent Q-learning, or multiple agents in the case of federated Q-learning. Agents are restricted to their assigned regions, and parallel training across CPU cores improves scalability and efficiency.

\begin{algorithm}
\caption{Recursive Environment Decomposition}
\label{alg:decomposition}
\begin{algorithmic}[1]
\REQUIRE Current node ($node$)
\ENSURE Tree of sub-environments

\STATE // Base case
\IF{node size ($node.rows,\ node.cols \leq 20$)}
    \RETURN 
\ENDIF

\STATE
\STATE // Define new borders
\STATE $midRow \gets (node.startRow + node.endRow)\ /\ 2$
\STATE $midCol \gets (node.startCol + node.endCol)\ /\ 2$

\STATE
\STATE // Create child nodes
\STATE $child1 \gets$ Node($node.startRow$, $node.startCol$, $mid\text{-}\allowbreak Row$, $midCol$, $node$)
\STATE $child2 \gets$ Node($node.startRow$, $midCol + 1$, $midRow$, $node.endCol$, $node$)
\STATE $child3 \gets$ Node($midRow + 1$, $node.startCol$, $node.end\text{-}\allowbreak Row$, $midCol$, $node$)
\STATE $child4 \gets$ Node($midRow + 1$, $midCol + 1$, $node.endRow$, $node.endCol$, $node$)

\STATE
\STATE // Create tree structure
\STATE $node.addChildren(child1, child2, child3, child4)$

\STATE
\STATE // Recursive calls
\FOR{each $child \in \{child1, child2, child3, child4\}$}
    \STATE $decompose(child)$
\ENDFOR
\end{algorithmic}
\end{algorithm}

\subsection{Using the Hierarchical Decomposition}

With the decomposition of the environment and tree structure defined, we now explain how this hierarchy is used for efficient path planning.

\subsubsection{Initial Training}

Each leaf node is assigned an agent that trains using Q-learning, as the global path planner is assumed to be unavailable. For the tree depicted in Figure~\ref{fig:tree}, this would result in 16 agents, each operating independently in its assigned sub-environment. During training, each agent is restricted to its region, preventing it from crossing into adjacent sub-environments. This ensures localized path optimization, but might cause problems when no charging stations are present in a sub-environment, or when charging stations are unreachable.

The parallel nature of this setup is a major advantage: each agent can be trained concurrently on separate \ac{CPU} cores, greatly improving scalability and efficiency, as training a single agent in a very large environment is highly inefficient. The overall training time, therefore, highly depends on the number and speed of available \ac{CPU} cores.

\subsubsection{Region-Aware Dynamicity Assumption}

To support dynamic environments, we adopt the same assumption as in Yarahmadi et al.’s work. We assume that changes in the environment can be localized to specific leaf nodes. If this were not true, we would be unable to determine the affected leaf nodes, and hence our methodology would be ineffective. Under this assumption, the system can identify the affected leaf sub-environments and trigger retraining only for those affected. This minimalist strategy, which focuses solely on leaf-level updates, forms the basis of the first approach developed in this report, namely \texttt{onlyTrainLeafNodes}.

However, restricting agents to leaf nodes can fail in more complex environments where charging stations are sparse or unreachable in certain leaf nodes. In such cases, retraining in larger regions is needed.

To address this, we leverage the tree structure to ascend the hierarchy and retrain in a larger part of the environment. Each step up expands the scope by a factor of four, potentially including a charging station that was previously inaccessible. While effective, retraining larger sub-environments is computationally more expensive. Therefore, the system must carefully decide when to trigger retraining.

\subsubsection{Retraining Condition} \label{sec:retraining-condition}

\paragraph{Computation of the Success Rate}

To determine when retraining is required, we monitor the success rate of a sub-environment under the current policy:
\begin{equation}
    \text{success rate} = \frac{\text{number of states with a path to a charging station}}{\text{number of states}} 
\end{equation}

A high success rate implies that, under the current policy, most positions allow successful navigation to a charging station. The success rate is always a value between zero and one. To compute the success rate, we use the learned policy and simulate greedy paths, i.e., paths generated by following the highest-value actions. If the path encounters an obstacle, the path is considered invalid. If it encounters a charging station, the found path is valid. A limitation of this greedy method is its reliance on optimal Q-values. If sub-optimal actions receive slightly higher Q-values, paths leading to a charging station may be missed.

To mitigate this issue, we introduce a secondary planning strategy. Instead of following only the single highest Q-value action, the method explores the two best actions under the current policy. This expands the search space and increases the chances of finding a valid path to a charging station without encountering obstacles. The trade-off, however, is higher computational cost, since multiple paths must be generated and evaluated.

An alternative solution would be to extend the Q-learning training duration, allowing the policy to converge more closely to the optimal one and thereby increasing the likelihood that the highest-value actions resemble the optimal ones. However, this would require running Q-learning for a longer period each time retraining is triggered, which would significantly slow down the overall system. For this reason, this approach is not adopted, and instead, we rely on the previously described planning mechanism.

Computing the success rate incurs a significant computational cost, as for each position/state in the sub-environment, multiple paths will be generated to hopefully reach a charging station based on the current policy. While feasible for small sub-environments, this cost increases substantially with larger sub-environments. Consequently, the system prioritizes retraining at the lowest tree levels, corresponding to the smallest sub-environments, to minimize computational overhead.

The \texttt{computeSuccessRate} function, located in the \texttt{treenode.cpp} file\footnote{\url{https://github.com/micss-lab/MARL4DynaPath/blob/main/src/treenode.cpp}}, implements the computation of the success rate. A key detail is that the policy used in the sub-environment is extracted from the root Q-table (explained in Section~\ref{sec:ensuring-consistent-policies}), rather than from the Q-table of the sub-environment itself. This ensures that we always apply the correct policy when retraining occurred at higher levels in the hierarchy, since the root Q-table integrates the Q-tables from different hierarchical levels. This prevents the agent from trying to generate paths within to its sub-environment only when the most recent policy it has learned was in a sub-environment at a higher level in the tree.

\paragraph{Comparing Success Rates}

The system evaluates the success rate of each affected sub-environment between consecutive time steps ($t$ and $t+1$). If the success rate decreases by more than a predefined threshold (set to 0.01 in this report), retraining is initiated. This low threshold ensures rapid adaptation to changes by tolerating only minor reductions in success rate. However, frequent retraining, as observed in Yarahmadi et al.’s approach~\cite{yarahmadi2024comp}, may introduce significant computational overhead.

The success rate accounts for varying impact levels of dynamic changes. A single obstacle change could sever a critical connection, while multiple changes elsewhere may have minimal impact. This allows the system to respond to meaningful declines in performance, rather than relying on arbitrary assumptions about the impact or frequency of environmental changes.

\paragraph{Minimum Expected Success Rate}

In addition to monitoring decreases in success rate, we establish a minimum acceptable success rate of 0.9 (90\%). If a sub-environment’s success rate falls below this threshold after a change or multiple changes, retraining is automatically triggered, irrespective of prior performance. This threshold assumes the general feasibility of finding paths to a charging station in a maze is high, deeming mazes where 10\% or more of the starting positions lack such a path as unrealistic.

Applying this threshold does not guarantee success rates above 0.9 in every sub-environment, but it aims to achieve this target. If the required success rate is not attained at a given level, retraining escalates to a higher tree level.

\subsubsection{Ensuring Consistent Policies} \label{sec:ensuring-consistent-policies}

When agents retrain in sub-environments, the corresponding Q-values must remain consistent across the hierarchy. To maintain this consistency, Q-tables are propagated both upward and downward.

\begin{itemize}
    \item When a leaf or intermediate node is trained or retrained, its Q-values are propagated upward to update the parent nodes, ultimately reaching the root.
    \item Conversely, when a higher-level node is retrained, its learned Q-values are propagated downward to its children.
\end{itemize}

The root Q-table, which represents the entire environment, consolidates the Q-tables from all levels of the hierarchical tree into a single unified Q-table. This table is used to assess the effectiveness of the learned policies. The motivation behind the root Q-table is to remain consistent with the hierarchical retraining process: when training in a leaf sub-environment proves insufficient, such as when no charging station is present, retraining occurs at a higher level. In such cases, the policy relevant to that region is not the one learned locally at the leaf level, but rather the one obtained from the higher-level sub-environment.

Although Q-values are propagated downwards, relying on the policy extracted directly from the leaf sub-environment would be counterintuitive, since we already know that local training there was ineffective. Instead, the root Q-table stores the most up-to-date policy, including any learned at higher levels, ensuring that agents always follow a meaningful strategy. This makes the downward propagation of Q-values, to some extent, unnecessary.

The propagation of Q-table values across the hierarchy is not the most efficient and could likely be optimized, or even avoided, through alternative techniques. Nonetheless, it reflects the intuition that agents should rely on the most effective policies, regardless of the hierarchical level at which they were learned. An advantage of maintaining a root Q-table is that in the policy visualization tool (explained in Section~\ref{sec:policy-visualization}), the learned policies can be extracted directly, without having to traverse the hierarchy to gather them locally.

The propagation of Q-table values is implemented in the \texttt{propagateQTableUpwards} and \texttt{propagateQTable\allowbreak Downwards} functions, located in the \texttt{treenode.cpp} file\footnote{\url{https://github.com/micss-lab/MARL4DynaPath/blob/main/src/treenode.cpp}}.

\subsubsection{Effective Tree Strategy} \label{sec:strategy}

By combining all previously discussed ideas and concepts, we can construct an effective strategy for utilizing the hierarchical tree structure, which is implemented in the \texttt{smartHierarchy} function, located in the \texttt{treestrategy.cpp} file\footnote{\url{https://github.com/micss-lab/MARL4DynaPath/blob/main/src/treestrategy.cpp}}. The detailed procedure is described in Algorithm~\ref{alg:tree-strategy}, which performs the following sequence of steps at every time step.

First, a distinction is made between an initial training scenario and a time step affected by environmental changes based on the initial training variable (\texttt{isInitialTraining}) in block 1 (lines 1-6). In the initial training scenario, no training has been conducted on the environment yet. Therefore, all leaf nodes, i.e., the smallest sub-environments in the hierarchy, are collected for training. In the second scenario, at least one element in the environment has changed. Leveraging the region-aware dynamicity assumption, the system identifies which leaf sub-environments were affected by these changes.

Once the relevant sub-environments are collected (based on either the initial or dynamic scenario), the system checks their success rate. Initially, the success rate is set to -1 to indicate that the sub-environment has not yet been trained. If the success rate is greater than or equal to zero, the system computes the new success rate after the environmental change(s). If the drop in success rate exceeds the threshold of 0.01 (i.e., at least a 1\% decrease) or the new success rate falls below 0.9, the sub-environment is marked for retraining. This retraining condition is located on lines 14-17 in block 2 (lines 9-22).

In rare cases, the new success rate may exceed the previous one and remain above 0.9. In such instances, the old success rate is simply updated. While uncommon, this scenario is handled to ensure correctness.

All sub-environments marked for training or retraining are processed in parallel across multiple \ac{CPU} cores via the \texttt{train} method on line 26 using the specified training mode (\texttt{mode}), selected from single-agent Q-learning, federated Q-learning with equal averaging, or federated Q-learning with importance-based averaging. After training, their updated success rates are checked. Sub-environments with success rates above 0.9 are considered sufficiently reliable and are skipped. Sub-environments with success rates lower than 0.9 that have a parent in the hierarchy mark their parent to be included in the next iteration. This hierarchical escalation reflects the strategy discussed earlier: if local retraining fails to yield sufficient valid paths, expanding to a larger sub-environment may increase the probability of success.

The same sequence of steps is applied to the marked sub-environments in the next iteration, which is generalized by the procedure located in block 3 (lines 29-49). Specifically, their success rate is evaluated, the retraining condition is checked, and retraining is performed if necessary. If a sub-environment’s success rate remains below 0.9 after retraining and it has a parent in the hierarchy, the parent is marked for the next iteration. In the worst case, this process escalates to the root of the tree, necessitating retraining of the entire environment. While this is computationally expensive, we assume that such full retraining will rarely be required. After each retraining step, Q-values are propagated both upward and downward to maintain policy consistency across the hierarchy.

\begin{algorithm}
\caption{Tree Strategy}
\label{alg:tree-strategy}
\begin{algorithmic}[1]
\REQUIRE Affected leaf nodes ($changedLeaves$), training mode ($mode$)
\ENSURE Updated tree with retrained nodes if needed

\STATE $isInitialTraining \gets (changedLeaves = \emptyset)$
\IF{$isInitialTraining$}
    \STATE $leafNodesToTrain \gets$ collect all leaf nodes
\ELSE
    \STATE $leafNodesToTrain \gets changedLeaves$
\ENDIF

\STATE
\STATE // Select leaves to retrain
\STATE $leavesToRetrain \gets \emptyset$
\FOR{each $leaf \in leafNodesToTrain$}
    \IF{$isInitialTraining$}
        \STATE add $leaf$ to $leavesToRetrain$
    \ELSE
        \STATE $oldRate \gets$ stored success rate
        \STATE $newRate \gets$ computeSuccessRate($leaf$)
        \IF{$(oldRate - newRate > 0.01)\ \|\ (newRate < 0.9)$}
            \STATE add $leaf$ to $leavesToRetrain$
        \ELSIF{$newRate > oldRate$}
            \STATE update success rate of $leaf$ to $newRate$
        \ENDIF
    \ENDIF
\ENDFOR

\STATE
\STATE // Retrain leaves
\IF{$leavesToRetrain \neq \emptyset$}
    \STATE $train(leavesToRetrain, mode)$
    \STATE $parentsToRetrain \gets$ parents of leaves with success rate $< 0.9$

    \STATE // Escalate retraining upward
    \WHILE{$parentsToRetrain \neq \emptyset$}
        \STATE $nodesToTrain,\ nextLevelParents \gets \emptyset$
        \FOR{each $node \in parentsToRetrain$}
            \IF{$node$ untrained}
                \STATE add $node$ to $nodesToTrain$
            \ELSE
                \STATE $oldRate \gets$ stored success rate
                \STATE $newRate \gets$ computeSuccessRate($node$)
                \IF{$(oldRate - newRate > 0.01)\ \|\ (newRate < 0.9)$}
                    \STATE add $node$ to $nodesToTrain$
                \ELSIF{$newRate > oldRate$}
                    \STATE update success rate of $node$ to $newRate$
                \ENDIF
            \ENDIF
        \ENDFOR
        \IF{$nodesToTrain \neq \emptyset$}
            \STATE $train(nodesToTrain, mode)$
            \STATE $nextLevelParents \gets$ parents of nodes with success rate $< 0.9$
        \ENDIF
        \STATE $parentsToRetrain \gets nextLevelParents$
    \ENDWHILE
\ENDIF
\end{algorithmic}
\end{algorithm}

\paragraph{Illustration}

\begin{figure}[h]
    \centering
    \includegraphics[width=0.85\linewidth]{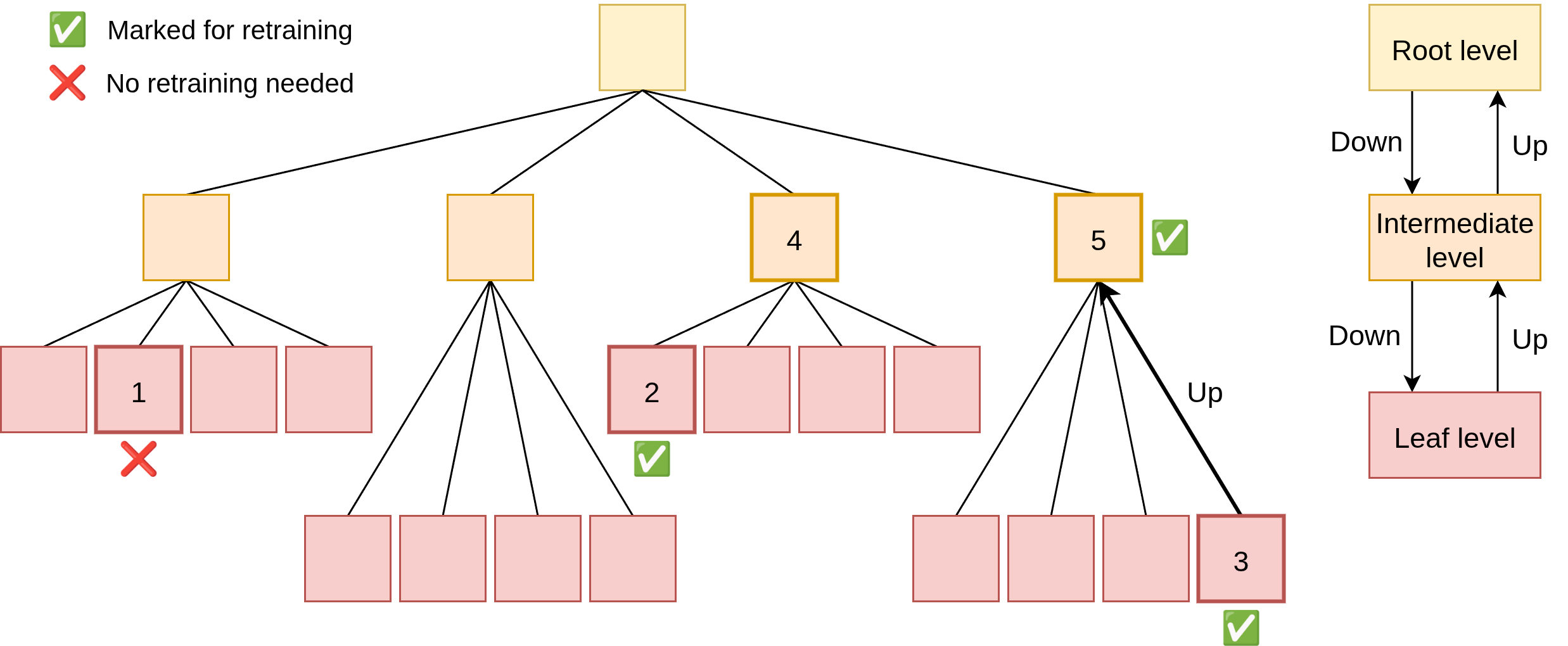}
    \caption{Illustration of the tree strategy discussed in Section~\ref{sec:strategy}.}
    \label{fig:illustration}
\end{figure}

To illustrate the tree strategy, consider the example shown in Figure~\ref{fig:illustration}. The diagram displays a hierarchical tree in which five sub-environments are labeled numerically. Three of these (with red borders) are leaf nodes and represent the smallest sub-environments. The remaining two (with orange borders) are intermediate nodes.

In this example, we assume that the initial training has already been completed; all leaf sub-environments have been trained at least once. During a particular simulation step, multiple obstacle changes occur. As a result, the environment is altered, and the affected leaf sub-environments are identified as sub-environments 1, 2, and 3.

The first iteration begins with these three sub-environments. For each, we compare the old and new success rates. Suppose sub-environment 1 had an old success rate of 0.98, sub-environment 2 had 0.94, and sub-environment 3 had 0.91. After the changes, their new success rates become 0.975, 0.91, and 0.77, respectively. The success rate for sub-environment 1 drops by only 0.005 and remains above 0.9, so it is not marked for retraining (the current policy still suffices). The success rate of sub-environment 2 drops by 0.03 (0.94 to 0.91), exceeding the 0.01 threshold, and is therefore marked for retraining. The success rate of sub-environment 3 drops below 0.9 (0.77) and is also marked for retraining.

After retraining, assume sub-environments 2 and 3 now have success rates of 0.99 and 0.87, respectively. Since sub-environment 2's new success rate exceeds 0.9, its parent (sub-environment 4) is not considered. Sub-environment 3’s parent (sub-environment 5), however, is marked for the next iteration because sub-environment 3's new success rate is still below the threshold (indicated by the arrow labeled Up).

In the second iteration, sub-environment 5 is processed. Suppose its old success rate was 0.93 and its new one is 0.91. Since the drop (0.02) exceeds the 0.01 threshold, it is marked for retraining. After retraining, its success rate improves to 0.97, which is above the 0.9 threshold, so the process terminates.

When sub-environments are retrained, their Q-tables are propagated upwards and downwards to ensure consistent policies. The upward propagation of Q-tables is the most important step to ensure the root Q-table consolidates the Q-tables from different hierarchical levels.

\subsection{Time Complexity Analysis}

The proposed framework relies on two interdependent procedures: the \textit{Recursive Environment Decomposition} (Algorithm~\ref{alg:decomposition}) and the \textit{Tree Strategy} (Algorithm~\ref{alg:tree-strategy}). 

Algorithm~\ref{alg:decomposition} recursively partitions the environment into sub-regions until each sub-region is at most the predefined size threshold, i.e., $20 \times 20$. At each recursive step, a node is split into four child nodes, leading to the recurrence relation given in Eq. (\ref{Eq8}):  
\begin{equation}\label{Eq8}
T(N) = 4T\!\left(\tfrac{N}{4}\right) + O(1)
\end{equation}

where $N = R \times C$ denotes the environment size in terms of rows and columns. By applying the Master Theorem~\cite{cormen2022introduction}, the recurrence in Eq. (\ref{Eq8}) simplifies to Eq. (\ref{Eq9}):
\begin{equation}\label{Eq9}
  T(N) = \Theta(N)
\end{equation}

Hence, Algorithm~\ref{alg:decomposition} runs in linear time with respect to the environment area, i.e., $\Theta(RC)$. The recursion depth is logarithmic in the maximum dimension, i.e., $\Theta(\log \max\{R,C\})$.

Algorithm~\ref{alg:tree-strategy} builds upon this decomposition and governs the training and retraining of tree nodes. In the initial training phase, all leaf nodes are traversed once and their ancestors are possibly retrained, which, in the worst case, visits every node in the tree. Therefore, the complexity is denoted by Eq. (\ref{Eq10}):
\begin{equation}\label{Eq10}
T_{\text{init}} = \Theta(N)
\end{equation}

which is again linear in the number of nodes. During incremental updates, only a subset of changed leaves needs to be processed. Let $K$ denote the number of affected (changed) leaf nodes. In this case, both the $K$ leaves and their ancestors are considered for possible retraining. In a balanced quadtree, the number of ancestors per leaf is bounded by the tree height $h = \Theta(\log M)$, where $M = \max\{R,C\}$. Eq. (\ref{Eq11}) shows the incremental retraining cost:
\begin{equation}\label{Eq11}
 T_{\text{incr}} = O\!\big(\min\{N, K \log M\}\big)
  \end{equation}
Thus, the proposed method achieves linear time complexity with respect to the environment size for the initial phase, while supporting incremental updates bounded by the logarithmic height and proportional to the number of affected leaves $K$. This ensures both scalability to large environments and efficiency during adaptive retraining.

\subsection{Single-Agent Reinforcement Learning} \label{sec:single-agent-rl}

The final remaining component of the methodology concerns how individual sub-environments are trained to adapt to changes. The first two \ac{RL}-based approaches evaluated in this report adopt single-agent Q-learning, meaning that each sub-environment is trained using a single agent that collects experience and learns a policy via the Q-learning algorithm. The implementation of single-agent Q-learning is located in the \texttt{singleagent.cpp} file\footnote{\url{https://github.com/micss-lab/MARL4DynaPath/blob/main/src/singleagent.cpp}}.

The first approach, referred to as \texttt{onlyTrainLeafNodes}, restricts itself to the leaf level of the hierarchical tree. At every time step, this method identifies the leaf sub-environments affected by changes and retrains them using single-agent Q-learning. Its simplicity makes it an ideal baseline for understanding whether localized retraining at the leaf level alone is sufficient for adapting to dynamic environments, or whether a more advanced strategy (as discussed in Section~\ref{sec:strategy}) is required. Its implementation is found in the \texttt{onlyTrainLeafNodes} function, located in the \texttt{treestrategy.cpp} file\footnote{\url{https://github.com/micss-lab/MARL4DynaPath/blob/main/src/treestrategy.cpp}}.

The second approach, called \texttt{singleAgent}, combines single-agent Q-learning with the full tree-based strategy from Section~\ref{sec:strategy}. This allows it to traverse the tree hierarchy, retraining in larger sub-environments when necessary. This approach will enable us to evaluate whether the hierarchical strategy improves adaptability and overall performance in dynamic settings.

\subsubsection{Additional Techniques}

In practice, Q-learning alone proved insufficient for efficient learning in dynamic environments. Therefore, several supporting techniques were introduced to accelerate convergence and improve the quality of the learned policies.

\paragraph{Experience Replay Buffer}

The first technique is the experience replay buffer. This buffer stores up to 1000 experiences (i.e., transitions), which are periodically replayed to reinforce learning. Once full, the buffer is cleared and filled with more recent experiences.

This mechanism speeds up convergence by allowing the agent to reuse valuable experiences. The experience replay buffer is only used in the \texttt{onlyTrainLeafNodes} and \texttt{singleAgent} approaches, but is omitted in the federated Q-learning setting, where maintaining separate buffers for multiple agents is computationally expensive.

\paragraph{Prioritized Replay}

The second technique builds on the idea of prioritized replay, which focuses learning on more difficult or infrequently successful experiences. Rather than prioritizing samples within the experience buffer, we use this concept to guide the selection of starting positions for episodes.

Some starting positions (e.g., those near a charging station) are naturally easier than others. To balance learning, we track how often each position is selected and how often a successful path to a charging station is planned from it. Positions with lower success are sampled more frequently, allowing the agent to focus on learning from harder-to-solve positions. This strategy improves final policy quality and ensures better overall success across the state space.

Without this technique, each position would be sampled uniformly, potentially requiring more time for the agent to learn from the harder positions.

\subsubsection{Convergence Detection}

To keep training efficient, we incorporate a convergence detection mechanism, as described in Section~\ref{sec:convergence}. Specifically, the Q-learning algorithm checks for convergence every 50 episodes using a threshold of $5 \times 10^{-4}$.

When the Q-values stabilize below this threshold, training is stopped. This mechanism ensures that each sub-environment is trained sufficiently to learn a reliable policy while maintaining efficiency.

\subsection{Multi-Agent Reinforcement Learning} \label{sec:marl-methodology}

The final two \ac{RL}-based path-planning approaches adopt a distinct strategy from single-agent Q-learning. In real-world scenarios, many simultaneous obstacle changes are rather unlikely; instead, changes typically occur sequentially, with one obstacle change per time step, though occasional bursts of multiple changes within a single time step may occur. This observation enables further optimization of our methodology. As multiple \ac{CPU} cores can operate in parallel, but most of the time only one sub-environment is affected by a change, most \ac{CPU} cores remain idle. Consequently, a Q-learning-based training procedure involving multiple agents training concurrently, each on a separate \ac{CPU} core, optimizes the utilization of all available CPU cores.

To align with these objectives, we draw on the work of Woo et al.~\cite{woo2025blessing}, which analyzes federated Q-learning. This approach enhances Q-learning by enabling multiple agents to learn policies and periodically aggregate their Q-tables. Woo et al. explore both synchronous and asynchronous variants of federated Q-learning. This technical report focuses on the asynchronous variant, as its experiments demonstrate superior performance, requiring fewer samples to estimate the optimal action-value function.

In the asynchronous setting, each agent independently collects trajectories using its local behavior policy. These behavior policies can differ across agents over time and are not necessarily optimal. Each agent updates its local Q-values using the standard Q-learning rule (see Equation~\eqref{eq:q-learning}). Periodically, agents' Q-estimates are aggregated into a shared Q-table. The method used for this aggregation is what distinguishes the different algorithm variants.

In the subsequent sections, we describe the \ac{FedAsynQ} algorithm and explain its adaptation and extension for our context. Additionally, we discuss two variants based on distinct aggregation schemes.

The implementation of both variations can be found in the \texttt{multiagent.cpp} file\footnote{\url{https://github.com/micss-lab/MARL4DynaPath/blob/main/src/multiagent.cpp}}.

\subsubsection{FedAsynQ}

The \ac{FedAsynQ} algorithm estimates the optimal action-value function by averaging local Q-tables from multiple agents at regular intervals. Each iteration $t$ of the algorithm proceeds as follows:

First, each agent $k \in [K]$ samples a transition $(s^k_{t-1}, a^k_{t-1}, r^k_{t-1}, s^k_t)$ from its trajectory and updates its local Q-estimate $Q^k_{t-1}$ to reach some intermediate estimate $Q^k_{t-\frac{1}{2}}$ using:

\begin{equation} \label{eq:fedasynq-local-update}
    Q^k_{t-\frac{1}{2}}(s,a) = 
    \begin{cases}
        (1-\eta)Q^k_{t-1}(s,a) + \eta(r^k_{t-1}+\gamma V^k_{t-1}(s^k_t)),& \text{if } (s,a) = (s^k_{t-1},a^k_{t-1})\\
        Q^k_{t-1}(s,a),& \text{otherwise}
    \end{cases}
\end{equation}

Then, every $\tau$ iterations (where $\tau \ge 1$), the agents’ intermediate Q-estimates are averaged using the following weighted scheme:

\begin{equation} \label{eq:fedasynq-averaging}
    \forall (s,a) \in S \times A:\quad Q^k_t(s,a) = 
    \begin{cases}
        \sum^K_{k=1}\alpha^k_t(s,a)Q^k_{t-\frac{1}{2}}(s,a),& \text{if } t \equiv 0 \Mod{\tau}\\
        Q^k_{t-\frac{1}{2}}(s,a),& \text{otherwise}
    \end{cases}
\end{equation}

In the equation above, $\alpha^k_t = [\alpha^k_t(s,a)]_{s\in S,a\in A} \in [0,1]^{|S||A|}$ is an entry-wise weight assigned to agent $k$ such that:

\begin{equation}
    \forall(s,a) \in S \times A: \quad \sum^K_{k=1}\alpha^k_t(s,a) = 1
\end{equation}

After a total of $T$ iterations, the final global Q-estimate is given by:

\begin{equation} \label{eq:fedasynq-final-q-table}
    Q_T(s,a) = \sum^K_{k=1} \alpha^k_T(s,a)Q^k_T(s,a)
\end{equation}

\begin{figure}[h]
    \centering
    \includegraphics[width=0.85\linewidth]{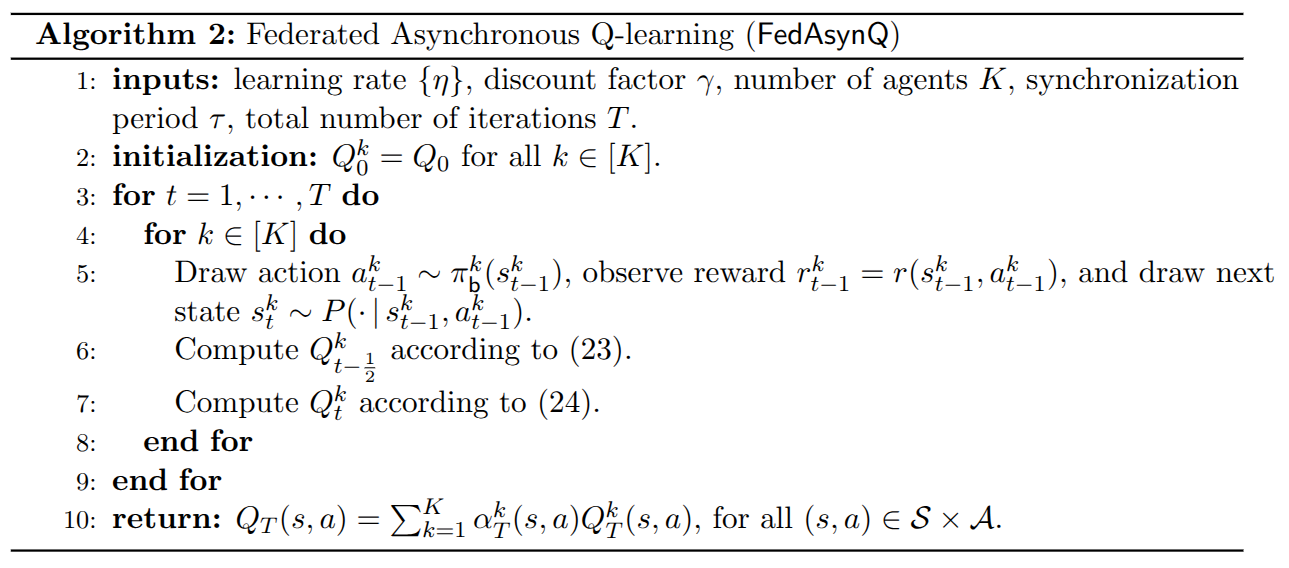}
    \caption{Overview of the FedAsynQ algorithm.}
    \label{fig:fedasynq}
\end{figure}

Figure~\ref{fig:fedasynq} shows a high-level overview of the \ac{FedAsynQ} algorithm. Each agent interacts with the environment, collects experience, performs updates, and periodically contributes to a global average Q-table.

Our implementation differs slightly from the original. We parallelize agent execution to accelerate learning, and each agent gathers experience independently for a fixed number of steps (set to 1000 in our case) before the global averaging step is performed. The reasoning behind this choice is that local updates are inexpensive, while averaging is costly. However, if averaging is delayed too long, agents’ policies may diverge too much, destabilizing the learning process. Hence, we apply averaging regularly.

We set the learning rate to $\eta = 0.4$, the discount factor to $\gamma = 0.9$, and use $K = 12$ agents. The synchronization period is $\tau = 1000$ steps. The total number of iterations $T$ is scaled with the sub-environment size: $T = r \times c \times 200$, where $r \times c$ are the sub-environment's dimensions. This scaling ensures that larger sub-environments are given more training iterations to determine a good policy.

Twelve agents were chosen as a balance between parallelism and practicality; more agents increase learning speed but require more \acp{CPU} and increase aggregation complexity. Also, unlike in single-agent Q-learning, we do not use convergence-based stopping. Due to the instability of federated learning and asynchronous updates, we found it difficult to define a reliable convergence condition. Instead, we rely on the fixed iteration scheme described above.

\paragraph{FedAsynQ-EqAvg}

In the basic version of \ac{FedAsynQ}, each agent is given equal importance when computing the average. That is:

\begin{equation}
    \alpha^k_t(s,a) = 1/K
\end{equation}
 
This simple uniform averaging scheme assumes that all agents contribute equally and simplifies computation. Woo et al.~\cite{woo2025blessing} also show that under certain conditions, this method has guaranteed finite-time convergence. However, it may not be sample-efficient, especially in settings where agents visit some state-action pairs far more frequently than others.

To evaluate how well this method performs in dynamic environments, we created a third \ac{RL}-based path planning approach called \texttt{FedAsynQ\_EqAvg}. This approach uses the hierarchical strategy described in Section~\ref{sec:strategy}, combined with \ac{FedAsynQ} and equal weighting.

\paragraph{FedAsynQ-ImAvg}

A more sophisticated weighting scheme uses importance averaging, which accounts for the frequency of visits to each state-action pair. In asynchronous settings, agents may explore the state space unevenly. Some state-action pairs may be visited frequently by a few agents and rarely or never by others. Uniform averaging would give equal weight to all Q-estimates, which can misrepresent the true value of rarely visited pairs.

To address this, the importance weights are computed as:

\begin{equation} \label{eq:fedasynq-imavg}
    \alpha^k_t(s,a) = \frac{(1-\eta)^{-N^k_{t-\tau,t}(s,a)}}{\sum^K_{k'=1}(1-\eta)^{-N^{k'}_{t-\tau,t}(s,a)}}    
\end{equation}

Here, $N^k_{t-\tau,t}(s,a)$ is the number of times agent $k$ visited state-action pair $(s,a)$ during the interval $[t - \tau, t)$. Intuitively, agents that visit a state-action pair more often contribute more to its final Q-value. For example, if an agent is the only one visiting the state-action pair $(s,a)$, its estimate will dominate the aggregated value.

This particular weighting scheme is computationally more expensive than uniform averaging, as it requires tracking visit counts for each agent and each state-action pair. Additionally, the weights must be recomputed at each synchronization step. However, it potentially offers higher accuracy and better sample efficiency.

To test this importance averaging weighting scheme, we created a fourth and final path planning approach called \texttt{FedAsynQ\_ImAvg}, which combines the hierarchical training strategy from Section~\ref{sec:strategy} with \ac{FedAsynQ} using importance-based weights.

\clearpage

\section{Evaluation} \label{sec:evaluation}

The work of Yarahmadi et al.~\cite{yarahmadi2024comp} exhibits both methodological and evaluation-related limitations. The methodological issues were addressed in Section~\ref{sec:methodology}; this section focuses on the shortcomings of their evaluation strategy and provides a more rigorous assessment of our proposed methodology.

Two main limitations were identified in their evaluation. The first is the reliance on overly simple environments, characterized by abundant free space, few obstacles, and many charging stations. Such settings make path planning considerably easier: with numerous charging stations, shorter paths are more likely, and the scarcity of obstacles increases the number of feasible routes. To address this, in our experiment setup, we design environments of varying difficulty, categorized as easy, medium, and hard, based on the densities of obstacles, charging stations, and free space. This broader set of test cases enables a more meaningful evaluation of robustness, i.e., whether our methodology performs consistently across different levels of complexity.

The second limitation concerns their treatment of dynamic environments. In their simulations, only one obstacle changes position per time step. While this simplifies the problem, it does not reflect realistic dynamics, where multiple changes may occur simultaneously. Although such cases are less frequent, they can significantly affect planning performance. To better capture this, our experiment setup allows multiple simultaneous obstacle changes per time step. Specifically, up to ten changes can occur, with probabilities decreasing as the number of changes increases. This design more closely mirrors real-world conditions, where small changes are more common than large-scale shifts.

\subsection{Experiment Setup}

The experiments in this technical report were conducted using a standardized setup implemented in the \texttt{runFull\allowbreak Experiment} function, located in the \texttt{experiments.cpp} file\footnote{\url{https://github.com/micss-lab/MARL4DynaPath/blob/main/src/experiments.cpp}}. The function is responsible for running simulations and is invoked from the \texttt{main.cpp} file\footnote{\url{https://github.com/micss-lab/MARL4DynaPath/blob/main/src/main.cpp}}. 

The experiment setup used ensures consistency across all tested approaches, providing a controlled environment for performance comparison. Additional technical details and instructions on how to replicate the experiments are provided in Section~\ref{sec:implementation-running-details}.

\subsubsection{Definition of the Environments}

The environments studied in this report are mazes represented as two-dimensional grids, following the format introduced by Yarahmadi et al. Each maze is composed of free spaces, dynamic obstacles, and charging stations, serving as a classical testbed for evaluating path planning approaches in dynamic settings where the goal is to reach a destination (charging station) under changing conditions.

Formally, each environment is modeled as an \ac{MDP}. For a maze of size $r \times c$, the \ac{MDP} is defined as follows:

\begin{itemize}
    \item $S = \{ (i, j) \mid i \in \{0, \dots, r-1\}, j \in \{0, \dots, c-1\} \}$ (the set of all grid positions in the maze)
    \item $A = \{0, 1, 2, 3, 4, 5, 6, 7\}$ (actions: 0 = N, 1 = NE, 2 = E, ..., 7 = NW)
    \item $R_a(s, s') = 
        \begin{cases}
            100.0,& \text{if the agent moves onto a charging station}\\
            -10.0,& \text{if the agent attempts to move into an obstacle (the agent remains in place)}\\
            -1.0,& \text{for all other valid moves}
        \end{cases}$
    \item $P_a(s, s') \in \{0,1\}$ (deterministic transitions)
    \item $\gamma = 0.9$ (discount factor)
\end{itemize}

This reward structure encourages agents to find the shortest path to a charging station. Reaching a charging station yields a high reward, while bumping into an obstacle incurs a heavy penalty. Each step taken also incurs a small penalty to discourage unnecessary movement.

\subsubsection{Difficulty of the Environments}

We evaluate our approaches in environments of three difficulty levels, determined by the relative proportions of free space, obstacles, and charging stations. These environments are randomly initialized using a seed value, with cell types assigned according to the specified probability distributions.

\begin{itemize}

    \item \textbf{Easy difficulty}
    \begin{itemize}
        \item Free space: 0.8
        \item Obstacles: 0.18
        \item Charging stations: 0.02
    \end{itemize}
    
    \item \textbf{Medium difficulty}
    \begin{itemize}
        \item Free space: 0.7
        \item Obstacles: 0.29
        \item Charging stations: 0.01
    \end{itemize}
    
    \item \textbf{Hard difficulty}
    \begin{itemize}
        \item Free space: 0.6
        \item Obstacles: 0.395
        \item Charging stations: 0.005
    \end{itemize}
    
\end{itemize}

Easy environments, with more free space, charging stations, and fewer obstacles, are expected to be easier for agents to navigate. Medium environments have more obstacles and fewer charging stations, presenting a moderate challenge. Hard environments feature dense obstacles and very few charging stations, making path planning significantly more difficult, especially in dynamic conditions where a single change could block a crucial bottleneck path.

\subsubsection{Size of the Environments}

Larger environments inherently increase computational complexity due to the growing number of states and state-action pairs. To evaluate the scalability of our approach, we test it across the following environment sizes:

\begin{itemize}
    \item $20 \times 20$, with a single-level tree (root only)
    \item $50 \times 50$, with a three-level tree (root, intermediate, and leaf level)
    \item $100 \times 100$, with a four-level tree (root, two intermediate levels, and leaf level)
    \item $200 \times 200$, with a five-level tree (root, three intermediate levels, and leaf level)
    \item $300 \times 300$, with a five-level tree (root, three intermediate levels, and leaf level)
\end{itemize}

Smaller environments typically yield simpler hierarchies, often comprising only a root node, whereas larger environments result in deeper hierarchies with multiple levels. This range of hierarchical tree depths enables assessment of the system’s scalability with respect to the environment size. Consistent performance across increasing environment sizes and deeper hierarchical trees suggests that the observed trends can be generalized to even larger mazes.

\subsubsection{Simulating Environment Changes}

To evaluate adaptability over time, we simulate multiple time steps for each environment size. Generally, obstacle changes in smaller environments have a larger relative impact, whereas changes in large environments affect only a small portion of the map.

To compensate for this, we simulate more time steps for larger environments. Specifically, we simulate $2 \times r$ time steps for an $r \times r$ environment. This ensures that across time, the environment undergoes enough changes to meaningfully test the adaptability of our approach. However, simulating large environments for many time steps is computationally expensive, so we restrict testing to environments no larger than $300 \times 300$.

The number of obstacle changes per time step is sampled from a skewed distribution that tries to reflect real-world likelihoods: smaller numbers of changes are more probable than larger ones. The following piece of code illustrates the sampling logic:

\begin{lstlisting}
    int r = rand() % 1000;
    int numChanges;
    if (r < 900) numChanges = 1;
    else if (r < 950) numChanges = 2;
    else if (r < 970) numChanges = 3;
    else if (r < 980) numChanges = 4;
    else if (r < 987) numChanges = 5;
    else if (r < 992) numChanges = 6;
    else if (r < 995) numChanges = 7;
    else if (r < 997) numChanges = 8;
    else if (r < 999) numChanges = 9;
    else numChanges = 10;
\end{lstlisting}

As shown, the likelihood of two or three simultaneous changes is small, while the likelihood of nine or ten changes is extremely rare. This more realistic simulation of dynamic changes ensures that our method is evaluated under diverse, yet plausible, conditions.

Lastly, to ensure meaningful adaptation occurs in every time step, we guarantee that at least one obstacle change happens. Time steps without any changes do not require replanning and are therefore excluded from the simulation.

The \texttt{simulateEnvironmentChanges} function, located in the \texttt{experiments.cpp} file\footnote{\url{https://github.com/micss-lab/MARL4DynaPath/blob/main/src/experiments.cpp}}, implements the simulation of environment changes.

\subsubsection{Baseline Algorithms}

In Section~\ref{sec:methodology}, we introduced four \ac{RL}-based approaches for path planning in dynamic environments. To assess their performance, we compare their relative performance and benchmark them against two baseline approaches based on the A* algorithm, a traditional path-planning method.

The A* algorithm is implemented in the \texttt{astar.cpp} file\footnote{\url{https://github.com/micss-lab/MARL4DynaPath/blob/main/src/astar.cpp}}.

\paragraph{A*}

The strongest baseline in terms of path quality and accuracy is the traditional A* path planning algorithm. A* is widely used in applications such as games due to its ability to quickly compute paths using an admissible heuristic. As a shortest-path algorithm, A* is guaranteed to find the optimal path between two points, provided the heuristic is chosen correctly.

As discussed earlier, traditional path planning algorithms like A* are not inherently designed to handle dynamic environments. A* requires full knowledge of the environment at each time step, specifically, which cells are free space, obstacles, or charging stations. This contradicts the assumptions of a truly dynamic setting, where such information is not always known in advance, and changes may be unpredictable.

Despite this limitation, we use two variants of A* in our experiments to serve as theoretical performance bounds in idealized conditions where the full environment is always known.

Our A* implementation uses the Chebyshev distance heuristic (Equation~\eqref{eq:chebyshev}), which is more appropriate than the Manhattan distance given that agents can move in eight directions. The implementation iterates over all positions in the environment, attempting to find a path to a charging station. To improve performance, we apply the principle that any sub-path of a shortest path is also a shortest path. This allows us to reuse previously discovered paths to reduce redundant computation. Although the implementation is still not the most efficient, it is sufficient to serve as a reasonable benchmark.

\begin{equation} \label{eq:chebyshev}
    d((x_1,y_1),(x_2,y_2)) = \max\{|x_1-x_2|,|y_1-y_2|\}
\end{equation}

\paragraph{A* Static}

The first variant, \texttt{A* Static}, computes all paths once at the beginning of the simulation and does not update them in response to changes. This baseline helps evaluate how a static plan degrades over time as dynamic changes accumulate. Ideally, the success rate will be high initially but will gradually decrease as the environment evolves and paths become blocked.

This variant also serves as a validation tool: if changes in the environment do not reduce its accuracy over time, there may be an issue with the simulation logic, as obstacle changes should eventually invalidate some paths.

\paragraph{A* Oracle}

The second variant, \texttt{A* Oracle}, assumes full knowledge of the environment at every time step. It recomputes paths across the entire grid whenever changes occur. While this approach is highly accurate and always finds the shortest paths if they exist, it is computationally expensive, especially in large environments. Nonetheless, it provides a strong benchmark, helping us assess how close our \ac{RL}-based approaches come to this ideal performance level in terms of accuracy and path quality.

\subsubsection{Edge Case}

\begin{figure}[h]
    \centering
    \includegraphics[width=0.85\linewidth]{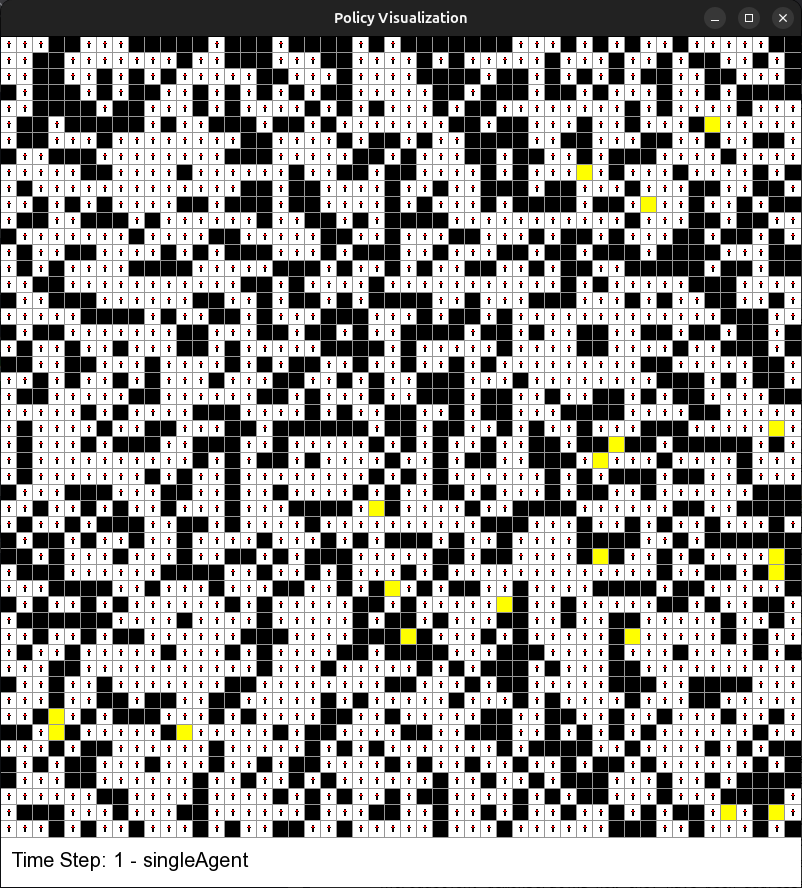}
    \caption{a $50 \times 50$ edge case environment: the top-left quadrant does not contain any charging stations (yellow), making path planning significantly more challenging.}
    \label{fig:edge-case}
\end{figure}

To thoroughly test the limits of our methodology, we designed a challenging $50 \times 50$ maze in which path planning is particularly difficult, i.e., finding valid paths is far from trivial. In this edge case, charging stations are placed such that, when the environment is conceptually divided into four quadrants (by splitting both dimensions in half), one of the quadrants contains no charging stations at all. Specifically, the top-left quadrant covering a $25 \times 25$ region lacks any charging stations. This configuration is illustrated in Figure~\ref{fig:edge-case}.

The purpose of this difficult environment is to demonstrate that our methodology, which is built on the hierarchical decomposition introduced in Section~\ref{sec:hierarchical-decomposition} and the tree strategy described in Section~\ref{sec:strategy}, can still achieve reasonable accuracy even under extremely adverse conditions. However, it is also expected that in such hard environments, retraining at higher levels in the hierarchy is required more frequently. Consequently, a drop in efficiency is expected to maintain acceptable accuracy levels.

\subsubsection{Hardware Configuration for Experimental Setup}

All experiments were conducted on a single machine with the following hardware specifications:

\begin{itemize}
    \item \textbf{Operating System (OS):} 64-bit Linux Ubuntu 24.04.2 LTS
    \item \textbf{CPU:} 11th Gen Intel(R) Core(TM) i7-1165G7 x 8 (2.80 GHz)
    \item \textbf{RAM:} 16.0 GiB
    \item \textbf{Storage:} 1.0 TB SSD
\end{itemize}

\subsection{Evaluation Measures}

To compare the performance of the two A* variants and the four \ac{RL}-based methods from Section~\ref{sec:methodology}, we define a comprehensive set of evaluation measures. Note that all time measurements refer to wall-clock time, which strongly depends on the machine configuration. Faster CPU cores can reduce execution time by performing more cycles per second, while the number of available cores also plays a critical role, as parallelization across agents can greatly decrease computation time. However, the speedup is not linear with respect to the number of cores, in accordance with Amdahl's law. In addition, for federated Q-learning, parallel training with more agents increases the aggregation overhead, since a larger number of Q-tables must be combined.

\subsubsection{Accuracy}

Accuracy is measured using the success rate, which is defined as the ratio of positions in the environment from which a valid path to a charging station exists under the current policy, to the total number of positions in the environment. This metric reflects the effectiveness of the path planning approach. Ideally, the success rate should remain high throughout the simulation. If a sudden drop occurs, perhaps due to a critical obstacle change, we expect the approach to recover quickly. Accuracy results are visualized as line graphs showing the average success rate across various environment sizes and difficulty levels.

\subsubsection{Adaptation Time}

This metric captures the time required to adapt to environmental changes at each time step. As the primary goal of this report is to develop an efficient \ac{RL}-based approach for dynamic path planning, adaptation time is a key performance indicator. It is visualized using box plots to show the mean, variance, and outliers, as well as line graphs to illustrate trends across different maze sizes.

\subsubsection{Cumulative Adaptation Time}

The cumulative adaptation time aggregates the computational time required for all adaptation steps up to time step $t$. It provides insight into the overall adaptational cost of an approach. Line graphs are used to depict how steeply the cumulative adaptation time grows over time. Ideally, the curve should be as flat as possible; sharp increases indicate inefficient adaptation.

\subsubsection{Average Path Length}

This metric quantifies the average length of all valid paths, one per starting position, determined at a given time step. Only successful paths reaching a charging station are included. Shorter average path lengths suggest more optimal planning. This metric enables comparison with the \texttt{A* Oracle} approach, which consistently yields the shortest possible paths.

\subsubsection{Initial Training Time}

This metric captures the time required to train on the environment for the first time. For both baseline methods, we assume that the environment is fully known in advance, as they require a complete representation to function properly. In contrast, our four developed approaches operate under the assumption that the environment is initially unknown. Through interaction, they gradually learn a policy to navigate the environment effectively. This assumption increases the general applicability of our methods. Ideally, the initial training time should be as low as possible to ensure fast deployment.

\subsection{Simulation Results} \label{sec:simulation-results}

This section presents the results obtained by the six approaches outlined below from simulating dynamic environments.

\begin{itemize}
    \item \textbf{A* Static:} Performs path planning only at the beginning of the simulation and does not update the paths afterward.
    \item \textbf{A* Oracle:} Replans paths each time step when obstacle changes occur, assuming full knowledge of the environment at all times.
    \item \textbf{onlyTrainLeafNodes:} Retrains only the leaf nodes of the hierarchical tree (the smallest sub-environments) whenever affected by changes, using single-agent Q-learning.
    \item \textbf{singleAgent:} Combines single-agent Q-learning with the hierarchical tree strategy, selectively retraining affected sub-environments based on the retraining condition.
    \item \textbf{fedAsynQ\_EqAvg:} Utilizes federated Q-learning with equal averaging of the local Q-estimates to retrain sub-environments. It combines the tree strategy with the retraining condition.
    \item \textbf{fedAsynQ\_ImAvg:} Similar to the previous approach, but uses federated Q-learning with importance averaging for retraining.
\end{itemize}

To keep the evaluation concise and focused, we mainly limit the discussion of simulation results to two environment sizes: $50 \times 50$ and $300 \times 300$, and two difficulty levels: easy and hard. These selected configurations provide a representative overview of performance across both small and large environments, as well as under varying complexity. While we tested additional sizes and difficulty levels, discussing all of them in detail would be overly exhaustive. The insights gained from the selected cases can be reasonably generalized to the remaining configurations.

\subsubsection{Accuracy}

\paragraph{Easy Complexity}

\begin{figure}[h]
    \centering
    \includegraphics[width=0.70\linewidth]{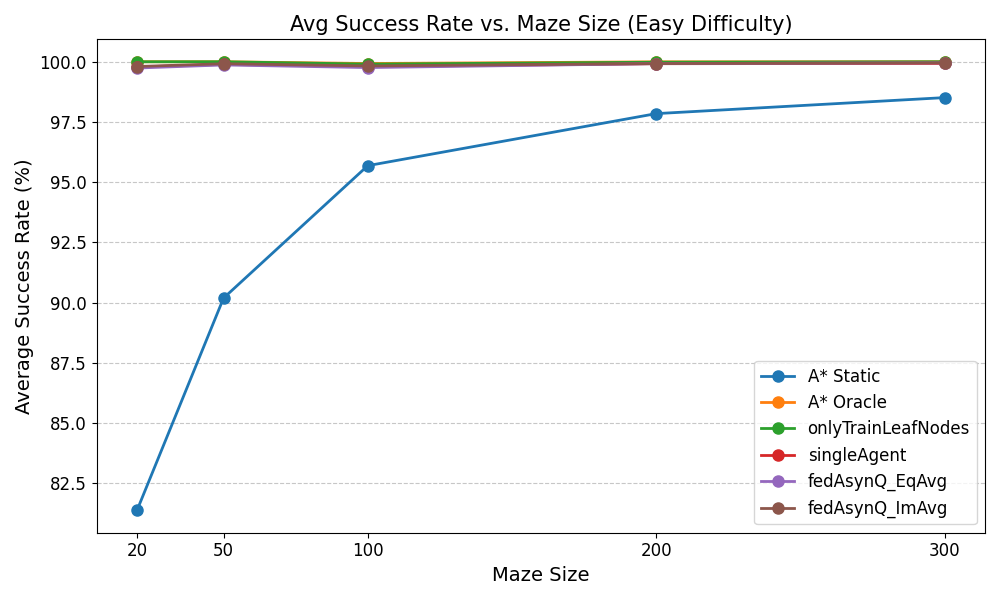}
    \caption{Success rate of different path planning approaches in easy difficulty environments.}
    \label{fig:evaluation-accuracy-easy}
\end{figure}

Figure~\ref{fig:evaluation-accuracy-easy} presents the accuracy results for easy maze environments. Among all approaches, \texttt{A* Static} consistently performs the worst. It achieves an average accuracy of roughly 81\% in the $20 \times 20$ maze, which increases to about 98\% for the $300 \times 300$ maze. This upward trend is not due to improved performance in larger environments but rather because obstacle changes have a less significant relative impact in larger environments. Although the number of simulated time steps is scaled with environment size, smaller mazes are still more sensitive to changes.

\texttt{A* Oracle} performs flawlessly across all environment sizes, maintaining a near 100\% success rate, which is unsurprising given its full knowledge of the environment and its ability to replan optimally.

Among the \ac{RL}-based approaches, \texttt{onlyTrainLeafNodes} achieves excellent results, matching the performance of \texttt{A* Oracle} in all environment sizes. This demonstrates that always initiating retraining in affected leaf sub-environments, even without assessing the impact, achieves the highest accuracy for easy mazes.

The three more advanced approaches (\texttt{singleAgent}, \texttt{fedAsynQ\_EqAvg}, and \texttt{fedAsynQ\_ImAvg}) also achieve high accuracy. However, for the $20 \times 20$ maze, their accuracy is slightly below 100\%, as retraining is only performed when the success rate drops, since they employ the retraining condition. This makes these methods more efficient while still maintaining high accuracy.

\paragraph{Medium Complexity}

\begin{figure}[h]
    \centering
    \includegraphics[width=0.70\linewidth]{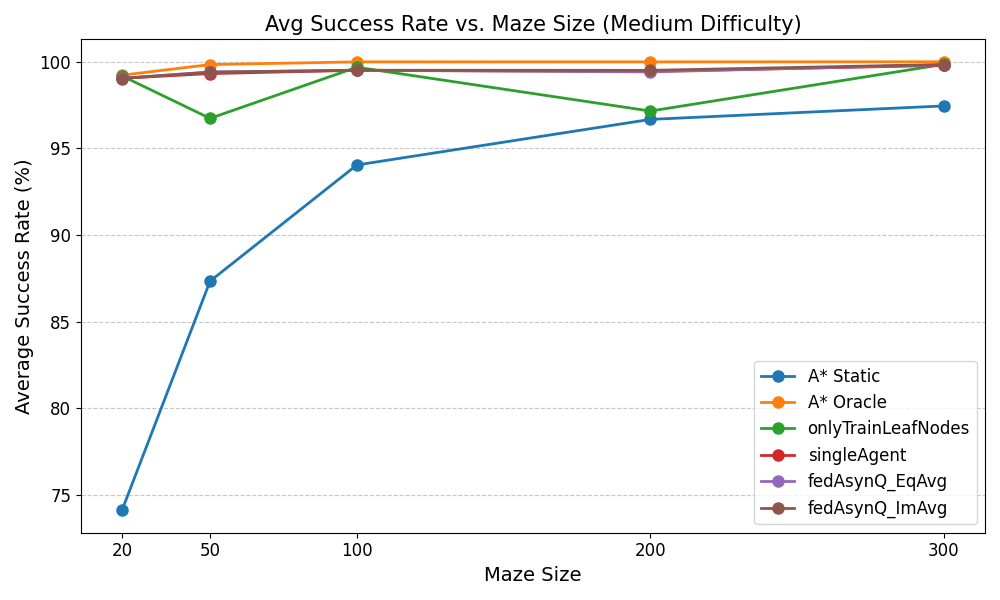}
    \caption{Success rate of different path planning approaches in medium difficulty environments.}
    \label{fig:evaluation-accuracy-medium}
\end{figure}

For the medium-complexity environments, shown in Figure~\ref{fig:evaluation-accuracy-medium}, \texttt{A* Oracle} remains the top-performing approach. \texttt{A* Static} continues to perform the worst across all maze sizes.

\texttt{onlyTrainLeafNodes} shows inconsistent performance, particularly poor on the $50 \times 50$ and $200 \times 200$ mazes, though it performs reasonably well on other sizes. This inconsistency suggests that only retraining at the leaf level may not be sufficient in more complex environments.

The three sophisticated approaches (\texttt{singleAgent}, \texttt{fedAsynQ\_EqAvg}, and \texttt{fedAsynQ\_ImAvg}) maintain high accuracy levels, closely matching the \texttt{A* Oracle} baseline, highlighting the effectiveness of the tree-based retraining strategy.

\paragraph{Hard Complexity}

\begin{figure}[h]
    \centering
    \includegraphics[width=0.70\linewidth]{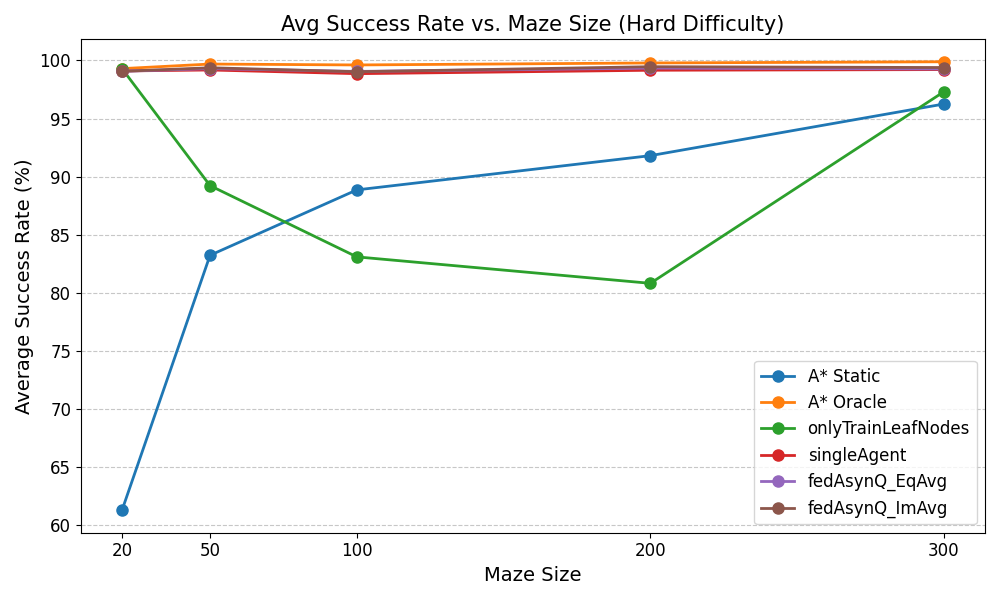}
    \caption{Success rate of different path planning approaches in hard difficulty environments.}
    \label{fig:evaluation-accuracy-hard}
\end{figure}

\texttt{A* Oracle} again delivers near-perfect accuracy for the hardest environments, shown in Figure~\ref{fig:evaluation-accuracy-hard}. \texttt{A* Static} maintains its pattern of low accuracy across all maze sizes, especially in smaller environments in which obstacle changes have more effect.

The \texttt{onlyTrainLeafNodes} approach performs poorly under these hard conditions. On the $50 \times 50$, $100 \times 100$, and $200 \times 200$ mazes, it achieves below 90\% accuracy, and in some cases, that is even worse than the \texttt{A* Static} baseline. This suggests that retraining solely at the leaf level is inadequate for environments with complex obstacle configurations and few charging stations.

In contrast, the three hierarchical methods demonstrate strong performance, achieving success rates close to those of \texttt{A* Oracle}. These results validate the benefit of selectively retraining at higher levels in the hierarchy when necessary.

\subsubsection{Adaptation Time}

\paragraph{50\ x\ 50 Environment}

\begin{figure}[h]
    \centering
    \includegraphics[width=0.70\linewidth]{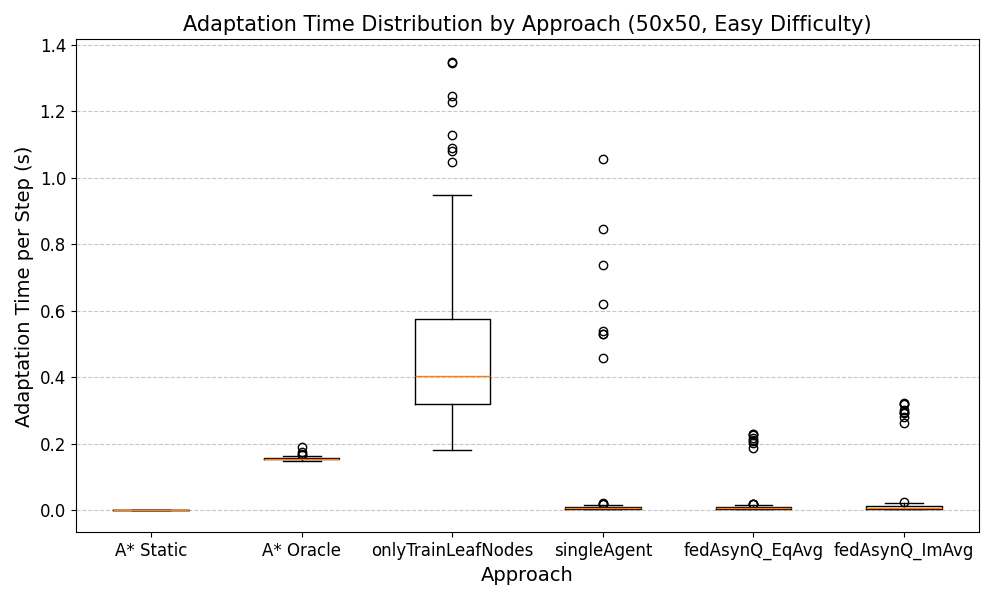}
    \caption{Box plots of adaptation times in a $50 \times 50$ easy environment.}
    \label{fig:box_50x50-easy}
\end{figure}

In the easy environment (Figure~\ref{fig:box_50x50-easy}), \texttt{A* Static} exhibits zero adaptation time and zero variance, as expected; it performs no replanning. \texttt{A* Oracle} has a stable adaptation time just below 0.2 seconds with negligible variance, since it performs the same global operations at every time step.

Among the learning-based methods, \texttt{onlyTrainLeafNodes} has the highest average adaptation time and the largest variance. This is due to its strategy of always retraining affected leaf nodes, regardless of the actual impact. In contrast, the three more sophisticated approaches retrain only when the success rate drops. This leads to fewer retraining events and thus lower average adaptation time and variance. These approaches still show outliers, likely due to time steps requiring retraining or escalation to higher tree levels.

\begin{figure}[h]
    \centering
    \includegraphics[width=0.70\linewidth]{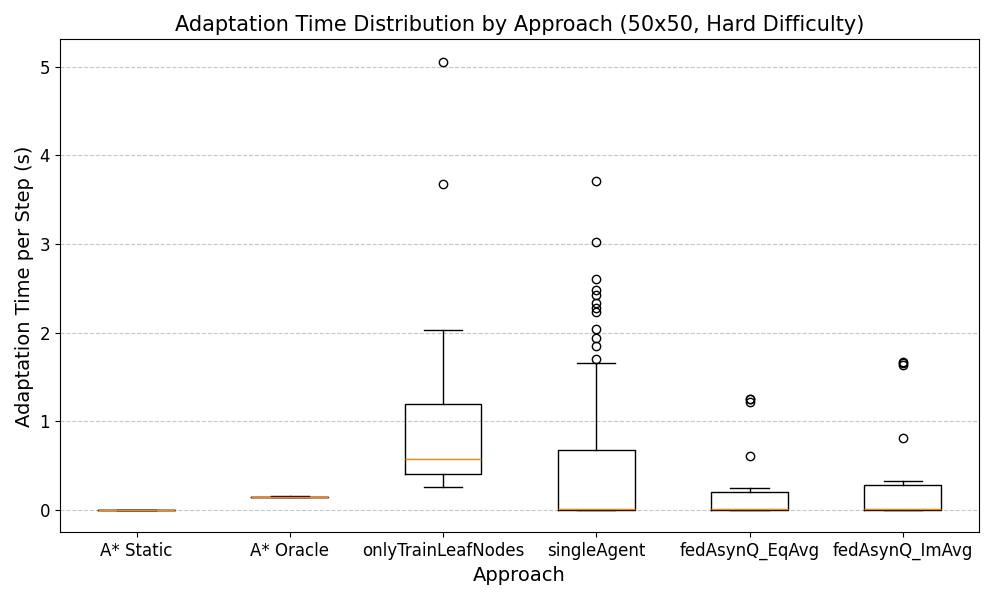}
    \caption{Box plots of adaptation times in a $50 \times 50$ hard environment.}
    \label{fig:box_50x50-hard}
\end{figure}

In the hard environment (Figure~\ref{fig:box_50x50-hard}), the three advanced approaches exhibit a more skewed distribution, indicating increased retraining or escalation to higher hierarchical levels, which increases adaptation time for specific time steps. Conversely, \texttt{onlyTrainLeafNodes} maintains more consistent behavior across difficulty levels, as its fixed retraining strategy yields similar box plot shapes.

\paragraph{300\ x\ 300 Environment}

\begin{figure}[h]
    \centering
    \includegraphics[width=0.70\linewidth]{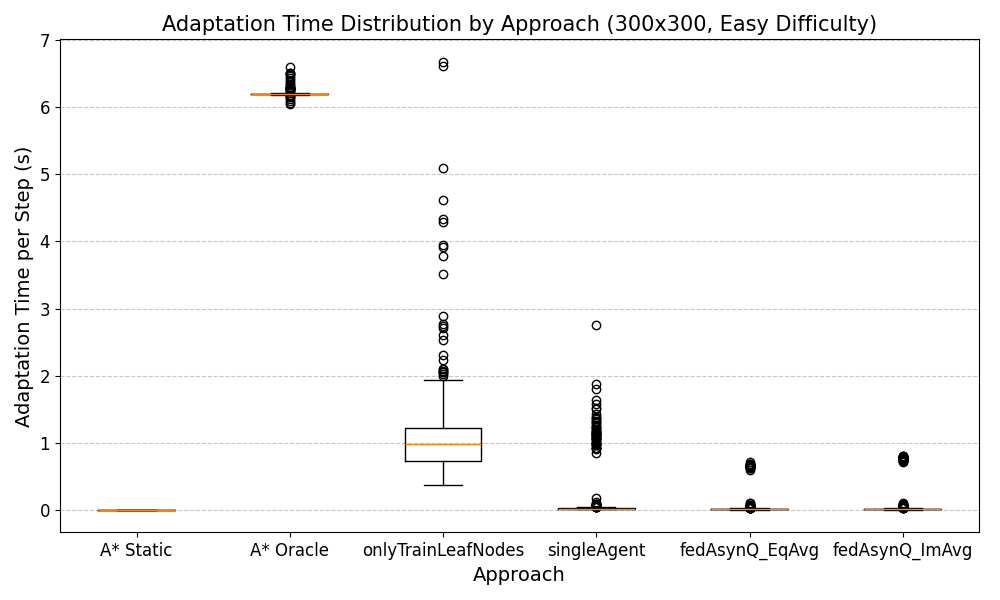}
    \caption{Box plots of adaptation times in a $300 \times 300$ easy environment.}
    \label{fig:box_300x300-easy}
\end{figure}

In the easy $300 \times 300$ environment (Figure~\ref{fig:box_300x300-easy}), \texttt{A* Oracle} shows a steep increase in adaptation time, averaging over six seconds per step, highlighting the scalability limitations of traditional methods. \texttt{A* Static} still performs no adaptation.

Among the learning-based methods, \texttt{onlyTrainLeafNodes} again has a higher average adaptation time than the other three approaches. The three advanced methods all perform similarly in average adaptation time, though with some notable outliers. The \texttt{singleAgent} approach shows more dispersed and higher outliers, some nearing three seconds, suggesting that certain retraining steps require more adaptation time. The \texttt{fedAsynQ\_EqAvg} and \texttt{fedAsynQ\_ImAvg} variants have tighter outlier distributions. Notably, \texttt{fedAsynQ\_EqAvg} has slightly lower outliers, likely due to its simpler averaging scheme that avoids computing importance weights.

\begin{figure}[h]
    \centering
    \includegraphics[width=0.70\linewidth]{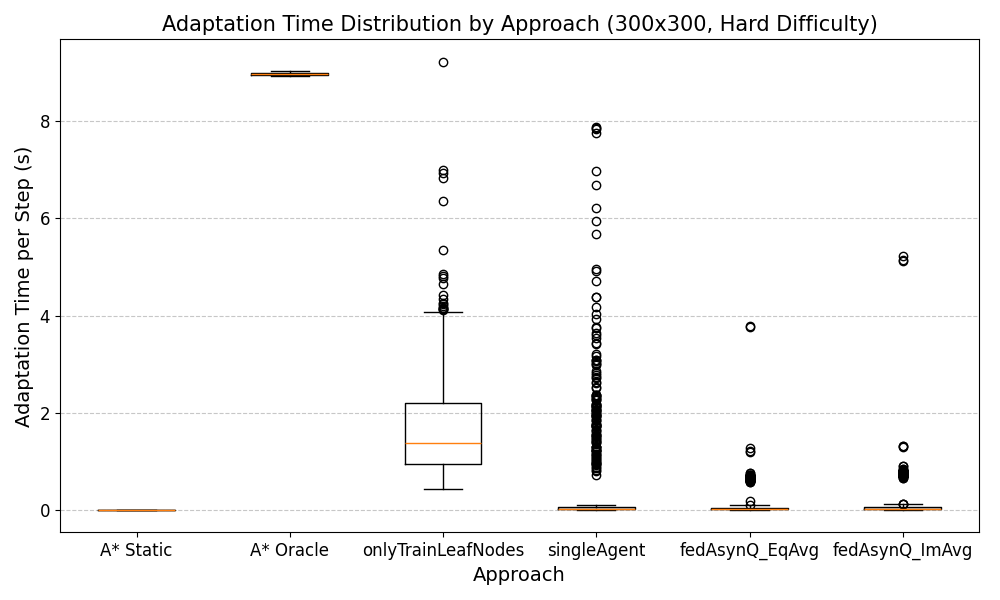}
    \caption{Box plots of adaptation times in a $300 \times 300$ hard environment.}
    \label{fig:box_300x300-hard}
\end{figure}

In the hard $300 \times 300$ environment (Figure~\ref{fig:box_300x300-hard}), the \texttt{A* Oracle} approach averages approximately nine seconds per time step. The three advanced approaches exhibit more outliers than their easy-environment counterparts, with \texttt{singleAgent} displaying the broadest spread. Conversely, the federated Q-learning variants demonstrate tighter and slightly lower outlier ranges.

\paragraph{Adaptation Time per Difficulty Level}

\begin{figure}[h]
    \centering
    \includegraphics[width=0.70\linewidth]{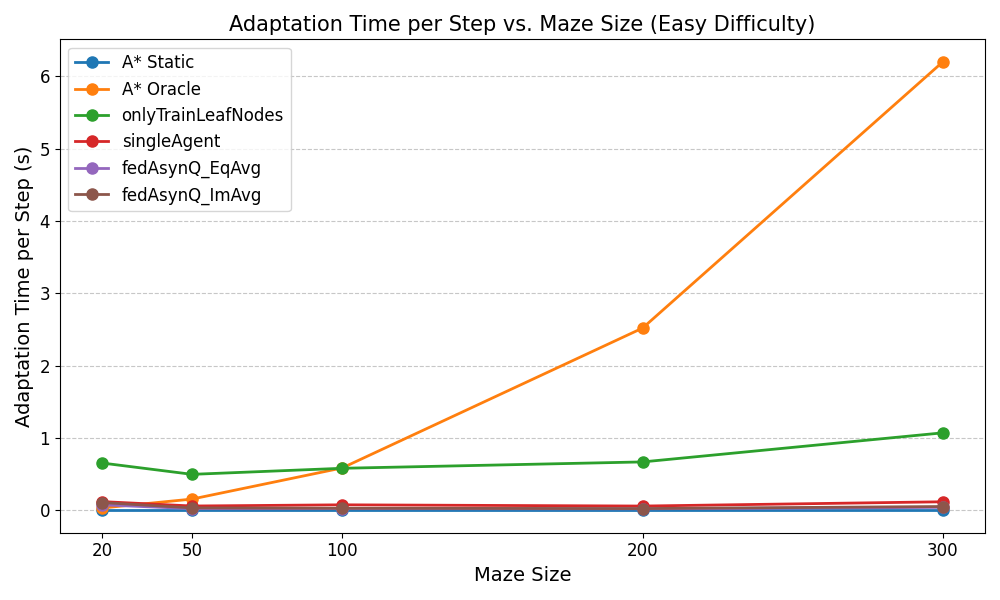}
    \caption{Adaptation time of different path planning approaches in easy environments.}
    \label{fig:evaluation-adaptation-time-easy}
\end{figure}

\begin{figure}[h]
    \centering
    \includegraphics[width=0.70\linewidth]{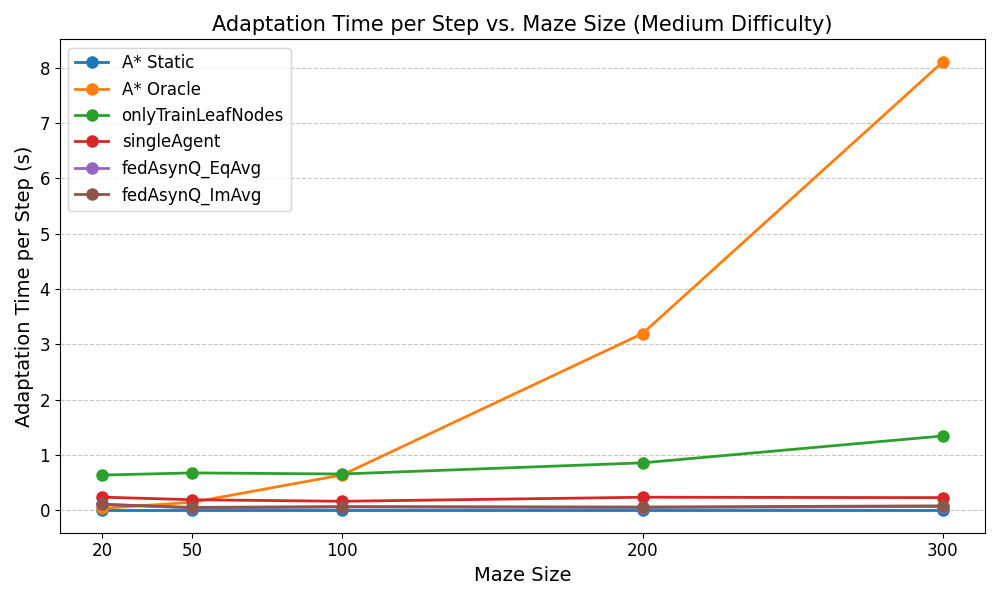}
    \caption{Adaptation time of different path planning approaches in medium environments.}
    \label{fig:evaluation-adaptation-time-medium}
\end{figure}

\begin{figure}[h]
    \centering
    \includegraphics[width=0.70\linewidth]{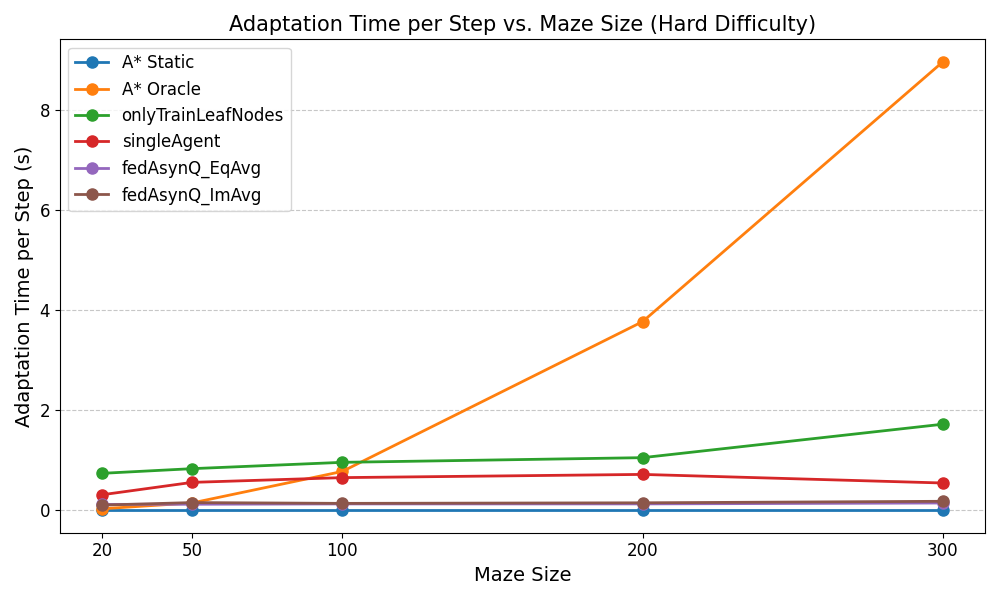}
    \caption{Adaptation time of different path planning approaches in hard environments.}
    \label{fig:evaluation-adaptation-time-hard}
\end{figure}

Figures~\ref{fig:evaluation-adaptation-time-easy},~\ref{fig:evaluation-adaptation-time-medium}, and~\ref{fig:evaluation-adaptation-time-hard} visualize the average adaptation time across all environment sizes for easy, medium, and hard complexity levels.

For easy environments (Figure~\ref{fig:evaluation-adaptation-time-easy}), the adaptation time for \texttt{A* Oracle} increases exponentially with the maze size, whereas the adaptation time for \texttt{A* Static} remains constant at zero. The \texttt{onlyTrainLeaf\allowbreak Nodes} approach exhibits higher average adaptation times across all environment sizes compared to the three advanced approaches, with a slight increase at the $300 \times 300$ size. The advanced approaches sustain low and stable adaptation times across all sizes.

For medium environments (Figure~\ref{fig:evaluation-adaptation-time-medium}), similar trends are observed, but a small gap emerges between the \texttt{singleAgent} approach (slightly higher adaptation time) and the two federated Q-learning approaches (\texttt{fedAsynQ\_EqAvg} and \texttt{fedAsynQ\_ImAvg}), which are more efficient.

For hard environments (Figure~\ref{fig:evaluation-adaptation-time-hard}), the differences between the \ac{RL}-based methods are most pronounced here. From highest to lowest average adaptation time, we have \texttt{onlyTrainLeafNodes}, \texttt{single\allowbreak Agent}, and \texttt{fedAsynQ\_ImAvg} with \texttt{fedAsynQ\_EqAvg}. This clearly shows the scalability and efficiency benefits of the federated Q-learning approaches under challenging conditions.

\subsubsection{Cumulative Adaptation Time}

\paragraph{50\ x\ 50 Environment}

\begin{figure}[h]
    \centering
    \includegraphics[width=0.70\linewidth]{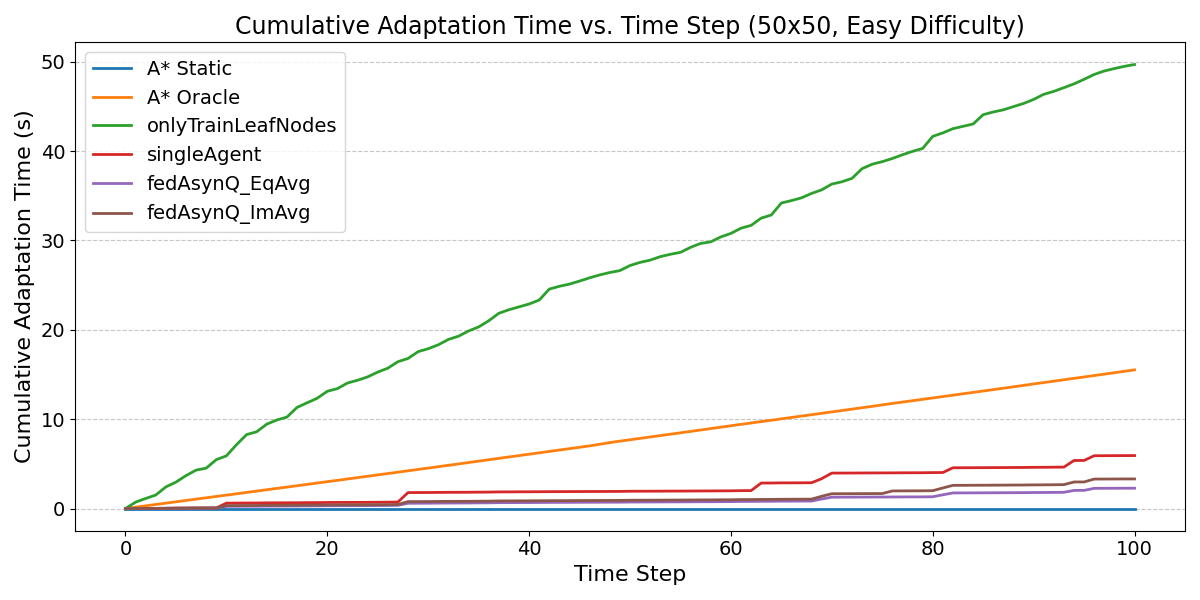}
    \caption{Cumulative adaptation times in a $50 \times 50$ easy environment.}
    \label{fig:cum_50x50-easy}
\end{figure}

In the easy setting (Figure~\ref{fig:cum_50x50-easy}), two straight lines emerge: a horizontal line for the \texttt{A* Static} approach and a slightly sloped line for the \texttt{A* Oracle} approach. The horizontal line reflects the fact that \texttt{A* Static} does not perform any replanning throughout the simulation. The linear slope of \texttt{A* Oracle} corresponds to its constant adaptation time at each time step, consistent with the earlier box plot analysis that showed almost zero variance in its adaptation time.

\begin{figure}[h]
    \centering
    \includegraphics[width=0.70\linewidth]{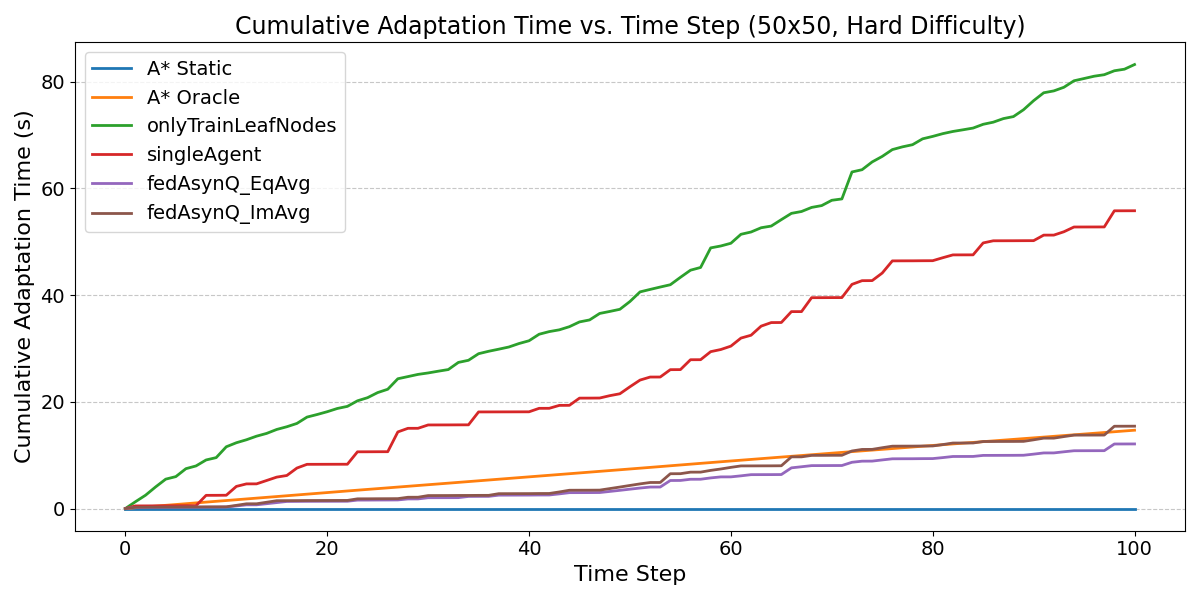}
    \caption{Cumulative adaptation times in a $50 \times 50$ hard environment.}
    \label{fig:cum_50x50-hard}
\end{figure}

The plot also demonstrates that the cumulative adaptation time for the \texttt{onlyTrainLeafNodes} approach exceeds that of \texttt{A* Oracle}. In contrast, the three more advanced methods exhibit significantly lower cumulative adaptation times, with \texttt{fedAsynQ\_EqAvg} achieving the lowest overall. This can be attributed to its simpler update step, which avoids recomputing importance weights for each Q-table entry. The \texttt{singleAgent} method has a slightly higher cumulative adaptation time than the federated approaches.

In the hard environment (Figure~\ref{fig:cum_50x50-hard}), the baseline methods behave the same as in the easy setting: \texttt{A* Static} remains flat, and \texttt{A* Oracle} shows a steady linear increase. However, the \ac{RL}-based methods now display more jagged curves, indicating more frequent retraining due to the higher environmental complexity. Notably, while the final cumulative adaptation time after 100 time steps remains the same for both baselines (approximately 0 seconds for \texttt{A* Static} and 15 seconds for \texttt{A* Oracle}), this is not true for the learning-based methods. In harder environments, their cumulative adaptation time is significantly higher.

\paragraph{300\ x\ 300 Environment}

\begin{figure}[h]
    \centering
    \includegraphics[width=0.70\linewidth]{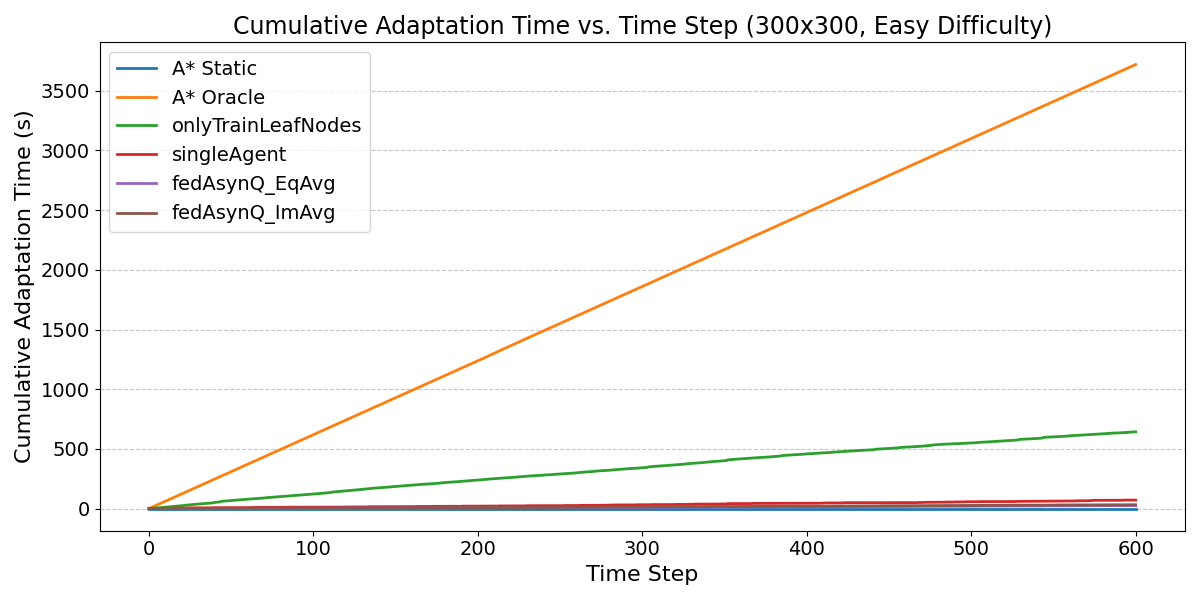}
    \caption{Cumulative adaptation times in a $300 \times 300$ easy environment.}
    \label{fig:cum_300x300-easy}
\end{figure}

\begin{figure}[h]
    \centering
    \includegraphics[width=0.70\linewidth]{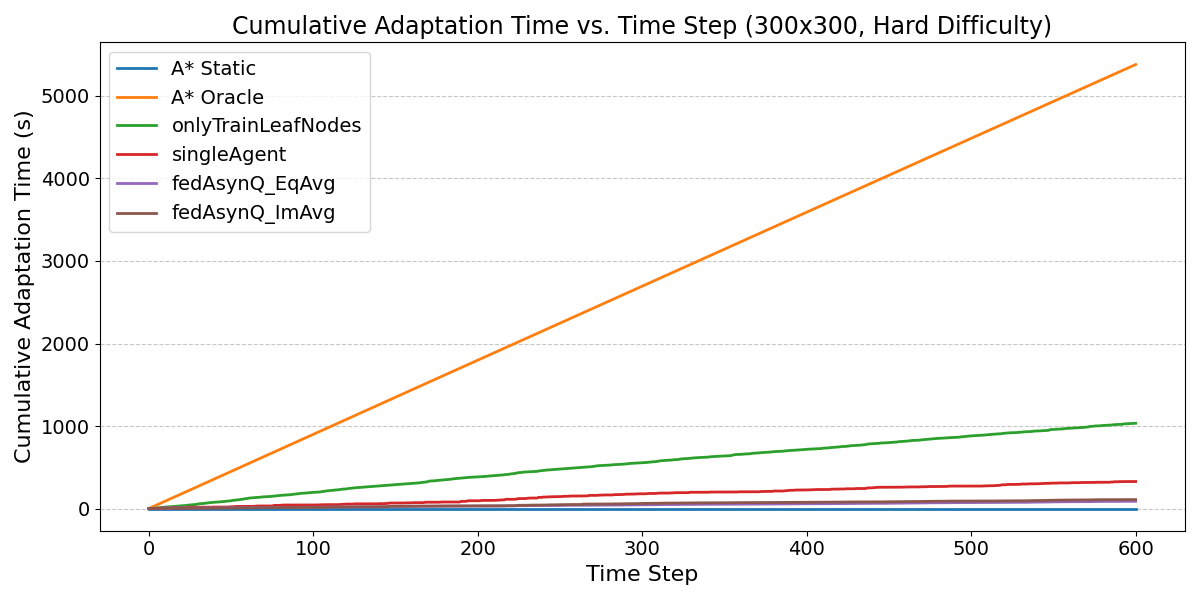}
    \caption{Cumulative adaptation times in a $300 \times 300$ hard environment.}
    \label{fig:cum_300x300-hard}
\end{figure}

Figures~\ref{fig:cum_300x300-easy} and~\ref{fig:cum_300x300-hard} present the cumulative adaptation times for both easy and hard $300 \times 300$ environments. 

The overall trends across both environmental complexities are similar, but the key differences are:

\begin{itemize}
    \item The total cumulative adaptation time is higher in hard environments than in easy ones, as increased complexity necessitates more frequent retraining. This applies to our learning-based approaches, but also to the \texttt{A* Oracle} approach, which is not based on a learning mechanism.
    \item The distinction among the \ac{RL}-based approaches is more evident in hard environments. Notably, the two federated Q-learning variants outperform others, achieving the lowest total cumulative adaptation times.
\end{itemize}

These results further reinforce the scalability and efficiency benefits of the \ac{RL}-based methods, particularly the federated Q-learning approaches.

\subsubsection{Average Path Length}

\paragraph{Easy Complexity}

\begin{figure}[h]
    \centering
    \includegraphics[width=0.70\linewidth]{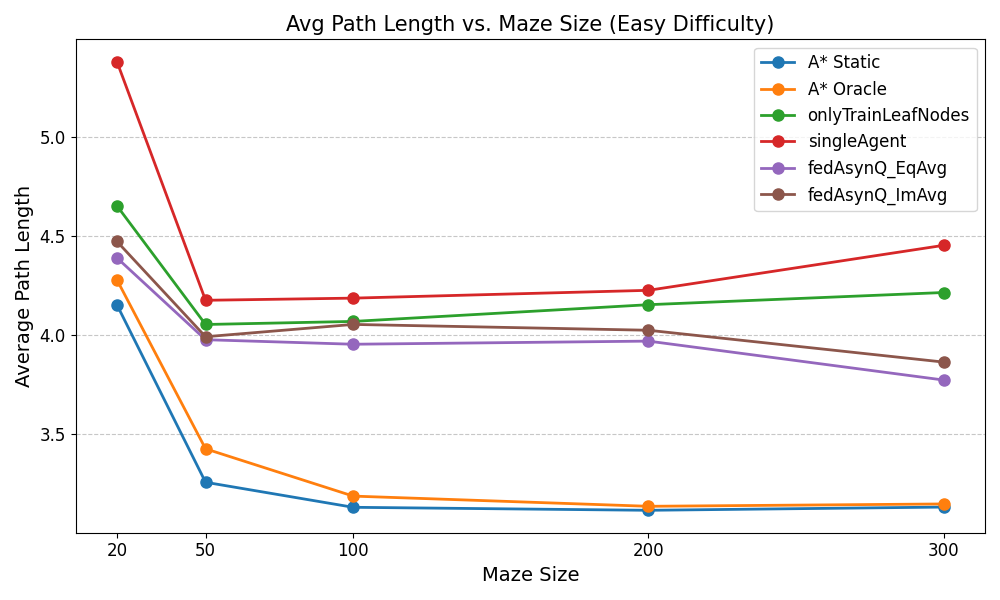}
    \caption{Average path length achieved by each approach in easy complexity environments.}
    \label{fig:evaluation-average-path-length-easy}
\end{figure}

In Figure~\ref{fig:evaluation-average-path-length-easy}, we see that both baseline methods (\texttt{A* Static} and \texttt{A* Oracle}) yield the shortest path lengths in easy environments. However, it is important to interpret these values in context. Achieving the shortest paths is only meaningful when paired with a high success rate.

\texttt{A* Oracle} meets both criteria, i.e., it consistently finds valid paths and guarantees minimal length. In contrast, \texttt{A* Static} has lower success rates, so its path lengths do not reliably reflect optimal performance.

Among the learning-based methods, the \texttt{singleAgent} approach produces slightly longer paths than the two federated Q-learning approaches. Between those, \texttt{fedAsynQ\_EqAvg} tends to find slightly shorter paths on average than \texttt{fedAsynQ\_ImAvg}, particularly in larger environments.

\paragraph{Medium Complexity}

For environments of medium complexity (Figure~\ref{fig:evaluation-average-path-length-medium}), the trends remain consistent:

\begin{figure}[h]
    \centering
    \includegraphics[width=0.70\linewidth]{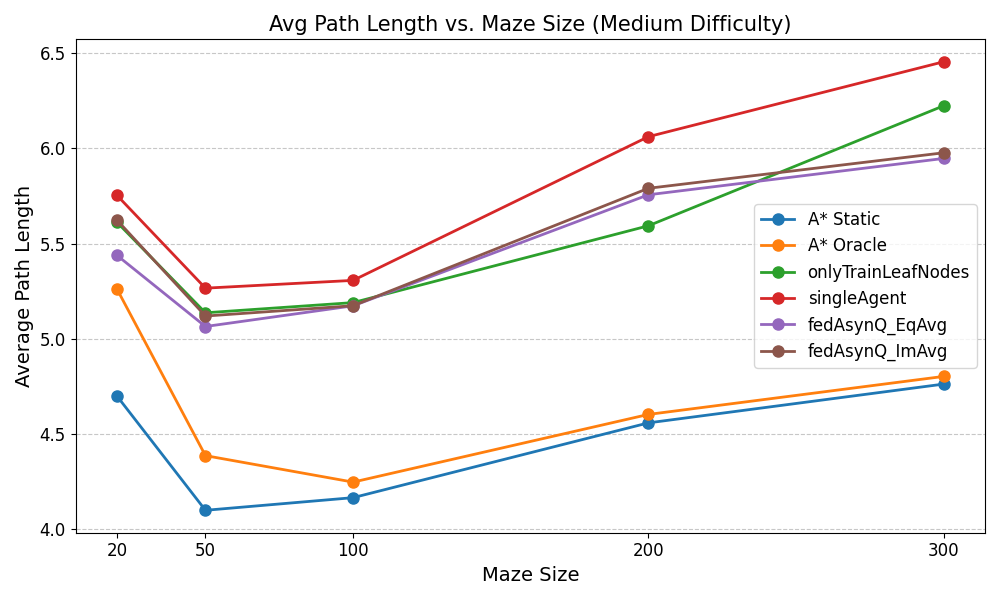}
    \caption{Average path length achieved by each approach in medium complexity environments.}
    \label{fig:evaluation-average-path-length-medium}
\end{figure}

\begin{itemize}
    \item Both baselines maintain the shortest path lengths.
    \item The \texttt{singleAgent} approach again finds the longest paths among the learning-based methods.
    \item The two federated Q-learning methods find more optimal routes, with \texttt{fedAsynQ\_EqAvg} slightly outperforming \texttt{fedAsynQ\_ImAvg}.
    \item Interestingly, the \texttt{onlyTrainLeafNodes} method produces path lengths similar to the federated Q-learning approaches, but since it has a lower overall success rate, the reliability of its results is diminished.
\end{itemize}

\paragraph{Hard Complexity}

\begin{figure}[h]
    \centering
    \includegraphics[width=0.70\linewidth]{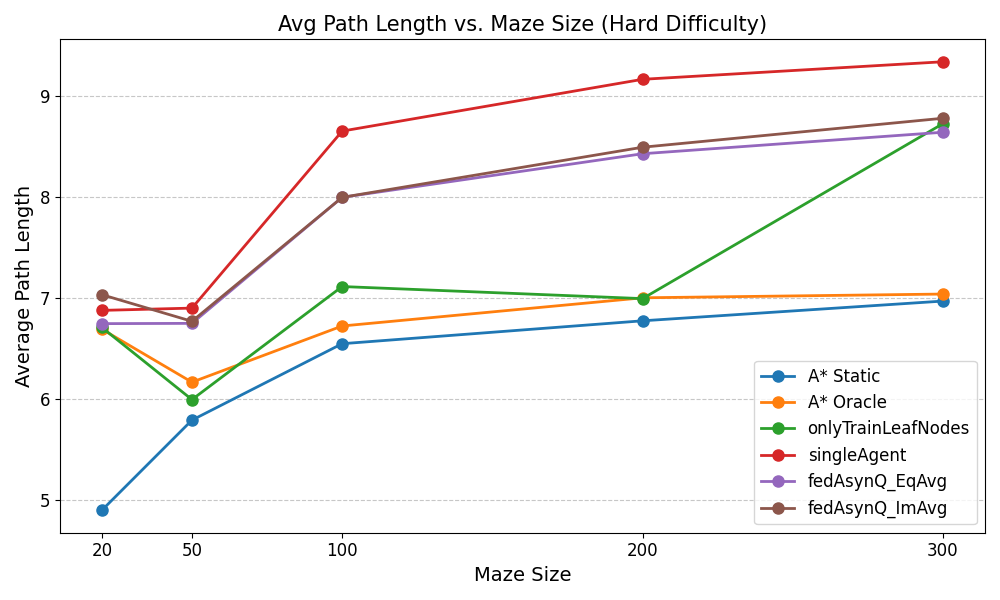}
    \caption{Average path length achieved by each approach in hard complexity environments.}
    \label{fig:evaluation-average-path-length-hard}
\end{figure}

In the hard environments (Figure~\ref{fig:evaluation-average-path-length-hard}), the two baselines again produce the shortest paths, with \texttt{A* Oracle} continuing to deliver optimal results.

Surprisingly, \texttt{onlyTrainLeafNodes} yields the shortest path lengths among all \ac{RL}-based methods. However, as shown in Figure~\ref{fig:evaluation-accuracy-hard}, this approach has a much lower success rate in these hard environments, meaning that the paths it does find are not representative of the overall quality of the solution.

Among the remaining methods, we observe that \texttt{singleAgent} continues to produce the longest paths, while the two federated approaches perform similarly, with \texttt{fedAsynQ\_EqAvg} having a slight edge in larger environments.

\subsubsection{Initial Training Time}

\paragraph{Easy Complexity}

\begin{figure}[h]
    \centering
    \includegraphics[width=0.70\linewidth]{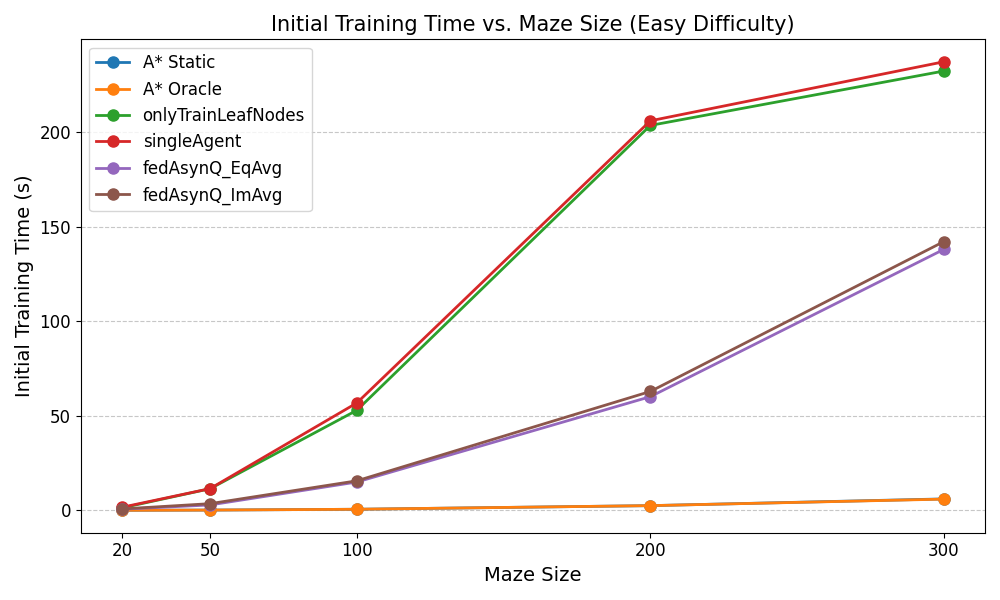}
    \caption{Initial training time of each approach in easy complexity environments.}
    \label{fig:evaluation-initial-training-time-easy}
\end{figure}

In easy environments (Figure~\ref{fig:evaluation-initial-training-time-easy}), we observe that the initial training times naturally group into three distinct categories.

The first group comprises the two baseline methods (\texttt{A* Static} and \texttt{A* Oracle}), which require the same initial training time. This is expected, as both follow the same procedure to perform path planning on the initial environment. Across all environment sizes, the initial training time is extremely low compared to the two other groups. 

The second group consists of the \texttt{fedAsynQ\_EqAvg} and \texttt{fedAsynQ\_ImAvg} approaches, with the former slightly outperforming the latter. As can be observed, the initial training time increases exponentially with the environment size.

Although the second group performs worse than the first group (baselines) in terms of initial training time, they still outperform the third group, comprising the \texttt{onlyTrainLeafNodes} and \texttt{singleAgent} approaches. Both methods require significantly longer initial training time, compared to the methods in the first and second groups.

\paragraph{Medium Complexity}

\begin{figure}[h]
    \centering
    \includegraphics[width=0.70\linewidth]{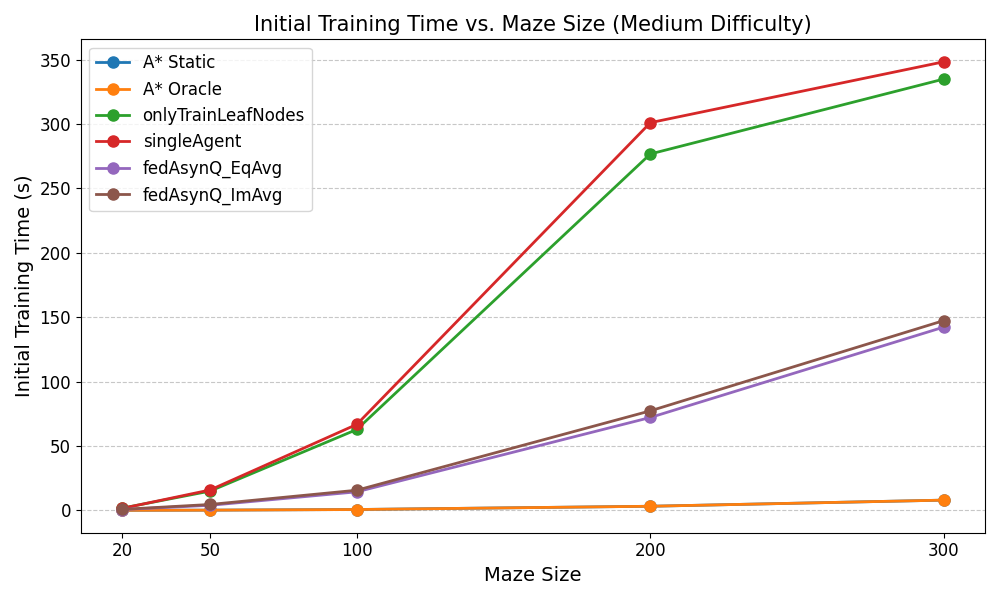}
    \caption{Initial training time of each approach in medium complexity environments.}
    \label{fig:evaluation-initial-training-time-medium}
\end{figure}

In environments of medium complexity (Figure~\ref{fig:evaluation-initial-training-time-medium}), the gap between the federated Q-learning and single-agent Q-learning approaches becomes more pronounced. The single-agent methods require approximately 350 seconds of initial training time in a $300 \times 300$ environment, while the federated Q-learning approaches need only around 150 seconds. This represents a significant improvement and shows the benefits of parallelized learning in a federated setup.

Both baselines (\texttt{A* Static} and \texttt{A* Oracle}) continue to require significantly less initial training time than the other two groups.

\paragraph{Hard Complexity}

\begin{figure}[h]
    \centering
    \includegraphics[width=0.70\linewidth]{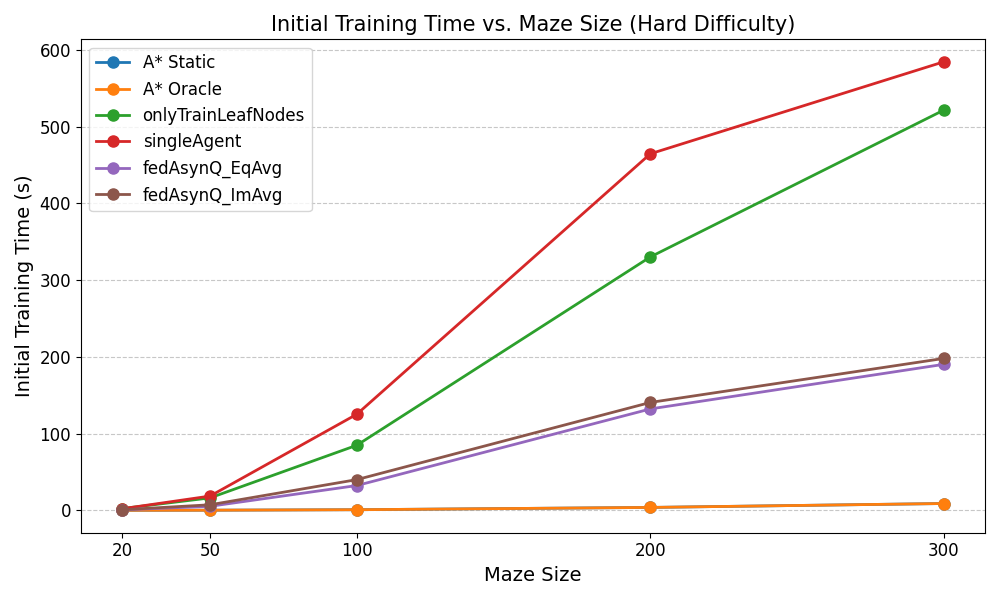}
    \caption{Initial training time of each approach in hard complexity environments.}
    \label{fig:evaluation-initial-training-time-hard}
\end{figure}

In hard environments (Figure~\ref{fig:evaluation-initial-training-time-hard}), the trends remain more or less consistent. Both baselines still exhibit the lowest initial training times. The two federated Q-learning approaches again perform well, requiring only about 200 seconds of initial training time in the $300 \times 300$ environment.

The key difference lies in the relative performance of the \texttt{onlyTrainLeafNodes} and \texttt{singleAgent} approaches. Unlike in easier environments, \texttt{onlyTrainLeafNodes} now requires less initial training time than \texttt{singleAgent}, which demands close to 600 seconds for a hard $300 \times 300$ environment. This difference arises because \texttt{singleAgent} escalates training to higher levels of the hierarchy when local policies are ineffective, whereas \texttt{onlyTrainLeafNodes} restricts training to the lowest level.

\subsection{Edge Case Results}

\subsubsection{Accuracy}

\begin{figure}[h]
    \centering
    \includegraphics[width=0.70\linewidth]{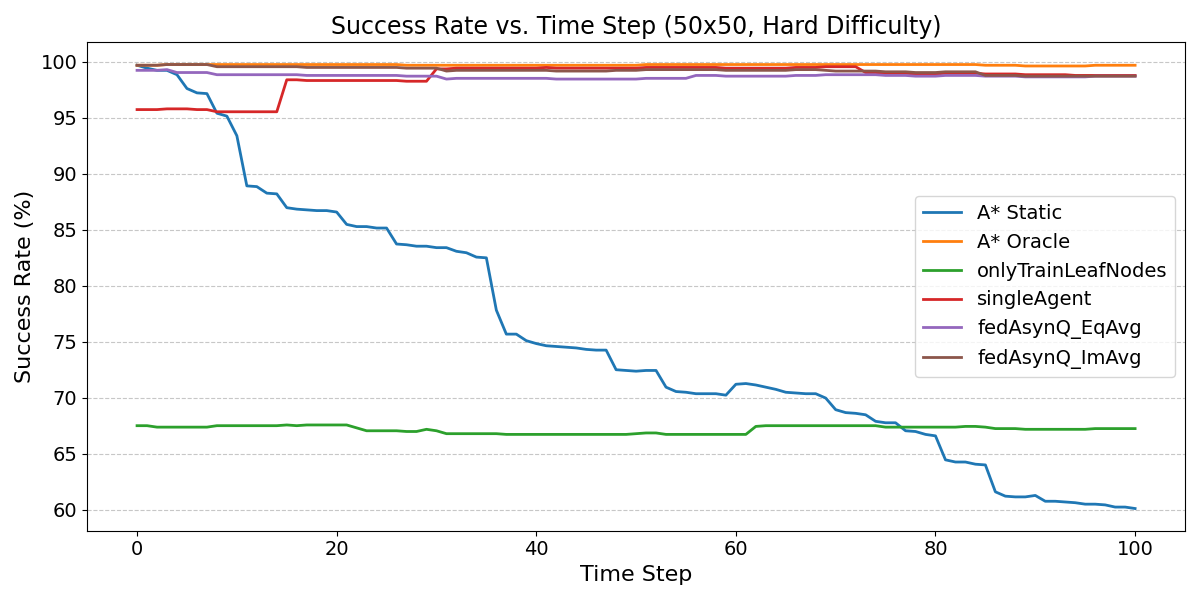}
    \caption{Success rate over time for each approach in the edge case environment.}
    \label{fig:accuracy-edge-case}
\end{figure}

We start by discussing the accuracy of the different approaches on the edge case example. As shown in Figure~\ref{fig:accuracy-edge-case}, this plot differs from earlier figures in that it focuses on a single environment: a hard $50 \times 50$ maze. This allows us to examine the evolution of accuracy over time in more detail, rather than examining average accuracy levels across all sizes and difficulties.

The graph displays the success rate of each of the six approaches, two baselines and four \ac{RL}-based methods, across simulation time steps, ranging from $t = 0$ to $t = 100$ (i.e., $2 \times 50$). As expected, the \texttt{A* Static} baseline shows a steady decline in accuracy, starting near 100\% and dropping to around 60\%. This behavior reflects the underlying mechanism of \texttt{A* Static}: it does not update its paths after initial planning, so its effectiveness decreases as more obstacle changes accumulate.

The \texttt{A* Oracle} approach maintains the highest accuracy throughout all time steps, remaining close to 100\%, which is consistent with its idealized design that assumes perfect, real-time knowledge of the environment.

The \texttt{onlyTrainLeafNodes} approach maintains a consistent accuracy around 67-68\%, which is notably lower than the other \ac{RL}-based methods. This is expected, as the method limits retraining to only the smallest sub-environments (leaf nodes), which is insufficient in challenging scenarios such as this one.

In contrast, the three remaining approaches (\texttt{singleAgent}, \texttt{fedAsynQ\_EqAvg}, and \texttt{fedAsynQ\_ImAvg}) achieve significantly higher accuracy over time. Although \texttt{singleAgent} starts with slightly lower accuracy in the early time steps, it catches up around $t = 20$, matching the performance of the federated Q-learning approaches. Both federated methods maintain high success rates throughout the simulation, nearly matching \texttt{A* Oracle}.

\subsubsection{Adaptation Time}

\begin{figure}[h]
    \centering
    \includegraphics[width=0.70\linewidth]{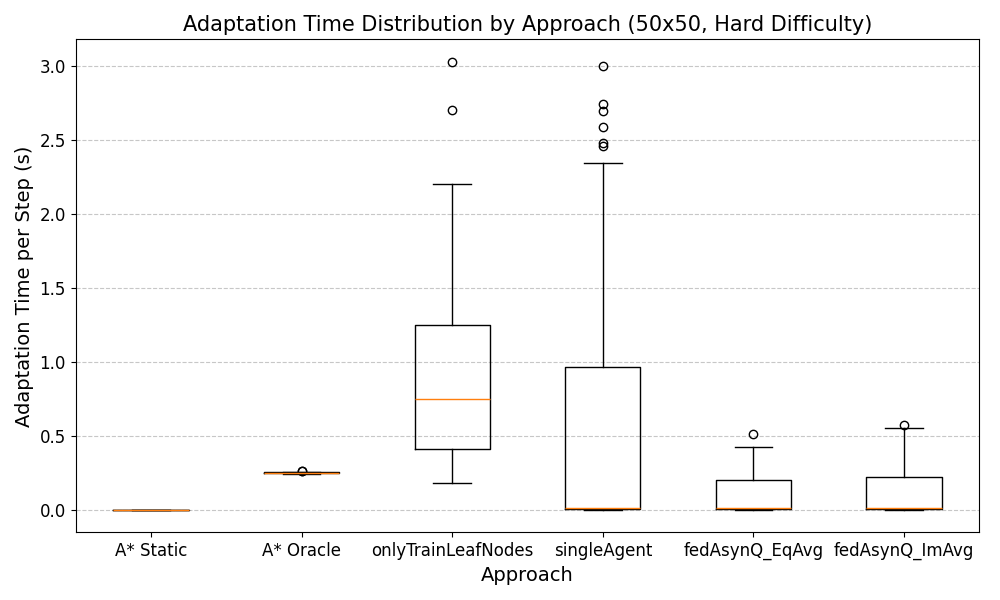}
    \caption{Adaptation time of the different approaches in the edge case environment.}
    \label{fig:adaptation-time-edge-case}
\end{figure}

Figure~\ref{fig:adaptation-time-edge-case} shows the adaptation time for each approach in the edge case environment. When comparing the adaptation time box plots to those from the regular hard $50 \times 50$ environment (Figure~\ref{fig:box_50x50-hard}), we observe that the shapes of the box plots are quite similar across all approaches. The primary difference lies in the outliers, which are slightly less pronounced in the edge case environment, though the whiskers of the \ac{RL}-based approaches appear more extended.

\subsubsection{Cumulative Adaptation Time}

\begin{figure}[h]
    \centering
    \includegraphics[width=0.70\linewidth]{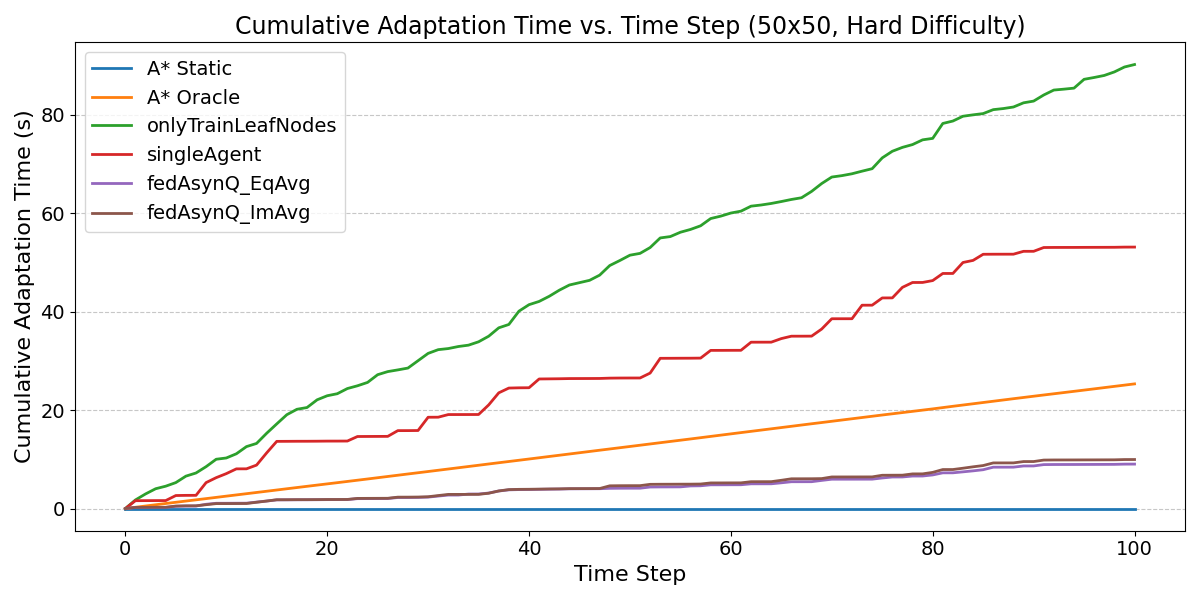}
    \caption{Cumulative adaptation time of the different approaches in the edge case environment.}
    \label{fig:cumulative-adaptation-time-edge-case}
\end{figure}

The cumulative adaptation times in the edge case environment (Figure~\ref{fig:cumulative-adaptation-time-edge-case}) exhibit patterns similar to those in the hard $50 \times 50$ environment (Figure~\ref{fig:cum_50x50-hard}). Specifically, the \texttt{onlyTrainLeafNodes} and \texttt{singleAgent} approaches incur higher cumulative adaptation times than \texttt{A* Oracle}, whereas both federated Q-learning variants demonstrate greater efficiency. Notably, these variants achieve lower cumulative adaptation times than \texttt{A* Oracle} in the edge case environment, unlike the original hard $50 \times 50$ environment, where their times are comparable.

These results are somewhat counterintuitive. One would expect that in a more challenging environment, such as this edge case, the adaptation and cumulative adaptation times would be higher. However, the results suggest otherwise. A likely explanation lies in the randomized nature of obstacle changes during simulation. If most changes occur in regions that already contain accessible charging stations, the disadvantage of the top-left quadrant lacking a charging station becomes less impactful. If we were to restrict changes in the environment to that quadrant specifically, retraining at higher levels in the hierarchy would be triggered more frequently, resulting in longer adaptation times. Restricting environmental changes to only specific regions in the environment is not implemented, as it is considered less realistic.

\subsubsection{Average Path Length}

\begin{figure}[h]
    \centering
    \includegraphics[width=0.70\linewidth]{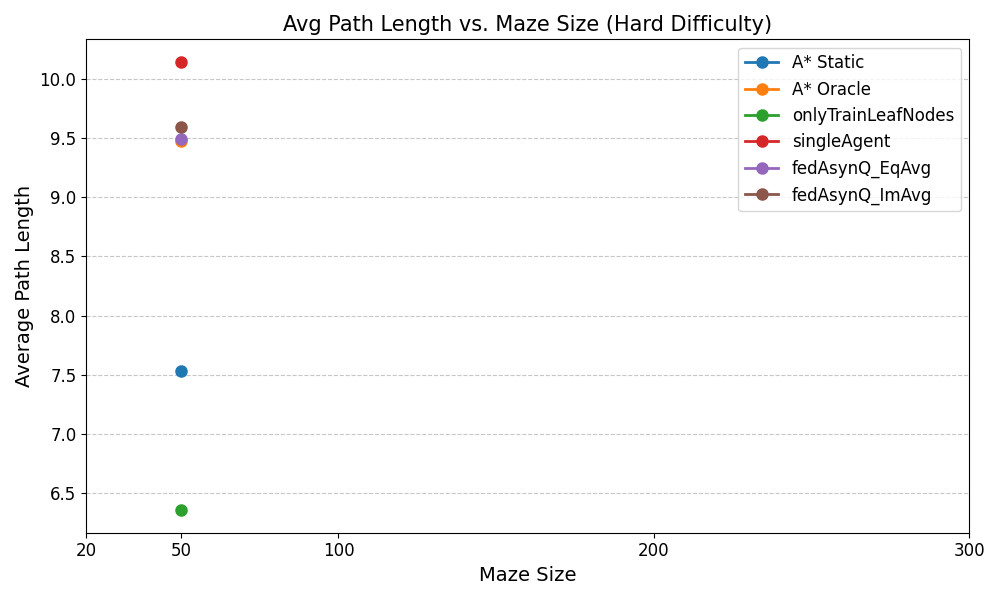}
    \caption{Average path length of the different approaches in the edge case environment.}
    \label{fig:path-length-edge-case}
\end{figure}

Figure~\ref{fig:path-length-edge-case} shows the average path length for each approach in the edge case scenario. Since only one $50 \times 50$ environment was tested, each method is represented by a single data point.

The ideal baseline, \texttt{A* Oracle}, achieves an average path length of approximately 9.5 steps. The federated Q-learning approaches (\texttt{fedAsynQ\_EqAvg} and \texttt{fedAsynQ\_ImAvg}) produce paths that are very close in length to this optimal value. The \texttt{singleAgent} approach yields slightly longer paths, while the \texttt{A* Static} and \texttt{onlyTrainLeafNodes} approaches produce significantly shorter paths. However, their low success rates explain this misleadingly low average path length.

Comparing these results to those of the original hard $50 \times 50$ environment in Figure~\ref{fig:evaluation-average-path-length-hard}, we observe that the average path lengths in the edge case are substantially higher. For example, \texttt{A* Oracle} exhibits an increase of approximately 3.5 steps per path. This clearly illustrates that the edge case is more challenging, particularly because the top-left quadrant lacks charging stations, requiring longer paths from that region to reach a charging station.

\subsubsection{Initial Training Time}

\begin{figure}[h]
    \centering
    \includegraphics[width=0.70\linewidth]{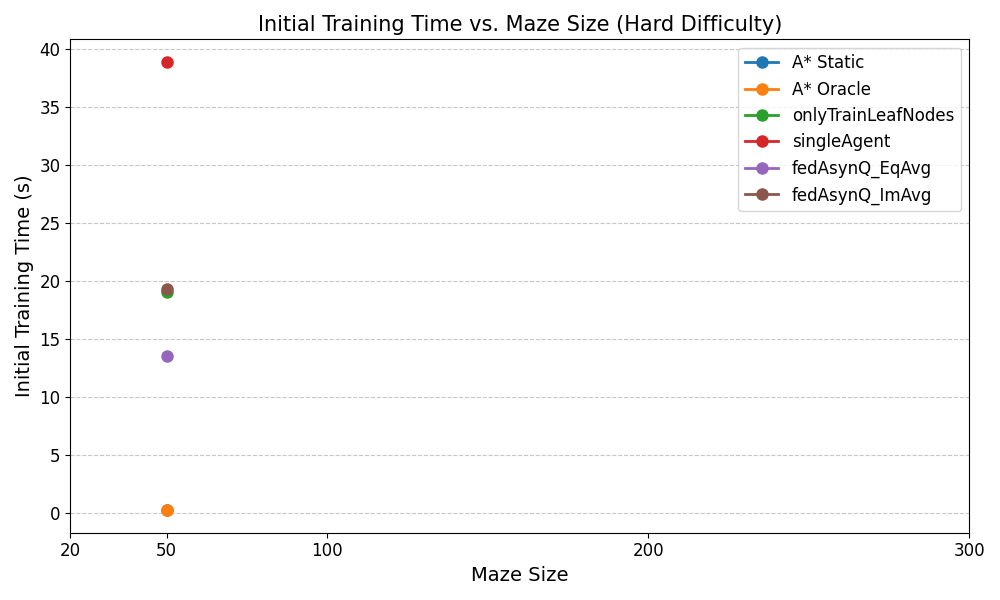}
    \caption{Initial training time of the different approaches in the edge case environment.}
    \label{fig:initial-training-time-edge-case}
\end{figure}

Figure~\ref{fig:initial-training-time-edge-case} shows the initial training time for each approach in the edge case environment. As observed in previous results, the two baseline methods achieve the lowest initial training times, since they assume full knowledge of the environment from the start. Among the \ac{RL}-based methods, the \texttt{singleAgent} approach performs the worst, requiring approximately 40 seconds of initial training time. Of the remaining approaches, \texttt{fedAsynQ\_EqAvg} achieves the best performance, completing initial training in just under 15 seconds. The \texttt{fedAsynQ\_ImAvg} approach follows, requiring approximately 20 seconds. While both are relatively efficient, this indicates that \texttt{fedAsynQ\_EqAvg} has a slight advantage in initial training time for this edge case.

\clearpage

\section{Discussion} \label{sec:discussion}

This section provides a high-level reflection on the results, connecting the identified limitations, proposed methodology, and experimental outcomes. We begin in Section~\ref{sec:performance-summary} with a summary of the performance of the six tested path planning approaches, highlighting key insights from the simulations. Section~\ref{sec:discussion-edge-case} then discusses the behavior of the approaches in the edge case environment. In Section~\ref{sec:positioning-marl}, we position our methodology within the landscape of \ac{MARL} paradigms. In Section~\ref{sec:addressing-limitations}, we examine how the proposed methodology addresses the limitations identified in the work of Yarahmadi et al., while Section~\ref{sec:methodology-limitations} outlines the remaining limitations of our own methodology. Finally, Section~\ref{sec:threats-validity} discusses potential threats to the validity of our methodology and experiments.

\subsection{Performance Summary per Approach} \label{sec:performance-summary}

Based on the simulation results presented in Section~\ref{sec:simulation-results}, we summarize the performance of the six tested path planning approaches as follows:

\begin{itemize}
    \item \textbf{A* Static:} This non-adaptive baseline performs no replanning during the simulation. It consistently yields the lowest accuracy levels of all approaches, while achieving short path lengths and low initial training times. However, due to its inability to respond to environmental changes, its results are unreliable in dynamic settings, and it cannot be considered a dynamic path planning approach.
    \item \textbf{A* Oracle:} This baseline consistently achieves the highest accuracy, shortest paths, and lowest initial training times (similar to \texttt{A* Static}), serving as an idealized benchmark. However, adapting to environmental changes in large environments is highly inefficient as replanning needs to be performed on the entire environment, leading to high adaptation times and poor scalability. Its assumption of perfect, real-time knowledge of the environment is unrealistic in dynamic scenarios, where uncertainty is inherent. Therefore, this approach is practically unusable and only serves as a theoretically optimal baseline to which we compare our other approaches.
    \item \textbf{onlyTrainLeafNodes:} This approach performs well in easy environments, realizing high levels of accuracy, but struggles in medium and hard settings. In small environments, it exhibits the highest adaptation times of all six approaches, while in larger environments, it achieves adaptation times lower than the \texttt{A* Oracle} baseline but worse than the federated Q-learning approaches. Despite producing path lengths comparable to more advanced methods and even to \texttt{A* Oracle} in certain hard environments, its low success rate undermines its significance. Its poor performance in complex environments reveals that retraining only at the leaf level (i.e., the smallest sub-environments) is insufficient and that a broader spatial context is needed for effective adaptation. The approach also exhibits high initial training times, which are improved by the federated Q-learning approaches.
    \item \textbf{singleAgent:} This method maintains high accuracy across all difficulty levels and demonstrates low average adaptation times. However, its abundance of outliers indicates inconsistent adaptation to changes across time steps, which are particularly prominent in hard environments. It produces reasonable path lengths across all environmental difficulties, but leaves room for further improvement. A significant drawback of the approach is that it requires a considerable amount of initial training time, particularly in larger environments, which is a limitation it shares with the \texttt{onlyTrainLeafNodes} approach.
    \item \textbf{fedAsynQ\_EqAvg:} This federated Q-learning approach consistently achieves high accuracy levels (comparable to \texttt{A* Oracle}) and low adaptation times across all environment sizes and difficulties. It also minimizes the spread of outliers present in the box plots of the adaptation time, suggesting that it exhibits more consistent adaptation across time steps. Among the three advanced path planning approaches, it produces the shortest paths across all three environmental difficulties, being slightly shorter than the paths generated by the \texttt{fedAsynQ\_ImAvg} approach, while still achieving similar levels of accuracy.
    \item \textbf{fedAsynQ\_ImAvg:} This method closely matches the performance of \texttt{fedAsynQ\_EqAvg}, though with slightly higher adaptation times and marginally longer paths across all environment difficulties. Nonetheless, it consistently outperforms \texttt{singleAgent}, further reinforcing the advantages of federated Q-learning via parallelized training. Moreover, both federated learning approaches significantly outperform the \texttt{onlyTrainLeafNodes} and \texttt{singleAgent} approaches in terms of initial training time.

\end{itemize}

\subsection{Discussion of the Edge Case} \label{sec:discussion-edge-case}

We briefly discuss the performance of our approaches on the edge case example. The \texttt{onlyTrainLeaf\allowbreak Nodes} approach performs poorly, achieving low accuracy levels and failing to find valid paths for many positions, particularly those located in the top-left quadrant that do not contain any charging stations. It also exhibits the highest adaptation times among the four \ac{RL}-based approaches. The \texttt{singleAgent} method performs somewhat better, with slightly lower adaptation times and higher accuracy. However, the paths it finds are longer than those produced by the optimal \texttt{A* Oracle} baseline.

Both federated Q-learning approaches demonstrate strong performance in this challenging environment. They maintain high accuracy throughout the entire simulation, require relatively low adaptation time, and produce path lengths that closely match those of \texttt{A* Oracle}. Furthermore, they achieve considerably lower initial training times than the \texttt{onlyTrainLeafNodes} approach, and especially outperform the \texttt{singleAgent} approach in this regard.

Overall, we conclude that the federated Q-learning approaches significantly outperform the other \ac{RL}-based methods and are the most effective solutions for path planning in dynamic environments of varying size and complexity.

\subsection{Positioning within MARL Paradigms} \label{sec:positioning-marl}

To position our methodology within the broader \ac{MARL} landscape, we distinguish between the two learning approaches employed: single-agent Q-learning and federated Q-learning. While both share the same decomposition framework and retraining strategy, they differ in how learning is carried out within each sub-environment.

The single-agent Q-learning approaches follow a structure similar to that of Yarahmadi et al.~\cite{yarahmadi2024comp}. Here, a single agent interacts with its sub-environment, learning a local policy independently, without access to the observations or actions of other agents. This setup corresponds to the \ac{DTDE} paradigm.

In contrast, the federated Q-learning approaches involve multiple agents in the same sub-environment to accelerate policy learning. Although the overall environment is still handled in a distributed manner through separate sub-environments, the learning process within each sub-environment reflects the \ac{CTDE} paradigm: the information from all agents is aggregated during training to optimize the shared policy, while execution remains decentralized, with each agent acting independently.

\subsection{Addressing the Identified Limitations} \label{sec:addressing-limitations}

This section reflects on how our methodology and results address the key limitations identified in the work of Yarahmadi et al. Rather than restating the problems in detail, we focus on showing how our contributions overcome them and how the evaluation supports this.  

\begin{itemize}
    \item \textbf{Dependence on the global planner:} We entirely removed the reliance on a global path planner, which is impractical in unknown environments. Instead, all planning is handled locally via Q-learning. We examined both single-agent Q-learning and federated Q-learning, the latter aggregating multiple agents’ Q-tables to accelerate learning. Results show that both methods perform well, but federated Q-learning achieves higher accuracy and faster adaptation, demonstrating that local \ac{RL}-based planning alone is sufficient and effective.
    \item \textbf{Naive replanning strategy:} The method proposed by Yarahmadi et al. replans whenever a sub-environment changes, regardless of the impact. We addressed this by introducing the retraining condition, which evaluates policy effectiveness after changes and decides whether retraining is necessary. This reduces redundant retraining and improves adaptation times without significantly compromising accuracy. Our federated Q-learning approaches (\texttt{fedAsynQ\_EqAvg}, \texttt{fedAsynQ\_ImAvg}) consistently maintain high accuracy across simulations, as shown for example in Figure~\ref{fig:accuracy-edge-case}.
    \item \textbf{Restricted agent scope:} To overcome the limited scope of agents restricted to isolated sub-environments, we introduced the hierarchical decomposition framework. This allows retraining at different levels of the hierarchical tree, enabling broader spatial awareness when needed. The results confirm its effectiveness: approaches that exploit the hierarchy adapt well to changes in complex environments, whereas approaches limited to the leaf nodes (\texttt{onlyTrainLeafNodes}) struggle, particularly in the edge case scenario.
    \item \textbf{Simplistic environment design:} We expanded the evaluation to environments of varying complexity (easy, medium, hard) and introduced a challenging edge case ($50 \times 50$ maze with one quadrant lacking charging stations). Our advanced approaches (\texttt{singleAgent}, \texttt{fedAsynQ\_EqAvg}, \texttt{fedAsynQ\_ImAvg}) achieve strong performance across all difficulties, showing robustness under more diverse and realistic conditions.
    \item \textbf{Single obstacle changes:} Finally, instead of simulating a single obstacle change per time step, our setup allows up to 10 simultaneous changes, with probabilities decreasing as the number of changes increases. Despite this greater realism, our most advanced approaches maintain low adaptation times and high accuracy levels.
\end{itemize}

\clearpage

\subsection{Limitations of our Methodology} \label{sec:methodology-limitations}

Despite the strengths of our proposed methodology, several limitations remain:

\begin{itemize}
    \item \textbf{Suboptimal path lengths:} Our best-performing approaches based on federated Q-learning consistently produce valid paths to charging stations, but these paths are often longer than those generated by the \texttt{A* Oracle} baseline. The gap arises because the learned policies do not always lead to optimal path lengths. Increasing the number of agent-environment interactions could improve policy quality, as more transitions would refine the Q-table and bring it closer to an optimal solution. However, this would slow down adaptation and reduce efficiency. Thus, we identify a trade-off between path optimality and adaptation speed. In line with the demands of dynamic environments, our focus prioritizes fast adaptation over exact optimality.
    \item \textbf{Initial training overhead:} Learning policies in previously unexplored environments requires substantial initial training. While training is parallelized across leaf sub-environments, the computational cost remains high in large environments (e.g., $300 \times 300$). This is an inherent limitation of tabular \ac{RL}, which depends on extensive exploration to converge to effective policies. Although federated Q-learning reduces training time compared to single-agent Q-learning, the overall initial training cost remains considerable.
    \item \textbf{Hierarchical decomposition constraints:} Our hierarchical decomposition assumes that the environment’s shape and dimensions are known in advance and are rectangular or square. As a result, the method is restricted to grid-like environments. Consequently, irregular or non-rectangular environments, such as circular layouts, cannot currently be handled within this framework.
    \item \textbf{Simplified environment representation:}
    Environments in our methodology and experiments are modeled as \acp{MDP}, a common abstraction in the \ac{RL} literature. This provides a convenient and flexible mathematical formulation, but introduces several simplifications. For example, agents are not physically situated in a maze but instead occupy abstract states and interact with the environment through the agent-environment interface (Figure~\ref{fig:RL-loop}). Obstacle changes are handled by modifying the reward function of the \ac{MDP} and by adjusting the positions in the grid, used by the visualization tool. Consequently, agents do not need sensors or LiDAR to perceive real-time changes, as such information is abstracted away within the \ac{MDP} model.  
    This abstraction also simplifies the federated Q-learning setup, where multiple agents interact with the same sub-environment without accounting for collisions with one another. In a physical deployment, however, agents would need to avoid both obstacles and other agents, which adds significant complexity. Thus, our methodology is best viewed as a software-based solution rather than one that can be directly applied to real-world robotic systems.  
    At the same time, the \ac{MDP} formulation is a strength: it makes our approach extensible beyond path planning. Additional environment elements (e.g., new obstacle types or dynamic features) can be incorporated simply by modifying the \ac{MDP}, without altering the agents’ internal logic. This makes the methodology adaptable to a broader range of problems, though this potential was not tested in this technical report. Nonetheless, the reliance on simplified \ac{MDP}-based abstractions highlights a gap between our simulation-based results and direct real-world applicability.
\end{itemize}

In conclusion, while our best-performing approaches do not represent a universal solution for all dynamic path planning problems, they provide a robust, scalable, and efficient alternative to existing methods, particularly in grid-like environments. The identified limitations point toward important directions for future research and practical adaptation.

\subsection{Threats to the Validity} \label{sec:threats-validity}

In this section, we discuss potential threats to the validity of our study. Three validity categories are considered: construct validity, internal validity, and external validity, where the latter two are widely recognized in research. The specific definitions employed for these categories are taken from a study conducted by Challenger et al., which proposes a systematic approach to evaluating domain-specific modeling language environments for \acp{MAS}~\cite{challenger2016systematic}.

\subsubsection{Construct Validity}

Construct validity refers to the extent to which the operational measures used in the study truly represent the concepts the researcher intends to investigate according to the research questions.

First, we address the lack of relationships between sub-environments to ensure that when path planning fails in a specific sub-environment, related sub-environments can be leveraged to adapt to changes in the environment. We address this limitation by employing the hierarchical decomposition framework described in Section~\ref{sec:hierarchical-decomposition}, combined with the tree strategy presented in Section~\ref{sec:strategy}, which allows retraining at different hierarchical levels.

To evaluate the effectiveness of our approaches in overcoming this issue, we consider the accuracy metric (success rate), which measures the proportion of positions in the environment from which an agent can successfully navigate to a charging station. Approaches that consistently achieve high levels of accuracy, particularly in hard environments, demonstrate their ability to address this limitation.

However, high accuracy alone does not guarantee efficiency, as the paths found might be unnecessarily long. Therefore, we also consider path length as an evaluation metric. Approaches that achieve both high accuracy and relatively short path lengths can be regarded as more optimal.

Second, we eliminate the reliance on a global path planner, which does not scale well as the environment size increases (similar to traditional approaches such as A*). The time required to find a path with a global planner grows exponentially with the number of states, which itself increases exponentially with environment size. Our methodology instead relies solely on local path planners operating within assigned regions, thereby addressing scalability concerns. To assess scalability and efficiency, we rely on two metrics: initial training time and adaptation time.

The initial training time indicates how long it takes an approach to explore the initially unknown environment and establish paths connecting positions to charging stations. Adaptation time measures how quickly an approach adjusts to changes during the environment simulation, reflecting the responsiveness of the system. Adaptation time is measured only after the initial training phase is complete.

These metrics provide insights into scalability by evaluating performance across environment sizes ranging from $20 \times 20$ to $300 \times 300$. They also indicate efficiency in terms of adaptation time. However, both metrics are influenced by computational resources, i.e., machines with faster or more numerous \ac{CPU} cores will naturally achieve shorter training and adaptation times. Therefore, the relative differences between approaches should be prioritized over absolute timing values.

Lastly, our evaluation procedure includes multiple environment complexities and incorporates more realistic dynamics by allowing multiple obstacle changes per time step. To understand the impact of this evaluation setup, all metrics should be considered together to draw meaningful conclusions.

\subsubsection{Internal Validity}

Internal validity concerns the degree to which a causal relationship between the treatment and observed outcomes can be established.

Switching from single-agent Q-learning to federated Q-learning resulted in improvements across all evaluation metrics, without altering the core methodology (i.e., hierarchical decomposition and tree strategy). The federated Q-learning approaches achieved slightly higher and more consistent accuracy over time, reduced initial training and adaptation times, lowered cumulative adaptation time, and produced shorter average path lengths.

However, both single-agent Q-learning and federated Q-learning involve inherent randomness in exploring the environment or sub-environments. This randomness means that results may vary slightly between runs: differences in adaptation time, accuracy, or other metrics may increase or decrease. Another factor contributing to differences between the two approaches is the stopping criterion: single-agent Q-learning approaches use a convergence-based criterion for retraining termination, whereas federated Q-learning approaches use a fixed number of iterations determined by the sub-environment size.

The use of hierarchical decomposition and the tree strategy also supports higher accuracy in harder environments. This is evident from the lower accuracy of the \texttt{onlyTrainLeafNodes} approach compared to other approaches. Again, due to randomness, the magnitude of these differences may vary slightly between runs.

\subsubsection{External Validity}

External validity concerns the extent to which the results of this study can be generalized to other settings.

Our results are based on a single simulation for each environment size and difficulty, with one environment configuration and one sequence of obstacle changes. This approach ensures reproducibility (via fixed seeds), consistency across approaches by using the same environment configuration and sequence of obstacle changes, and facilitates detailed analysis of results for a specific simulation. However, it limits generalizability.

An alternative would be to run multiple simulations per size and difficulty, averaging or aggregating results across runs. While this could offer a broader view, it is computationally expensive, especially for large and complex environments in which many time steps are simulated, and could take a significant amount of time for even a modest number of repetitions (e.g., 5-10 runs). Moreover, averaging can hide important outliers, as a single poor-performing run may be masked by several good runs.

Although our evaluation covers various sizes and difficulty levels, the results cannot be generalized to any environmental type. The hierarchical decomposition mechanism assumes square or rectangular environments with known dimensions. Results may differ in environments with different geometries, obstacle distributions, or complexity definitions. Furthermore, while our dynamic environment simulation is more realistic than simpler setups, it does not fully replicate the unpredictability and complexity of real-world dynamic environments, which further limits generalizability.

\clearpage

\section{Conclusion and Future Work} \label{sec:conclusion-future-work}

In this technical report, we addressed several limitations identified in the methodology and evaluation framework proposed by Yarahmadi et al. Their approach relied on a combination of global and local path planners, where the local \ac{RL}-based planners handled individual sub-environments. However, this dependency on a global planner is not always realistic, especially in large, dynamic environments, where the global planner becomes inefficient due to increased uncertainty and poor scalability concerning environment size.

Moreover, in their methodology, local \ac{RL} agents were strictly bound to their sub-environments. This design posed challenges when changes rendered charging stations unreachable within a specific sub-environment, or when no charging station was present at all. The absence of a mechanism to handle such cases meant the system could fail to find valid paths under dynamic conditions.

In terms of evaluation, their testing was limited: the environments were relatively simple, and the simulation of dynamic changes was overly constrained, only allowing one obstacle change at a time.

To overcome these limitations, we contributed the following advancements:

\begin{itemize}
    \item We designed a hierarchical decomposition mechanism that represents the environment as a tree of sub-environments. This enables precise control over the retraining level and sub-environment size, preventing unnecessary retraining on large sections of the environment and improving adaptability.
    \item We introduced a retraining mechanism based on the success rate of a sub-environment. This criterion ensures retraining is performed only when changes significantly degrade the current policy's effectiveness, promoting efficient learning.
    \item We developed both single-agent Q-learning and federated Q-learning approaches using the above contributions. The single-agent method builds on standard Q-learning with additional optimizations, while the federated Q-learning approaches let multiple agents learn in parallel and aggregate their local Q-table estimates to accelerate learning.
    \item We evaluated our methods across multiple environment complexities and sizes, benchmarking them against two baselines: \texttt{A* Static} and \texttt{A* Oracle}, the latter serving as a theoretically optimal benchmark.
    \item We implemented a visual policy tracker that visualizes how the learned policy evolves in response to environmental changes. This visualization can be applied to the three more advanced approaches (\texttt{singleAgent}, \texttt{fedAsynQ\_EqAvg}, and \texttt{fedAsynQ\_ImAvg}) and used in $20 \times 20$ and $50 \times 50$ environments. While the tracker can technically be applied to larger environments, the resulting visualizations become cluttered and are therefore not recommended.
\end{itemize}

From our simulation results, we conclude that the federated Q-learning approaches, \texttt{fedAsynQ\_EqAvg} and \texttt{fedAsynQ\_ImAvg}, performed best overall. These approaches consistently achieved high accuracy across all tested environment sizes and difficulties, with accuracy levels close to that of the \texttt{A* Oracle} baseline. Their adaptation times were low, due to the success rate-based retraining condition, and their path lengths were relatively short, though not as optimal as A*’s.

Between the two, \texttt{fedAsynQ\_EqAvg} showed slightly better performance. It required less cumulative adaptation time and achieved marginally shorter path lengths. Moreover, it is computationally simpler, assuming equal contribution from each agent during aggregation. These results suggest that even a basic federated averaging scheme can outperform more complex ones in this setting.

Despite the promising results, there remain several directions for improvement and further exploration:

\begin{itemize}
    \item \textbf{Integration of \ac{Deep RL}:} Our work is based on tabular Q-learning, which, although effective, does not scale well to large or continuous state spaces. \ac{Deep RL} methods, such as \ac{DQN} or actor-critic algorithms, are increasingly popular due to their scalability and generalization capabilities. Incorporating \ac{Deep RL} into our framework could improve performance and adaptability in larger and more complex environments.
    \item \textbf{Enhancing Path Optimality:} The \ac{RL}-based approaches still struggle to consistently find the shortest paths in dynamic environments. Even the best-performing methods, \texttt{fedAsynQ\_EqAvg} and \texttt{fedAsynQ\_ImAvg}, achieve reasonable average path lengths but leave room for improvement compared to the \texttt{A* Oracle} baseline. Future work could explore techniques such as action masking to refine action selection, by dynamically restricting agents to valid or promising actions, thereby reducing exploration of implausible states and improving both efficiency and path optimality.
    \item \textbf{Reducing Initial Training Time:} A key limitation of the current framework is the time-consuming initial training phase, particularly in large environments with many sub-environments. Future research could investigate transfer learning strategies that reuse policies or Q-tables from previously trained sub-environments to accelerate learning in new ones. This would be especially beneficial in large-scale scenarios where simultaneous training of all leaf nodes is computationally infeasible.
    \item \textbf{Scaling to Larger Environments:} Although the proposed methods were tested on environments up to $300 \times 300$ in size, further experiments on larger maps such as $500 \times 500$, $1000 \times 1000$, or beyond could provide deeper insights into the scalability and robustness of our framework under more demanding conditions.
    \item \textbf{Extending to Alternative Environment Shapes:} The current hierarchical decomposition assumes known, rectangular, or square environments. A first promising extension involves expanding the two-dimensional framework into three dimensions, enabling \acp{UAV} to navigate dynamic 3D environments. However, this introduces substantial complexity: the decomposition must recursively partition 3D space, the retraining condition and learning procedures must be adapted, the state space grows cubically rather than quadratically, and the action set expands from 8 (2D) to 26 (3D). Another potential extension is supporting environments with arbitrary or irregular shapes, thereby broadening the framework’s applicability beyond grid-like structures.
    \item \textbf{Exploring Alternative Aggregation Strategies:}   While the current federated Q-learning approaches use equal or importance-weighted averaging to aggregate Q-tables, future work could explore more advanced aggregation mechanisms such as adaptive weighting or meta-learning-based aggregation to further enhance learning efficiency and stability.
    \item \textbf{Real-World Robotic Implementation:} A key direction for future work is implementing the framework in real-world robotic systems. The current implementation abstracts dynamic environments using \acp{MDP}, which simplifies several real-world complexities, such as environmental disturbances (e.g., currents in aquatic environments) and sensor noise. In a physical system, agents could leverage local perception technologies such as LiDAR or sonar for real-time environment sensing and more informed decision-making. Moreover, while agents in the current framework operate in parallel, they do not explicitly model other agents’ positions or behaviors. Incorporating inter-agent awareness and collision avoidance mechanisms would be crucial for deploying this system in real-world multi-robot scenarios.
\end{itemize}

In summary, this technical report presents a scalable and adaptive \ac{RL}-based path planning framework for dynamic environments, capable of operating efficiently across varying complexities and sizes. While tailored to path planning, the explored principles of hierarchical decomposition, conditional retraining, and federated Q-learning are more broadly applicable to other domains that require distributed, scalable decision-making under uncertainty.

\clearpage

\phantomsection
\addcontentsline{toc}{section}{References}
\bibliographystyle{unsrt}
\bibliography{mybib}

\begin{thebibliography}{10}

\bibitem{morad1992path}
AA~Morad, AB~Cleveland~Jr, YJ~Beliveau, VD~Fransisco, and SS~Dixit.
\newblock Path-finder: Ai-based path planning system.
\newblock {\em Journal of computing in civil engineering}, 6(2):114--128, 1992.

\bibitem{sivakumar2003automated}
P~{\'a}L Sivakumar, Koshy Varghese, and N~Ramesh Babu.
\newblock Automated path planning of cooperative crane lifts using heuristic search.
\newblock {\em Journal of computing in civil engineering}, 17(3):197--207, 2003.

\bibitem{kim2003construction}
Sung-Keun Kim, Jeffrey~S Russell, and Kyo-Jin Koo.
\newblock Construction robot path-planning for earthwork operations.
\newblock {\em Journal of computing in civil engineering}, 17(2):97--104, 2003.

\bibitem{song2019construction}
Siyuan Song and Eric Marks.
\newblock Construction site path planning optimization through bim.
\newblock In {\em ASCE International Conference on Computing in Civil Engineering 2019}, pages 369--376. American Society of Civil Engineers Reston, VA, 2019.

\bibitem{cai2023prediction}
Jiannan Cai, Ao~Du, Xiaoyun Liang, and Shuai Li.
\newblock Prediction-based path planning for safe and efficient human--robot collaboration in construction via deep reinforcement learning.
\newblock {\em Journal of computing in civil engineering}, 37(1):04022046, 2023.

\bibitem{lu2025fire}
Juan Lu, Yale Lan, and Minghai Li.
\newblock Fire evacuation path dynamic planning system based on improved a* algorithm.
\newblock In {\em International Conference on Civil Engineering and Architecture}, pages 143--154. Springer, 2025.

\bibitem{ting2002path}
Yung Ting, WI~Lei, and HC~Jar.
\newblock A path planning algorithm for industrial robots.
\newblock {\em Computers \& Industrial Engineering}, 42(2-4):299--308, 2002.

\bibitem{gochev2017path}
Ivan Gochev, Gorjan Nadzinski, and Mile Stankovski.
\newblock Path planning and collision avoidance regime for a multi-agent system in industrial robotics.
\newblock {\em Machines. Technologies. Materials.}, 11(11):519--522, 2017.

\bibitem{zhang2018path}
Haojian Zhang, Yunkuan Wang, Jun Zheng, and Junzhi Yu.
\newblock Path planning of industrial robot based on improved rrt algorithm in complex environments.
\newblock {\em IEEE Access}, 6:53296--53306, 2018.

\bibitem{fu2018improved}
Bing Fu, Lin Chen, Yuntao Zhou, Dong Zheng, Zhiqi Wei, Jun Dai, and Haihong Pan.
\newblock An improved a* algorithm for the industrial robot path planning with high success rate and short length.
\newblock {\em Robotics and Autonomous Systems}, 106:26--37, 2018.

\bibitem{chen2023intelligent}
Yun Chen, Jinfeng Wu, Chaoshuai He, and Si~Zhang.
\newblock Intelligent warehouse robot path planning based on improved ant colony algorithm.
\newblock {\em IEEE Access}, 11:12360--12367, 2023.

\bibitem{tordesillas2021mader}
Jesus Tordesillas and Jonathan~P How.
\newblock Mader: Trajectory planner in multiagent and dynamic environments.
\newblock {\em IEEE Transactions on Robotics}, 38(1):463--476, 2021.

\bibitem{yarahmadi2024comp}
Hossein Yarahmadi, Hussein Marah, and Moharram Challenger.
\newblock Comparative analysis of classic and reinforcement learning approaches for robot navigation in dynamic environments.
\newblock University of Antwerp and Flanders Make Strategic Research Center, 2024.

\bibitem{van2008multi}
Wiebe Van~der Hoek and Michael Wooldridge.
\newblock Multi-agent systems.
\newblock {\em Foundations of Artificial Intelligence}, 3:887--928, 2008.

\bibitem{jin2025comprehensive}
Weiqiang Jin, Hongyang Du, Biao Zhao, Xingwu Tian, Bohang Shi, and Guang Yang.
\newblock A comprehensive survey on multi-agent cooperative decision-making: Scenarios, approaches, challenges and perspectives.
\newblock {\em arXiv preprint arXiv:2503.13415}, 2025.

\bibitem{pashenkova1996value}
Elena Pashenkova, Irina Rish, and Rina Dechter.
\newblock Value iteration and policy iteration algorithms for markov decision problem.
\newblock In {\em AAAI’96: Workshop on Structural Issues in Planning and Temporal Reasoning. Citeseer}, 1996.

\bibitem{watkins1992q}
Christopher~JCH Watkins and Peter Dayan.
\newblock Q-learning.
\newblock {\em Machine learning}, 8:279--292, 1992.

\bibitem{panov2018grid}
Aleksandr~I Panov, Konstantin~S Yakovlev, and Roman Suvorov.
\newblock Grid path planning with deep reinforcement learning: Preliminary results.
\newblock {\em Procedia computer science}, 123:347--353, 2018.

\bibitem{lin2019end}
Juntong Lin, Xuyun Yang, Peiwei Zheng, and Hui Cheng.
\newblock End-to-end decentralized multi-robot navigation in unknown complex environments via deep reinforcement learning.
\newblock In {\em 2019 IEEE International Conference on Mechatronics and Automation (ICMA)}, pages 2493--2500. IEEE, 2019.

\bibitem{gan2019new}
Xingli Gan, Hongliang Guo, and Zhan Li.
\newblock A new multi-agent reinforcement learning method based on evolving dynamic correlation matrix.
\newblock {\em IEEE Access}, 7:162127--162138, 2019.

\bibitem{luis2021multiagent}
Samuel~Yanes Luis, Daniel~Guti{\'e}rrez Reina, and Sergio L~Toral Mar{\'\i}n.
\newblock A multiagent deep reinforcement learning approach for path planning in autonomous surface vehicles: The ypacara{\'\i} lake patrolling case.
\newblock {\em IEEE Access}, 9:17084--17099, 2021.

\bibitem{li2023ace}
Chuming Li, Jie Liu, Yinmin Zhang, Yuhong Wei, Yazhe Niu, Yaodong Yang, Yu~Liu, and Wanli Ouyang.
\newblock Ace: Cooperative multi-agent q-learning with bidirectional action-dependency.
\newblock In {\em Proceedings of the AAAI conference on artificial intelligence}, volume~37, pages 8536--8544, 2023.

\bibitem{shen2024autonomous}
Zhilong Shen, Yongwei Chi, Yu~Bai, Xiaojing Liao, Peiyu Zhao, Jianjiang Lu, Qian Niu, and Qinya Dai.
\newblock Autonomous navigation with minimal sensors in dynamic warehouse environments: a multi-agent reinforcement learning approach with curriculum learning enhancement.
\newblock {\em Research Square}, 2024.

\bibitem{yin2024cooperative}
Jiaming Yin, Weixiong Rao, Yu~Xiao, and Keshuang Tang.
\newblock Cooperative path planning with asynchronous multiagent reinforcement learning.
\newblock {\em arXiv preprint arXiv:2409.00754}, 2024.

\bibitem{wang2025lns2+}
Yutong Wang, Tanishq Duhan, Jiaoyang Li, and Guillaume Sartoretti.
\newblock Lns2+ rl: Combining multi-agent reinforcement learning with large neighborhood search in multi-agent path finding.
\newblock In {\em Proceedings of the AAAI Conference on Artificial Intelligence}, volume~39, pages 23343--23350, 2025.

\bibitem{chang2021reinforcement}
Lu~Chang, Liang Shan, Chao Jiang, and Yuewei Dai.
\newblock Reinforcement based mobile robot path planning with improved dynamic window approach in unknown environment.
\newblock {\em Autonomous robots}, 45:51--76, 2021.

\bibitem{bae2019multi}
Hyansu Bae, Gidong Kim, Jonguk Kim, Dianwei Qian, and Sukgyu Lee.
\newblock Multi-robot path planning method using reinforcement learning.
\newblock {\em Applied sciences}, 9(15):3057, 2019.

\bibitem{liu2020mapper}
Zuxin Liu, Baiming Chen, Hongyi Zhou, Guru Koushik, Martial Hebert, and Ding Zhao.
\newblock Mapper: Multi-agent path planning with evolutionary reinforcement learning in mixed dynamic environments.
\newblock In {\em 2020 IEEE/RSJ International Conference on Intelligent Robots and Systems (IROS)}, pages 11748--11754. IEEE, 2020.

\bibitem{guan2022ab}
Huifeng Guan, Yuan Gao, Min Zhao, Yong Yang, Fuqin Deng, and Tin~Lun Lam.
\newblock Ab-mapper: Attention and bicnet based multi-agent path planning for dynamic environment.
\newblock In {\em 2022 IEEE/RSJ International Conference on Intelligent Robots and Systems (IROS)}, pages 13799--13806. IEEE, 2022.

\bibitem{gao2025scalable}
Qiang Gao, Shuhao Li, Yuehui Ji, Junjie Liu, and Yu~Song.
\newblock Scalable path planning algorithm for multi-unmanned surface vehicles based on multi-agent deep deterministic policy gradient.
\newblock {\em Ocean Engineering}, 320:120243, 2025.

\bibitem{choi2021reinforcement}
Jaewan Choi, Geonhee Lee, and Chibum Lee.
\newblock Reinforcement learning-based dynamic obstacle avoidance and integration of path planning.
\newblock {\em Intelligent Service Robotics}, 14:663--677, 2021.

\bibitem{wang2020mobile}
Binyu Wang, Zhe Liu, Qingbiao Li, and Amanda Prorok.
\newblock Mobile robot path planning in dynamic environments through globally guided reinforcement learning.
\newblock {\em IEEE Robotics and Automation Letters}, 5(4):6932--6939, 2020.

\bibitem{chang2023hierarchical}
Lu~Chang, Liang Shan, Weilong Zhang, and Yuewei Dai.
\newblock Hierarchical multi-robot navigation and formation in unknown environments via deep reinforcement learning and distributed optimization.
\newblock {\em Robotics and Computer-Integrated Manufacturing}, 83:102570, 2023.

\bibitem{zhou2024novel}
Yatong Zhou, Xiaoran Kong, Kuo-Ping Lin, and Liangyu Liu.
\newblock Novel task decomposed multi-agent twin delayed deep deterministic policy gradient algorithm for multi-uav autonomous path planning.
\newblock {\em Knowledge-Based Systems}, 287:111462, 2024.

\bibitem{guo2024decentralized}
Dong Guo, Shouwen Ji, Yanke Yao, and Cheng Chen.
\newblock A decentralized path planning model based on deep reinforcement learning.
\newblock {\em Computers and Electrical Engineering}, 117:109276, 2024.

\bibitem{cai2020mobile}
Kuanqi Cai, Chaoqun Wang, Jiyu Cheng, Clarence~W De~Silva, and Max Q-H Meng.
\newblock Mobile robot path planning in dynamic environments: A survey.
\newblock {\em arXiv preprint arXiv:2006.14195}, 2020.

\bibitem{madridano2021trajectory}
{\'A}ngel Madridano, Abdulla Al-Kaff, David Mart{\'\i}n, and Arturo De~La~Escalera.
\newblock Trajectory planning for multi-robot systems: Methods and applications.
\newblock {\em Expert Systems with Applications}, 173:114660, 2021.

\bibitem{boroujeni2017flexible}
Zahra Boroujeni, Daniel Goehring, Fritz Ulbrich, Daniel Neumann, and Raul Rojas.
\newblock Flexible unit a-star trajectory planning for autonomous vehicles on structured road maps.
\newblock In {\em 2017 IEEE international conference on vehicular electronics and safety (ICVES)}, pages 7--12. IEEE, 2017.

\bibitem{qie2019joint}
Han Qie, Dianxi Shi, Tianlong Shen, Xinhai Xu, Yuan Li, and Liujing Wang.
\newblock Joint optimization of multi-uav target assignment and path planning based on multi-agent reinforcement learning.
\newblock {\em IEEE access}, 7:146264--146272, 2019.

\bibitem{cruz2017path}
David~Luviano Cruz and Wen Yu.
\newblock Path planning of multi-agent systems in unknown environment with neural kernel smoothing and reinforcement learning.
\newblock {\em Neurocomputing}, 233:34--42, 2017.

\bibitem{almazrouei2023dynamic}
Khawla Almazrouei, Ibrahim Kamel, and Tamer Rabie.
\newblock Dynamic obstacle avoidance and path planning through reinforcement learning.
\newblock {\em Applied Sciences}, 13(14):8174, 2023.

\bibitem{wang2019improved}
Qingqing Wang, Hong Liu, Kaizhou Gao, and Le~Zhang.
\newblock Improved multi-agent reinforcement learning for path planning-based crowd simulation.
\newblock {\em IEEE Access}, 7:73841--73855, 2019.

\bibitem{cormen2022introduction}
Thomas~H Cormen, Charles~E Leiserson, Ronald~L Rivest, and Clifford Stein.
\newblock {\em Introduction to algorithms}.
\newblock MIT press, 2022.

\bibitem{woo2025blessing}
Jiin Woo, Gauri Joshi, and Yuejie Chi.
\newblock The blessing of heterogeneity in federated q-learning: Linear speedup and beyond.
\newblock {\em Journal of Machine Learning Research}, 26(26):1--85, 2025.

\bibitem{challenger2016systematic}
Moharram Challenger, Geylani Kardas, and Bedir Tekinerdogan.
\newblock A systematic approach to evaluating domain-specific modeling language environments for multi-agent systems.
\newblock {\em Software Quality Journal}, 24(3):755--795, 2016.

\end{thebibliography}

\clearpage

\begin{appendices}

\section{Policy Visualization} \label{sec:policy-visualization}

This section quickly covers the policy visualization tool that I created, which is implemented in the \texttt{policy\allowbreak visualizer.cpp} file\footnote{\url{https://github.com/micss-lab/MARL4DynaPath/blob/main/src/policyvisualizer.cpp}}. The idea of the tool is that for each time step of an environment simulation, we display, for each position, the highest-value action. In other words, we display the policies learned by the agents during adaptation to changes in the environment.

For practical purposes, the visualization is only utilized for the three approaches using the tree strategy described in Section~\ref{sec:strategy}, which are \texttt{singleAgent}, \texttt{fedAsynQ\_EqAvg}, and \texttt{fedAsynQ\_ImAvg} to be specific. Since we want to show the highest-value action for each state/position in the environment in a visible way, the visualization is only shown for an environment of size $20 \times 20$ and for the edge case environment of size $50 \times 50$ in Figure~\ref{fig:edge-case}. For larger sizes, the visualization would become too cluttered, and changes would barely be noticed. Additionally, visualizing the policy for larger environments also requires more time, as the number of positions grows exponentially, causing the visualization to be slower, but this is not necessarily an issue.

\begin{figure}[t]
    \centering
    \includegraphics[width=0.49\linewidth]{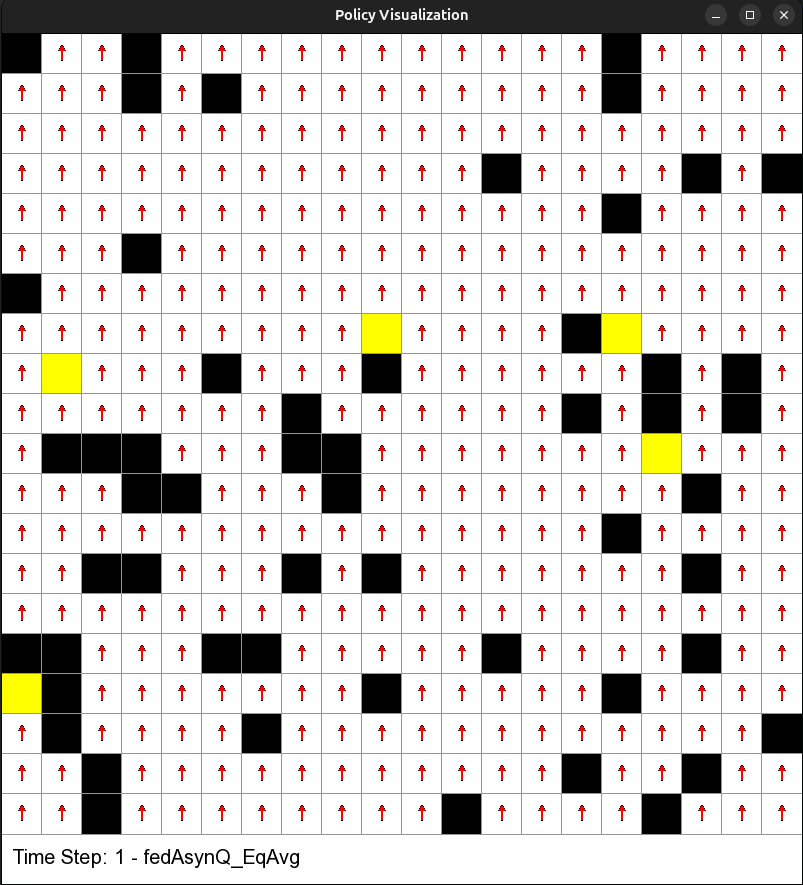}
    \includegraphics[width=0.49\linewidth]{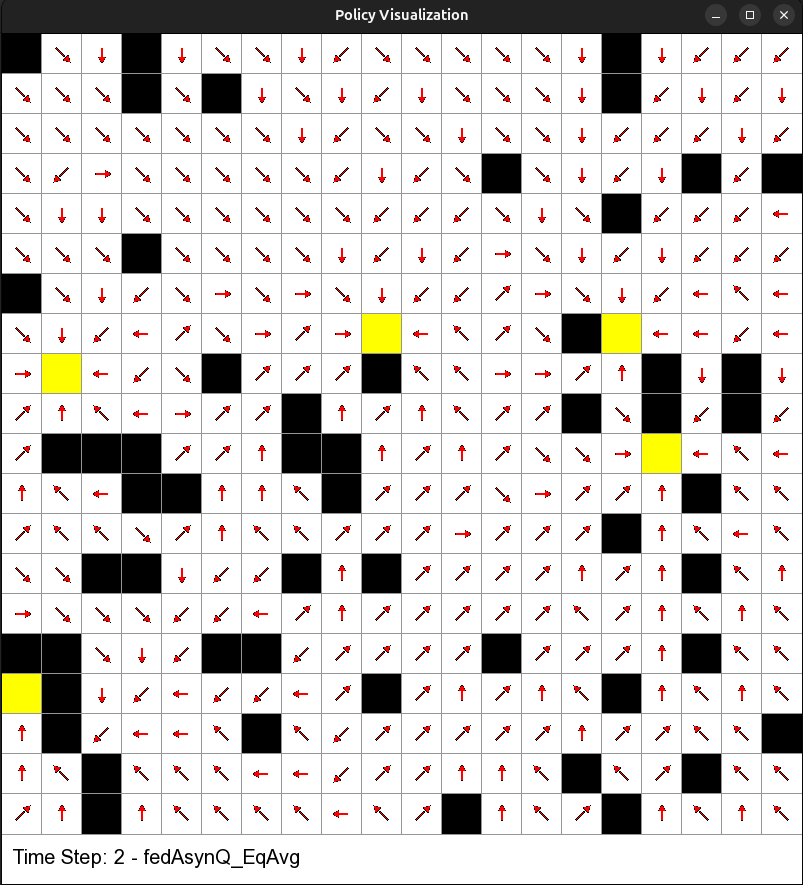}
    \includegraphics[width=0.49\linewidth]{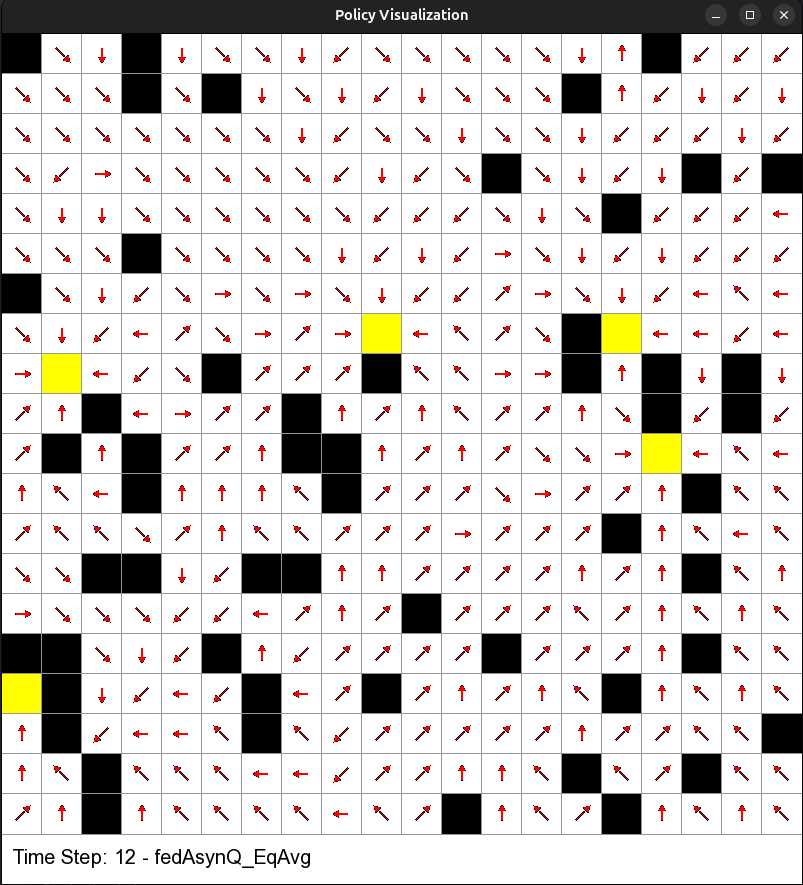}
    \includegraphics[width=0.49\linewidth]{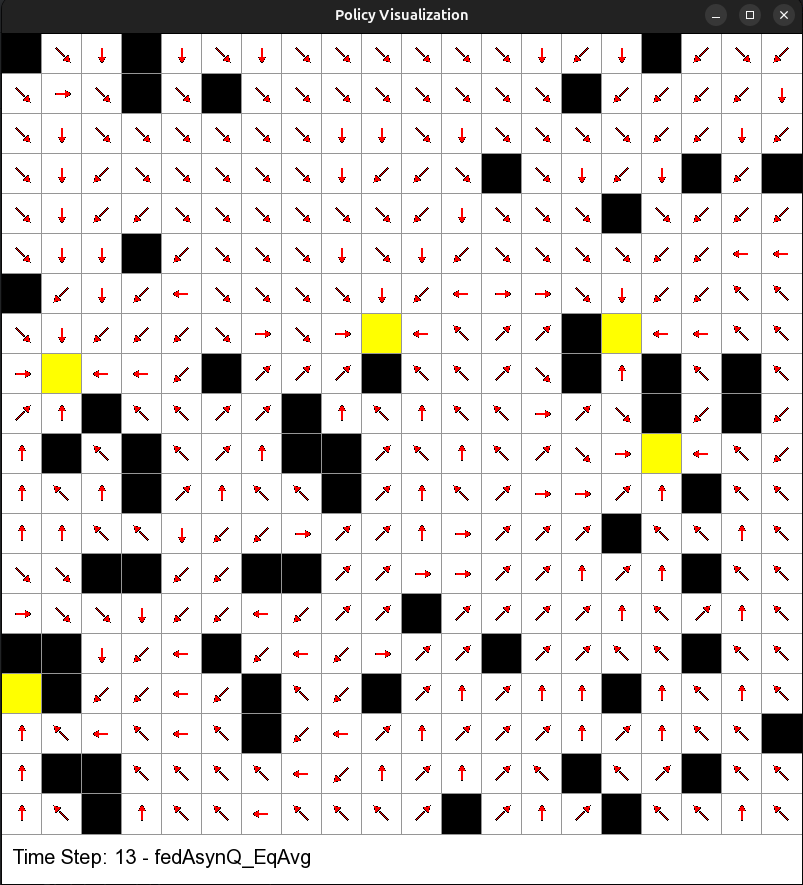}
    \caption{Policy visualization of the \texttt{fedAsynQ\_EqAvg} approach on an easy $20 \times 20$ environment across multiple time steps.}
    \label{fig:visualization-easy-20x20}
\end{figure}

A policy visualization of the \texttt{fedAsynQ\_EqAvg} approach on an easy $20 \times 20$ environment across multiple time steps is shown in Figure~\ref{fig:visualization-easy-20x20}. In the top-left, the untrained policy is shown before the start of the simulation. As can be observed, all arrows (in red) are pointing upwards. This is because the Q-table is initialized with all zeros, meaning that the first encountered action (being 0) is the one that is taken. This corresponds to the upward action.

In the top-right, the simulation started, and an obstacle change took place. The learned policy at that time step is also shown, as for all positions, the arrows changed directions. For the majority of the positions, we can see that the action displayed is relatively good, pointing in the direction of a charging station or at least following a path towards one. This shows that the initial training was successful and that the highest-value action corresponds in most positions to a plausible action.

After a certain number of time steps, we enter time step 12, visualized in the bottom-left. As can be seen, the policy has not changed with respect to time step 2, which means that the retraining condition determined that retraining is not necessary, as the changes did not sufficiently affect the accuracy of the current policy. When entering time step 13 (bottom-right), another obstacle change took place, and retraining of the environment was triggered. Because the environment is $20 \times 20$ in size, the hierarchical tree only consists of a root node, representing the entire environment. Therefore, when retraining is triggered, the entire environment is retrained, which is not an issue, since the entire environment is only $20 \times 20$ in size. After retraining at time step 13, the adapted policy is visualized.

This simulation then continues until time step $2 \times 20 = 40$ and stops. Because we are considering a $20 \times 20$ environment, the regional adaptation is not visible, as the entire environment is always retrained. To provide an intuition of what the visualization will show when operating on larger environments, e.g., $50 \times 50$, $100 \times 100$, etc., we have to think in terms of the hierarchical tree that is created for the environment. All of our approaches prioritize retraining at the lowest level of the hierarchy, corresponding to the smallest sub-environments. If (re)training these sub-environments seems insufficient, we go up the hierarchy and retrain at a larger scope/region of the environment.

During the simulation of the environment, this would mean that when the retraining condition for a specific sub-environment is satisfied at a certain time step, it will be (re)trained. In the visualization, this will be shown as only a region adapting or changing its policy, instead of the entire environment. An example run on a $50 \times 50$ environment is shown in the README file\footnote{\url{https://github.com/micss-lab/MARL4DynaPath/blob/main/README.md}} on GitHub.

\clearpage

\section{Implementation Details and Running Experiments} \label{sec:implementation-running-details}

This section provides an overview of the implementation developed in this report, along with instructions for running the experiments. It also explains how randomness and seed values are used to ensure reproducibility.

\subsection{Implementation Details}

The implementation developed in this report is available as an open-source project on GitHub\footnote{\url{https://github.com/micss-lab/MARL4DynaPath}}. The code is written in C++, chosen for its execution speed and support for multithreading, which are two key factors for our proposed approaches. The repository also contains a README file\footnote{\url{https://github.com/micss-lab/MARL4DynaPath/blob/main/README.md}} with additional information about the file structure and detailed installation and running instructions. Two examples of the policy visualization tool are also included.

The policy visualization tool, described in Section~\ref{sec:policy-visualization}, is implemented using version 2.6.1 of the \ac{SFML} library\footnote{\url{https://www.sfml-dev.org/}}, which integrates well with C++. To install \ac{SFML} on Ubuntu, run the following command:

\begin{verbatim}
sudo apt-get install libsfml-dev
\end{verbatim}

For installation on other platforms or for manual setup instructions, refer to the Download page\footnote{\url{https://www.sfml-dev.org/download/}} or the Tutorial page\footnote{\url{https://www.sfml-dev.org/tutorials/2.6/}}. The provided \texttt{CMakeLists.txt} file in the project already includes the necessary configuration to locate and link \ac{SFML}.

The repository also includes the \texttt{visualizations.py} Python script\footnote{\url{https://github.com/micss-lab/MARL4DynaPath/blob/main/src/visualizations.py}}, which generates all plots (e.g., accuracy, adaptation time, path lengths). The script depends on the \texttt{matplotlib} and \texttt{pandas} packages, which can be installed with:

\begin{verbatim}
pip3 install matplotlib pandas
\end{verbatim}

\subsection{Running Experiments}

We recommend using the CLion IDE for building and running the experiments, as it simplifies setup and execution:

\begin{enumerate}
    \item Fork or download the repository, then navigate to the project directory using a terminal.
    \item Launch the project in CLion with: \verb|clion .|
    \item When prompted, click Trust Project. 
    \item When prompted with the Project Wizard, tick the checkbox to reload the CMake project on editing \texttt{CMakeLists.txt} and click OK.
    \item Change the \verb|CMAKE_RUNTIME_OUTPUT_DIRECTORY| variable in the \texttt{CMakeLists.txt} file to the absolute path of the project's root directory.
    \item Click the hammer icon to build the project (the initial build may take some time).
    \item Click the green play icon to start running the experiments.
\end{enumerate}

\subsection{Modifying Experiment Settings}

\begin{itemize}
    \item \textbf{Enable or disable policy visualization:}
    Set the argument of the \texttt{runFullExperiment} function in the \texttt{main.cpp} file to \texttt{true} to enable visualization (recommended only for small environments) or \texttt{false} to disable it.

    \item \textbf{Change environment sizes and difficulties:}
    In \texttt{experiments.cpp}, modify the sizes and difficulties lists within the \texttt{runFullExperiment} function.

    \item \textbf{Select approaches to test:}
    You can include or exclude any of the following:
    \begin{itemize}
        \item \texttt{A* Static}
        \item \texttt{A* Oracle}
        \item \texttt{onlyTrainLeafNodes}
        \item \texttt{singleAgent}
        \item \texttt{fedAsynQ\_EqAvg}
        \item \texttt{fedAsynQ\_ImAvg}
    \end{itemize}
\end{itemize}

After running the experiments, two files should be created in the project's root directory: \texttt{results.csv} and \texttt{results\_detailed.csv}. These are used by the Python script to generate plots. In the script, you can specify the output directory for these plots to be located. To create the plots, run the script from the directory containing both CSV files.

\subsection{Running the Edge Case}

To run the edge case experiment:

\begin{enumerate}
    \item In \texttt{experiments.cpp}, locate the line that sets the seed (line 101) and change it to \texttt{srand(d + 100)}, as indicated by the comment.
    \item Modify the list of environment sizes to only include sizes 20 and 50, while keeping all three difficulty levels.
    \item Build and run the experiments.
\end{enumerate}

The final environment considered will be the hard $50 \times 50$ environment, which serves as the edge case. Currently, there is no simpler method to find a similar edge case environment. To generate plots for the edge case, remove all preceding lines in \texttt{results.csv} and \texttt{results\_detailed.csv} that do not correspond to the hard $50 \times 50$ environment, and adjust the plots directory in the Python script to \verb|plots-edge-case|. Then, run the \texttt{visualizations.py} script from the project's root directory.

\subsection{Reproducibility and Randomness}

To ensure reproducibility, a fixed seed is used when generating the initial environment configuration and sequence of dynamic changes. This is set in the \texttt{experiments.cpp} file at line 101:

\vspace{10pt}
\begin{lstlisting}
    srand(d + 50);
\end{lstlisting}

Using a fixed seed ensures that the same environment and sequence of obstacle changes are generated in every run. All six approaches discussed in this report are evaluated on the same environment instance and change sequence, ensuring a fair and consistent comparison. This is essential for drawing meaningful conclusions about their relative performance.

\end{appendices}

\end{document}